\documentclass{article}

\usepackage[main, final]{neurips_2026}

\usepackage[utf8]{inputenc} 
\usepackage[T1]{fontenc}    
\usepackage{hyperref}       
\usepackage{url}            
\usepackage{booktabs}       
\usepackage{amsfonts}       
\usepackage{nicefrac}       
\usepackage{microtype}      
\usepackage{xcolor}         
\usepackage{graphicx}
\usepackage{tcolorbox} 
\usepackage{hyperref}
\usepackage{float}
\usepackage{caption}
\usepackage{enumitem}
\usepackage{etoc}
\definecolor{citeblue}{HTML}{114082}
\definecolor{boxgray}{HTML}{F2F4F8}

\hypersetup{
  colorlinks=true,
  urlcolor=black,
  linkcolor=citeblue,
  citecolor=citeblue
}
\usepackage{natbib}
\setcitestyle{authoryear,open={(},close={)}}
\usepackage{fontawesome5}

\newcommand{\figref}[1]{Fig.~\ref{#1}}
\newcommand{\tabref}[1]{Tab.~\ref{#1}}
\newcommand{\aref}[1]{\S\hyperref[#1]{\ref*{#1}}}

\usepackage{rotating}
\usepackage{array}
\usepackage{adjustbox}
\usepackage{makecell}
\usepackage{rotfloat}
\usepackage{tablefootnote}
\usepackage{siunitx}
\usepackage{pifont}
\title{\texttt{spora}:\\A Unified Multimodal Dataset\\and Platform for Spatial Proteomics}

\author{%
  \textbf{Benedikt von Querfurth}\textsuperscript{\normalfont1*}
  \quad
  \textbf{Johann Wenckstern}\textsuperscript{\normalfont1*}
  \quad
  \textbf{Eeshaan Jain}\textsuperscript{\normalfont1*}
  \quad
  \textbf{Lukas Klein}\textsuperscript{\normalfont1,2,3*}
  \\[0.4em]
  \textbf{Phil Cheng}\textsuperscript{4,5}
  \quad
  \textbf{Yexiang Cheng}\textsuperscript{1}
  \quad
  \textbf{Cedric Vincent-Cuaz}\textsuperscript{1}
  \quad
  \textbf{Martina Haberecker}\textsuperscript{2}
  \\[0.4em]
  \textbf{Pascal Frossard}\textsuperscript{1}
  \quad
  \textbf{Andreas Wicki}\textsuperscript{2,3} 
  \quad
  \textbf{Olivier Michielin}\textsuperscript{4,5} 
  \quad
  \textbf{Charlotte Bunne}\textsuperscript{1} 
  \\[0.8em]
  \normalfont\normalsize
  \textsuperscript{1}EPFL \quad
  \textsuperscript{2}University Hospital Zurich \quad
  \textsuperscript{3}University of Zurich \\
  \textsuperscript{4}University Hospital Geneva \quad
  \textsuperscript{5}University of Geneva
}

\usepackage{xspace}
\newcommand{\spora}{\texttt{spora}\xspace}
\newcommand{\sporaio}{\texttt{spora} \texttt{[io]}\xspace}
\newcommand{\sporadata}{\texttt{spora} \texttt{[data]}\xspace}
\newcommand{\sporabench}{\texttt{spora} \texttt{[bench]}\xspace}

\begin{document}

\maketitle

\begin{abstract}
\looseness -1 Spatial proteomics (SP) captures the molecular composition and spatial organization of tissues at single-cell resolution, opening a direct view of the cellular ecosystems that drive disease. Protein panels, acquisition platforms, and tissue contexts differ from one cohort to the next, and a coherent picture of tissue biology can only emerge from models trained to bridge this heterogeneity. Foundation models offer the natural path forward, but their training depends on large, harmonized corpora that span platforms, panels, and cohorts. No public resource currently approaches that scale; available data sit in small, narrowly scoped releases with incompatible formats and conventions.
We present \spora, a large-scale multimodal SP dataset of 13,546 standardized multiplex images from 5,477 patients across 35 cohorts and five major SP technologies. 
Curated with biomedical experts, \spora provides model-training-ready formats together with cell and tissue segmentations, structured clinical metadata, and, where available, paired H\&E enabling the training of virtual staining models.
It is complemented by \sporaio, a unified interface for tiling, sampling, and standardization across all cohorts and modalities, and \sporabench, a benchmark suite spanning cell phenotyping, diagnostic and therapeutic prediction as well as virtual staining tasks across twelve cohorts that places current SP foundation models on common ground against task-specific baselines.

\vspace{5pt}
\centering
\href{https://spora.bioimagearchive.org/}{\mbox{Data: %
  \raisebox{-0.2em}{\includegraphics[height=1em]{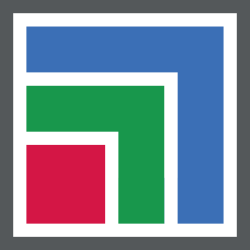}}~%
  \texttt{\textcolor{citeblue}{spora.bioimagearchive.org}}%
}}\\
\vspace{0.3em}
Docs: \href{https://spora.epfl.ch/docs/}{\texttt{\textcolor{citeblue}{\raisebox{-0.1em}{\faFile}\;spora.epfl.ch/docs}}}
\end{abstract}

\etocdepthtag.toc{mtmain}

\section{Introduction}

Tissue function arises from the molecular states of individual cells and their spatial organization. Spatial proteomics resolves both at single-cell resolution, measuring dozens of proteins across intact tissue sections through technologies such as imaging mass cytometry (IMC), CODEX (now PhenoCycler), MIBI, and Orion \citep{giesen2014highly, black2021codex, keren_mibi-tof_2019, lin2023high}. The resulting images expose immune cell phenotypes, checkpoint expression, stromal composition, and the architecture of the tumor microenvironment \citep{wu2024spatial, karimi2024method}. These features are increasingly relevant to questions in translational oncology, including immunotherapy response prediction, microenvironment characterization, and the identification of spatial patterns associated with recurrence and treatment resistance \citep{cords2023cancer, salie2025spatial, xiao2022multiplexed}. Across these applications, SP is moving from research use into clinical decision support \citep{quail2024revolutionizing, liu2025spatial}.

Realizing this trajectory at scale depends on AI, and the AI that has reshaped adjacent areas of biomedicine is fundamentally data-driven: performance scales with the volume and diversity of the data on which models are trained, and no single SP study supplies either at the level required. 
Protein panels, acquisition platforms, and tissue contexts differ from one cohort to the next, so the corpora needed to train such models can only be built by aggregating across studies. Foundation models (FMs) \citep{bommasani2021opportunities} are the clearest expression of this paradigm, pretrained at scale on diverse data and applied to specific downstream tasks, and their development has consistently followed the same pattern where large standardized (public) corpora enable successive waves of model development \citep{campanella2025clinical}: In histopathology, the Cancer Genome Atlas (TCGA) \citep{tomczak2015review} and HistAI \citep{nechaev2025histaiopensourcelargescaleslide} provide hundreds of thousands of standardized hematoxylin and eosin (H\&E) whole-slide images, seeding a generation of general-purpose FMs \citep{chen2024uni, zimmermann2024virchow2, hoptimus0, nechaev2024hibou}. In spatial transcriptomics, HEST-1k \citep{jaume2024hest} and STOmicsDB \citep{xu2024stomicsdb} consolidate paired transcriptomic and histology samples in standardized formats and supported the first FM pretraining and benchmarks for the modality. In single-cell RNA sequencing, CELLxGENE \citep{program_cz_2025} aggregates millions of cells from thousands of studies into a single queryable resource, providing the basis for cross-study atlases and FMs.

SP has yet to undergo this consolidation, and is structurally harder to unify than the modalities above. Acquisition technologies differ in resolution, intensity distribution, and file format. Antibody panels vary from one study to the next (\figref{fig:marker_overview}). Pairing with H\&E is inconsistent across cohorts. Existing machine learning models trained on SP reflect this fragmentation \citep{sorin2023single}: they are trained on small single-disease cohorts \citep{cords2023cancer}, single-cell imaging \citep{kraus2024masked}, or proprietary data the community cannot access or extend \citep{shaban2025foundation}, and the benchmarks built around them inherit the same limitations. For related work, see \autoref{sec:related_work}.

\begin{figure}[t]
    \centering
    \includegraphics[trim=0.2cm 2cm 0.2cm 0cm,clip,width=\textwidth]{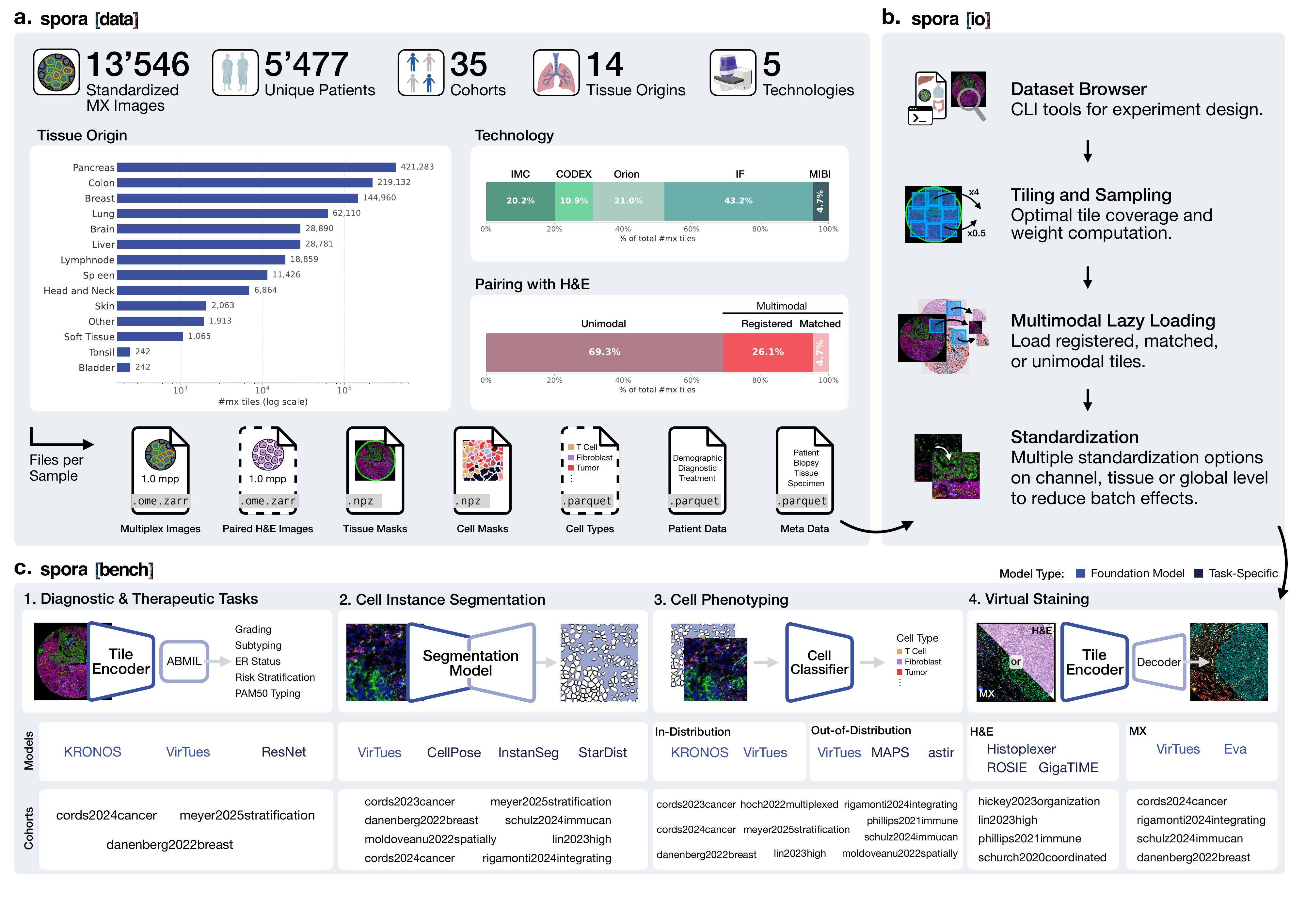}
    \caption{
    \textbf{Overview of \spora.}
    \textbf{a.} Unified dataset composition and per-sample file structure of \sporadata.
    \textbf{b.} Library pipeline of \sporaio from dataset browsing and tiling to standardized model input.
    \textbf{c.} \sporabench benchmark tasks for foundation model evaluation.
    }
    \label{fig:overview}
\end{figure}

In this work, we present the \textbf{S}patial \textbf{P}rote\textbf{o}mics \textbf{R}eference \textbf{A}tlas (\textbf{\spora}), the first unified, multimodal, AI-ready resource for spatial proteomics, and, to our knowledge, the largest publicly available SP corpus assembled to date (\figref{fig:overview}). \spora brings together a large body of publicly released SP cohorts under a common resolution, file layout, and metadata schema, with cell and tissue segmentations, structured clinical annotations, and paired H\&E provided wherever the source studies allow. By unifying SP at this scale, \spora not only enables FM training across panels, technologies, and tissues, but also lays the groundwork for spatial tissue atlases: reference maps that capture how cells are organized within tissues across health and disease. Such atlases would provide reference frames for interpreting new patient samples, expose microenvironmental archetypes that recur across cancers and organs, and document how treatments reshape tissue at single-cell resolution. The accompanying library \sporaio exposes the dataset as an AI-ready interface, and \sporabench evaluates current SP foundation models against task-specific baselines on a set of cell- and tissue-level tasks.
We summarize our main contributions as follows:
\begin{itemize}[leftmargin=*]
    \item \textbf{\sporadata}, a large-scale unified spatial proteomics dataset. We assemble 13,546 standardized SP multiplex (mx) images from 5,477 unique patients across 35 cohorts, 13 tissue origins, and five major SP technologies (IMC, CODEX, Orion, MIBI, IF). All images are resized to a common 1.0 \textmu m/px resolution and stored as \texttt{.ome.zarr} files, accompanied per sample by cell instance and tissue segmentation masks as well as structured patient, clinical, and acquisition metadata. Depending on the cohort, paired H\&E images (as pixel-level registered or matched modalities) and cell-type annotations are additionally provided, enabling multimodal learning at a scale not previously available in SP.
    \item \textbf{\sporaio}, a multimodal data loading library. A Python package built around the \spora dataset with four core capabilities: (i) precomputed tissue-aware tile coverage that improves over standard grid tiling by adapting to irregular tissue geometry; (ii) cross-cohort combination with weighted sampling for flexible experiment design; (iii) multimodal lazy loading directly from \texttt{.ome.zarr} images, jointly streaming paired H\&E and cell segmentation masks; and (iv) standardization strategies at multiple levels (channel, tissue, cohort, and global) to minimize batch effects across technologies and marker panels.
    \item \textbf{\sporabench}, a spatial proteomics benchmark for oncology. A modular benchmark suite spanning cell-level (segmentation and phenotyping) and tissue-level (diagnosis, therapy response, and virtual staining) tasks across cancer types, organs and technologies, evaluated against current SP foundation models and task-specific baselines.
\end{itemize}

Released open-source at \href{https://spora.bioimagearchive.org/}{\texttt{\textcolor{citeblue}{spora.bioimagearchive.org}}}, \spora turns spatial proteomics from a collection of bespoke single-study problems into a shared resource for the ML community to study marker-set heterogeneity, technology shift, multimodal pairing, and multiscale supervision.
\section{\spora Dataset}

\begin{figure}[t] 
  \centering
  \includegraphics[trim=0.0cm 7.5cm 19cm 0.0cm,clip,width=\textwidth, page = 2]{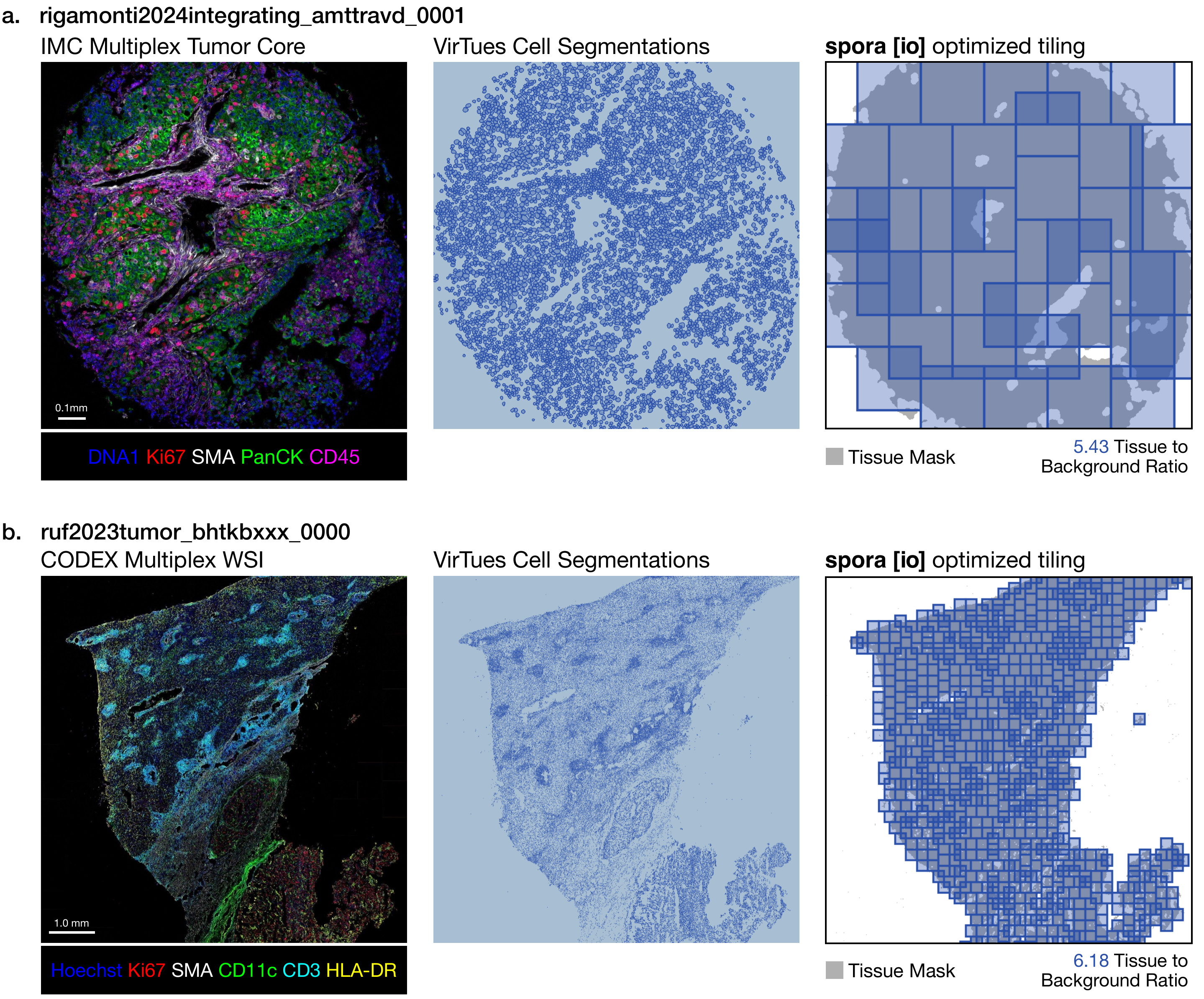}
  \caption{
  \textbf{Exemplary tissues and annotations.} Multiplex images of \textbf{a.} an IMC tumor core \citep{rigamonti2024integrating} and \textbf{b.} a CODEX whole-slide image \citep{ruf2023tumor}, with cell instance segmentations from VirTues and tissue-aware tile distributions from \sporaio. The tiling adapts to tissue geometry, achieving tissue-to-background ratios well above standard grid tiling.
  }
  \label{fig:tissue_examples}
\end{figure}

Building a unified SP dataset is not principally a problem of aggregation but of harmonization: public SP cohorts disagree on file format, marker nomenclature, segmentation strategy, and identifier conventions. 
These disagreements are what have so far prevented cross-cohort training. 
\sporadata resolves them through a set of design choices we describe in this section, applied to a corpus of 13,546 multiplex images from 5,477 patients across 35 cohorts, 13 tissue origins, 295 markers, and five SP technologies (IMC, CODEX, Orion, MIBI, IF; \figref{fig:overview}, \figref{fig:marker_overview}). Each sample carries a multiplex image, tissue and cell instance masks, structured patient data, and metadata; where available, paired H\&E and per-cell phenotype annotations are additionally provided. \tabref{tab:cohort_overview} summarizes the included cohorts.

All cohorts are released under permissive licenses and were obtained through open-access repositories or directly from the original study authors via dataset access requests and correspondence. Over two years, in close collaboration with biomedical experts, they were integrated into a unified, biomedically validated, AI-ready resource. This required substantial preprocessing: folder structures, file formats, metadata schemas, and image representations were harmonized across cohorts, and tissue and cell segmentations were recomputed and improved over those provided in the original releases. Full procedures are detailed in the following paragraphs and in the dataset documentation\footnote{\url{https://spora.epfl.ch/docs/}}.

\paragraph{Meta and patient data.}
\sporadata provides metadata at two levels. 
Every sample in \sporadata is de-identified and pseudonymized, and is assigned a unique tissue identifier composed of a cohort ID, patient ID, and specimen ID.
A multiplex image and a paired H\&E image share the same specimen ID if and only if both were acquired from the exact same tissue slide and are spatially registered. Co-registration between modalities is therefore encoded in the identifier itself, with no separate registration table to maintain. Cohort-level metadata additionally records the source publication, license, and publication date.
Each tissue entry further stores the acquisition modality, possible alignment level to H\&E, original patient and tissue identifiers from the source study (for back-linking to the original results), and a harmonized primary cancer type derived from OncoTree \citep{kundra2021oncotree} for solid tumors and community-standard taxonomies for hematologic malignancies. 
\figref{fig:overview} and \figref{fig:descriptive_stats} report the distribution of tiles and patients across these dimensions.

Clinical annotations are recorded at the \emph{patient} level and fall into three groups (see \figref{fig:descriptive_stats}c). 
\emph{Diagnostic} annotations (available for 20 cohorts) include cancer subtyping, grading, staging, histopathological descriptors, and
cell-state or immune infiltration levels. 
\emph{Therapeutic} annotations (available for 15 cohorts) include administered treatment, treatment response, and overall, progression-free, and disease-free survival. 
\emph{Demographic} annotations (available for 17 cohorts) include sex, age, and smoking status. 
All tissue- and patient-level annotations are stored in a single per-cohort \texttt{.parquet} file, supporting efficient filtering and stratification during experiment design and providing the label infrastructure for the benchmark tasks described in \aref{sec:5}.

\paragraph{Spatial proteomics images.}
\sporadata includes both whole-slide images (WSI) and tumor cores from tissue microarrays, each identified by its tissue ID. 
Because acquisition technologies and source datasets report different resolutions, all images are provided both in their native solution as well as resampled to a common 1.0~\textmu m/px resolution and stored in \texttt{.ome.zarr} \citep{moore2023ome}, the community standard for large multiplex imaging data. 
\texttt{.ome.zarr} not only supports chunked lazy loading optimal for ML model training but also is natively compatible with standard bioimaging tools including QuPath, napari, or Fiji~\citep{bankhead2017qupath,sofroniew2022napari,schindelin2012fiji}.
Each tissue is accompanied by a precomputed binary tissue mask, used downstream to restrict tile sampling to foreground tissue and avoid background regions.
See \aref{app:sec:2} for exemplary visualizations of all technologies.

Per-channel marker information is embedded within each \texttt{.ome.zarr} image and additionally saved as cohort-level \texttt{.parquet} tables for efficient programmatic access. 
These tables record the cohort-consistent marker names and official UniProt identifier where applicable. 
Nuclear stains such as DAPI or Hoechst bind DNA directly rather than a protein target and therefore lack UniProt identifiers; they are flagged with a dedicated binary column, allowing cell segmentation methods to locate the nuclear channel regardless of which specific stain a given cohort used.
A quality-control column further flags channels that did not pass our or the original publications QC, allowing them to be excluded from analyses.
Additional tissue-level tables are used when marker panels vary between patients within a single cohort, ensuring correct marker-channel assignment.

\paragraph{Cell segmentation and phenotyping.}
Every multiplex image in \spora is accompanied by a cell instance segmentation at 1.0~\textmu m/px resolution. 
For each study, we provide standardized cell masks predicted using the VirTues-based segmentation model \citep{wenckstern2026virtual} (\figref{fig:example_mx_imc}--\ref{fig:example_mx_orion}), as well as the cell masks published alongside the source study to preserve the original authors' annotations.
VirTues was selected after systematic comparison with alternative segmentation approaches and consultation with domain experts, as it produced the most consistent results across acquisition technologies and organs; in several cohorts, expert review judged VirTues-based masks to be of higher quality than those released with the original study (\aref{app:sec:virtues_seg_eval}, \figref{fig:virtues_segmentations_cords}-\ref{fig:virtues_segmentations_liu}), motivating our decision to include VirTues segmentations uniformly across all cohorts.

For 16 cohorts, per-cell phenotype annotations are additionally available with granularity matching the source study's original labels (\tabref{tab:cell_types}). 
These typically capture both cell type (e.g., CD15\textsuperscript{+} myeloid cells) and activation state (e.g., MHC-II\textsuperscript{+} myeloid cells), enabling downstream tasks that distinguish phenotypic identity from functional state. 
Phenotype labels are released as panoptic segmentation masks alongside structured \texttt{.parquet} tables keyed on cell-instance identifiers.

Together these components turn 35 disjoint releases into a single queryable resource: every tissue has a globally unique identifier, every marker has a stable name, and every cell and tissue have a segmentation produced by a uniform pipeline. What remains is to make this resource consumable at training scale, which \sporaio addresses next.
\section{\sporaio Library} \label{sec:spora_io}

\newcommand{\Tcal}{\mathcal{T}}
\newcommand{\Ycal}{\mathcal{Y}}
\newcommand{\Xcal}{\mathcal{X}}
\newcommand{\Ccal}{\mathcal{C}}
The value of a large spatial proteomics corpus depends not only on its scale, but also on whether models can access it through a fast and stable interface. 
Consuming the dataset at training scale raises challenges that go beyond data loading: tiling WSI images requires adapting to irregular tissue geometry; combining cohorts of differing sizes calls for principled sampling strategies; jointly streaming SP, paired H\&E, and segmentation masks demands lazy multimodal access to avoid memory bottlenecks; and training across cohorts and technologies requires consistent standardization to minimize batch effects. 
A dedicated IO library is therefore essential to translate \spora's curated data into reproducible, training-scale model development to separate model performance from pre-processing choices.

\sporaio addresses these challenges with a single Python interface that exposes \sporadata as an AI-ready dataset rather than a collection of files. It plays three roles: streaming samples lazily from the canonical on-disk format, applying standardized spatial and intensity preprocessing, and providing the data abstractions on which all benchmark tasks are built.

\paragraph{Data abstractions.} \sporaio is organized around typed dataset objects that expose the same logical operations across modalities while preserving modality-specific metadata. The fundamental unit of access it the tissue identifier, for which H\&E and immunohistochemistry (IHC) loaders return RGB image objects, while multiplex loaders return $C\times H\times W$ tensors with channel names, binary masks capturing which channels are measured and pass quality control, along with UniProt identifiers. Further, the same tissue identifier can be used to retrieve the tissue mask, cell-level masks, and tiles from any modality. Further, a composed dataset class is provided, that builds on the unimodal loaders to provide a multimodal loader for registered or unregistered modalities, enabling benchmarking tasks to switch between unimodal, crossmodal, and multimodal supervision.  
 
\paragraph{Multimodal tiling and sampling.}
Tiling determines the input distribution of every SP model, yet it has received limited methodological attention. The standard approach, inherited from H\&E pipelines, enumerates a fixed grid and discards tiles below a tissue-content threshold. This works for H\&E, where tissues are typically dense gigapixel WSIs, but breaks on SP: tumor cores and resections exhibit irregular morphology, interior voids (necrotic cores, folds), and biologically critical margins where invasive fronts and immune infiltrates concentrate. Grid tiling can discard or contaminate the regions that carry the strongest clinical signal. On the tissues in \figref{fig:tissue_examples}, the \sporaio tiler reaches tissue-to-background ratios of $5.43$ to $6.18$. 

We address this by posing tiling as a budgeted coverage optimization problem.
Let $M \in \{0,1\}^{H \times W}$ be the binary tissue mask and let
$\Ccal_\tau$ be the set of stride-grid candidate tiles whose tissue purity is at
least $1-\tau$. Each candidate $(y,x)\in\Ccal_\tau$
covers a set of valid pixels
$S_{(y,x)} := T(y,x)\cap \{(i,j):M_{ij}=1\}$, and for a budget $B$ we seek
\[
    \max_{\Tcal\subseteq\Ccal_\tau,\ |\Tcal|\le B}
    f(\Tcal)
    =
    \left|
    \bigcup\nolimits_{(y,x)\in\Tcal} S_{(y,x)}
    \right|.
\]
This is a budgeted maximum coverage problem over the purity-feasible candidate set and is NP-hard in general \citep{hochbaum1998analysis}. However, the coverage objective $f$ is monotone and submodular, so for any fixed tile budget $B$, the greedy prefix of length $B$ achieves the classical $(1-1/e)\approx0.63$ approximation guarantee to the optimal $B$-tile coverage~\citep{nemhauser1978analysis}. In practice, we do not require $B$ to be fixed a priori. Instead, the algorithm constructs the greedy sequence and terminates only once the coverage exceeds $\gamma$ and the best remaining marginal gain falls below $\alpha c^2$. The first condition prevents premature stopping on sparse or fragmented tissue, while the second avoids adding mostly redundant tiles once only negligible new tissue remains. Two design choices make the procedure practical at WSI scale, where $|\Ccal_\tau|$ can reach hundreds of thousands. First, all initial candidate gains are computed with summed-area tables. Second, after selecting a tile, we update only spatially overlapping candidates using a local integral image of the uncovered mask, rather than recomputing every marginal gain from scratch. Thus, each greedy step requires a vectorized best-candidate scan plus local updates over the affected candidates. For details on the \spora tiling algorithm, see \aref{app:sec:tiling}.

Tile coordinates are precomputed and released with \spora, and can be deterministically reproduced from $M$, $\tau$, $\gamma$, and $\alpha$, ensuring that benchmark numbers remain comparable across papers.

\paragraph{Multiplex standardization.}
Multiplex intensities are not comparable across cohorts, markers, or acquisition technologies, so any model trained on raw values learns acquisition identity as readily as biology. 
We precompute standardization statistics rather than estimating them inside training code, making preprocessing reproducible across models, machines, and benchmark runs. 
Channel-wise statistics are exposed at the tissue and cohort level, letting users trade batch-effect removal against preservation of biological variation, and two default transforms are provided: quantile-clipped linear normalization for general use, and a log-compressed variant for the heavy-tailed marker distributions common in multiplex proteomics. 
Formulas, validity-mask handling, and routines for computing statistics on new datasets are given in \aref{app:sec:standardization}.

\section{\spora Benchmark} \label{sec:5}

The spatial proteomics benchmark, \sporabench, is based on the images and associated cell- and tissue-level metadata provided by \sporadata as well as the \sporaio interface. It spans the field's core analysis tasks: (i) \emph{cell segmentation}, (ii) \emph{cell phenotyping} (e.g., assigning cell types and functional states), (iii) \emph{diagnostic and therapeutic prediction} (e.g., risk stratification, cancer subtyping), and (iv) \emph{virtual staining} (i.e., predicting unmeasured markers from observed measurements). We provide a modular implementation of these tasks consisting of an abstract model wrapping class with pre-defined method signatures, for each FM an implementation of this abstract class, and six evaluation pipelines that then run each wrapped model and collect task-level benchmarking results.

In total, \sporabench contains eight segmentation, 18 cell phenotyping, six diagnostic and therapeutic prediction and eight virtual staining tasks, and is designed to effortlessly integrate new tasks on both existing and new patient cohorts. 

\paragraph{Guidelines on data splits and exposure.}
For all datasets in \sporadata, we provide fixed 80/20 train--test splits at the patient level, enabling fair and reproducible comparisons and preventing models from exploiting patient-specific batch effects. Validation data, where needed, should be carved out from the training split. These splits must be respected during benchmark finetuning and, where possible, during pretraining. For public models the latter is often not verifiable, and we therefore recommend explicit reporting when such models are compared. Even when splits are followed strictly, pretraining exposure, i.e., whether a model has encountered a cohort during pretraining, may confound comparisons. Because the large SP cohorts most valuable for finetuning are also the most valuable for pretraining, and remain scarce, we ask users to disclose pretraining exposure rather than excluding benchmark datasets altogether. For \sporabench, \tabref{tab:disclosure_pretraining_exposure} reports pretraining exposure for each cohort and foundation model whose developers disclose it publicly.

\paragraph{Cell phenotyping.} We distinguish two cell phenotyping settings, each implemented by its own pipeline: (i) \emph{linear-probe cell classification} and (ii) \emph{cross-cohort cell annotation}. The former assesses the biological information content of the cell-level representations produced by an FM by fitting a logistic regression probe within a single cohort. The latter targets the practical setting of annotating a newly acquired cohort for which no labels are available, and therefore additionally measures how well methods generalize across cohorts and panels.

\begin{table}[t]
\centering
\caption{\textbf{Linear-probe cell phenotyping results.} Macro average F1-score per dataset and model. Best performing model per dataset is shown in bold. The 95\% confidence intervals calculated via bootstrapping ($n=1000$ repeats) are shown in brackets. \vspace{0.2cm}}
\label{tab:all_datasets_macro_f1}
\begin{tabular}{lll}
\toprule
Model & KRONOS & VirTues \\
Dataset &  &  \\
\midrule
\citet{cords2023cancer} & 0.601 {\footnotesize [.597, .605]} & \textbf{0.782} {\footnotesize [.778, .785]} \\
\citet{cords2024cancer} & - & \textbf{0.757} {\footnotesize [.755, .758]} \\
\citet{danenberg2022breast} & 0.355 {\footnotesize [.352, .358]} & \textbf{0.611} {\footnotesize [.607, .615]} \\
\citet{hoch2022multiplexed} & 0.574 {\footnotesize [.566, .582]} & \textbf{0.836} {\footnotesize [.830, .841]} \\
\citet{lin2023high} & 0.400 {\footnotesize [.397, .402]} & \textbf{0.642} {\footnotesize [.638, .645]} \\
\citet{meyer2025stratification} & 0.480 {\footnotesize [.476, .484]} & \textbf{0.718} {\footnotesize [.714, .722]} \\
\citet{moldoveanu2022spatially} & 0.523 {\footnotesize [.507, .538]} & \textbf{0.716} {\footnotesize [.699, .733]} \\
\citet{phillips2021immune} & 0.437 {\footnotesize [.419, .454]} & \textbf{0.715} {\footnotesize [.701, .729]} \\
\citet{rigamonti2024integrating} & 0.577 {\footnotesize [.569, .585]} & \textbf{0.699} {\footnotesize [.690, .708]} \\
\citet{eling2025multi} & 0.546 {\footnotesize [.539, .553]} & \textbf{0.761} {\footnotesize [.756, .768]} \\
\bottomrule
\end{tabular}
\end{table}

For the linear-probe setting, we select ten cohorts \citep{cords2024cancer, cords2023cancer, danenberg2022breast, hoch2022multiplexed, lin2023high, meyer2025stratification, moldoveanu2022spatially, phillips2021immune, rigamonti2024integrating, eling2025multi} and use the phenotype annotations provided by the original studies. These span a wide range of granularity, from four coarse classes \citep{cords2023cancer} to fourteen fine-grained ones \citep{eling2025multi}, and comprise 12.4 million cells in total. For the cross-cohort annotation setting, we select eight of these cohorts \citep{cords2023cancer, cords2024cancer, danenberg2022breast, meyer2025stratification, lin2023high, moldoveanu2022spatially, rigamonti2024integrating, eling2025multi} and harmonize their annotations into seven common classes.

We benchmark two FMs for spatial proteomics on the linear-probe tasks, VirTues~\citep{wenckstern2026virtual} and KRONOS~\citep{shaban2025foundation}. To this aim, we compute cell-level representations following each model's prescribed procedure, normalize them feature-wise, and train and evaluate a logistic regression classifier on the respective data split. 
On the cross-cohort cell annotation tasks, we evaluate VirTues under its semantic-segmentation-based annotation protocol, together with two classical specialized tools, astir \citep{geuenich2021automated} for unsupervised cell annotation, and MAPS \citep{shaban2024maps} for cell label transfer. KRONOS is omitted here, as it provides no method for unsupervised cell annotation or label transfer across varying panels.

Macro-averaged F1 scores across all cohorts and models are reported in \tabref{tab:all_datasets_macro_f1} for the linear-probe tasks and in \tabref{tab:cross_cohort_cell_annotation_macro_f1} for the cross-cohort annotation tasks. Per-cohort and per-cell-type results are provided in \aref{app:sec:spora_benchmark} (\tabref{tab:cords2023cancer_cell_types}--\tabref{tab:schulz2024immucan_cell_types}). On the linear-probe tasks, VirTues achieves the highest score in every cohort, across all levels of cell type granularity. On the cross-cohort annotation tasks, VirTues is again the most consistent performing method, ranking first on four of the eight cohorts and second on three.

\paragraph{Cell instance segmentation.} Segmenting cell instances is a key step in nearly every SP analysis requiring understanding of cell shapes across varying morphology markers. We benchmark this task across eight cohorts \citep{cords2023cancer, cords2024cancer, danenberg2022breast, meyer2025stratification, lin2023high, moldoveanu2022spatially, rigamonti2024integrating, eling2025multi}, using the instance segmentation masks provided by the original publications as ground truth. We evaluate the FM VirTues \citep{wenckstern2026virtual} alongside three established specialized segmentation tools: Cellpose \citep{pachitariu2025cellpose}, InstanSeg \citep{goldsborough2024instanseg}, and StarDist \citep{weigert2022nuclei}. To reflect the real-world setting of segmenting a newly imaged cohort and to test the generalization ability of methods across cohorts, the task is evaluated zero-shot, i.e., none of the methods is trained on any segmentation masks from the target cohort.

We report results in terms of instance F1-scores at intersection-over-union (IoU) threshold $\tau = 0.5$ in \tabref{tab:segmentation_f1_tau_05}. VirTues performs best overall, leading on six of the eight cohorts, often by a substantial margin. Cellpose is the strongest of the remaining methods and accounts for the other two cohorts.

\begin{figure}[t] 
  \centering
  \includegraphics[trim=0cm 29.5cm 3.6cm 0cm,clip,width=\textwidth, page = 3]{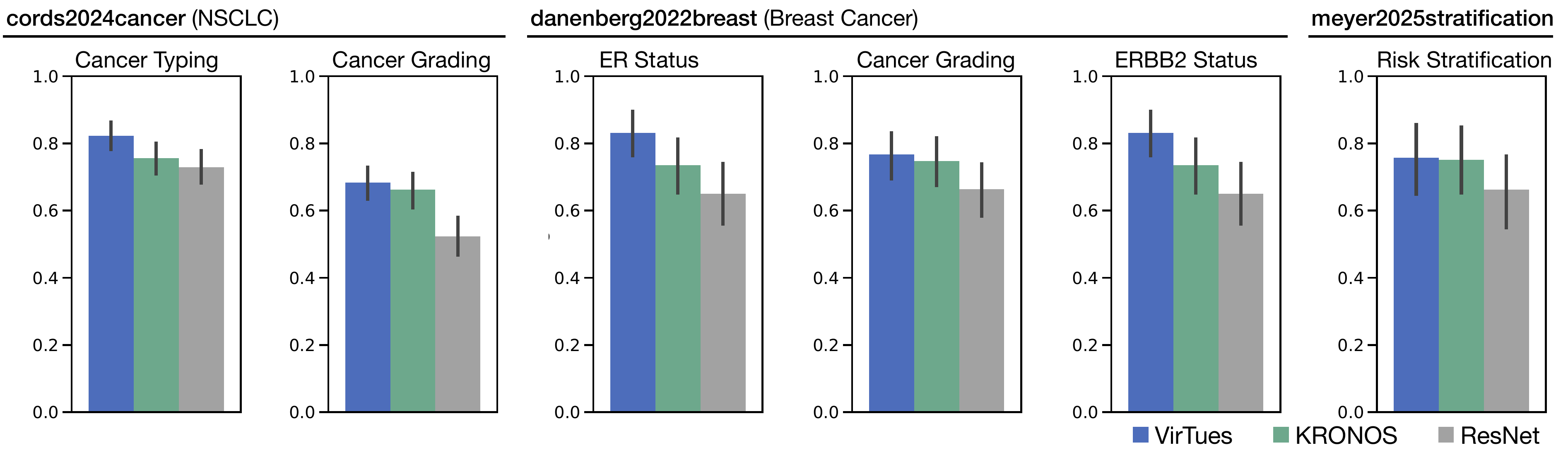}
  \caption{
  \textbf{Tissue-level diagnostic and therapeutic results.} Performance across diagnostic and therapeutic tasks on three cohorts spanning non-small cell lung cancer (NSCLC), triple-negative breast cancer (TNBC), and a heterogeneous breast cancer cohort. Each task contains two classes.
  }
  \vspace{-0.4cm}
  \label{fig:benchmark_tissue_level}
\end{figure}

\paragraph{Diagnostic and therapeutic prediction.} 
To assess the extent to which SP models capture tissue-level signal, we include six clinically relevant pathology tasks spanning cancer grading and subtyping across a non-small cell lung cancer cohort \citep{cords2024cancer}, a diverse breast cancer cohort \citep{danenberg2022breast}, and a triple-negative breast cancer cohort \citep{meyer2025stratification}, comprising 3,284 tissues from 2,002 patients in total.
As in the cell phenotyping tasks, we use the unified wrapper to evaluate the two SP foundation models alongside a ResNet feature encoder \citep{sorin2023single}, a traditional baseline for tissue-level SP prediction \citep{klein2026explainable}.

We follow the tissue-level evaluation protocol of VirTues and KRONOS, iteratively embedding tissue tiles in a grid-crop manner to extract patch or tile representations respectively. The resulting set of embeddings for each whole tissue is aggregated through attention-based multiple instance learning (ABMIL) \citep{ilse2018attention}, the standard framework for tissue-level prediction, to train a downstream classifier. For the ResNet baseline, we substitute the patch representations with tile-level features extracted by the encoder.
\figref{fig:benchmark_tissue_level} reports performance across all six tasks. VirTues outperforms KRONOS on four tasks and matches it on patient risk stratification and breast cancer grading.

We observe that on \sporabench, foundation models outperform classical baselines across cell-, tissue-, and patient-level tasks. Yet performance is far from saturated, with substantial headroom remaining across all tasks. One likely reason is that pretraining data has so far been limited in both scale and standardization, a constraint that \spora addresses directly. The rest of the gap is harder to attribute and likely reflects architectural and aggregation choices, which lie outside what a dataset alone can solve. As \spora grows, the data-side of this gap should narrow further, though closing it fully will likely also require advances in architecture and aggregation.

\begin{figure}[t] 
  \centering
  \includegraphics[trim=0.0cm 13.5cm 15cm 0.0cm,clip,width=0.97\textwidth, page = 14]{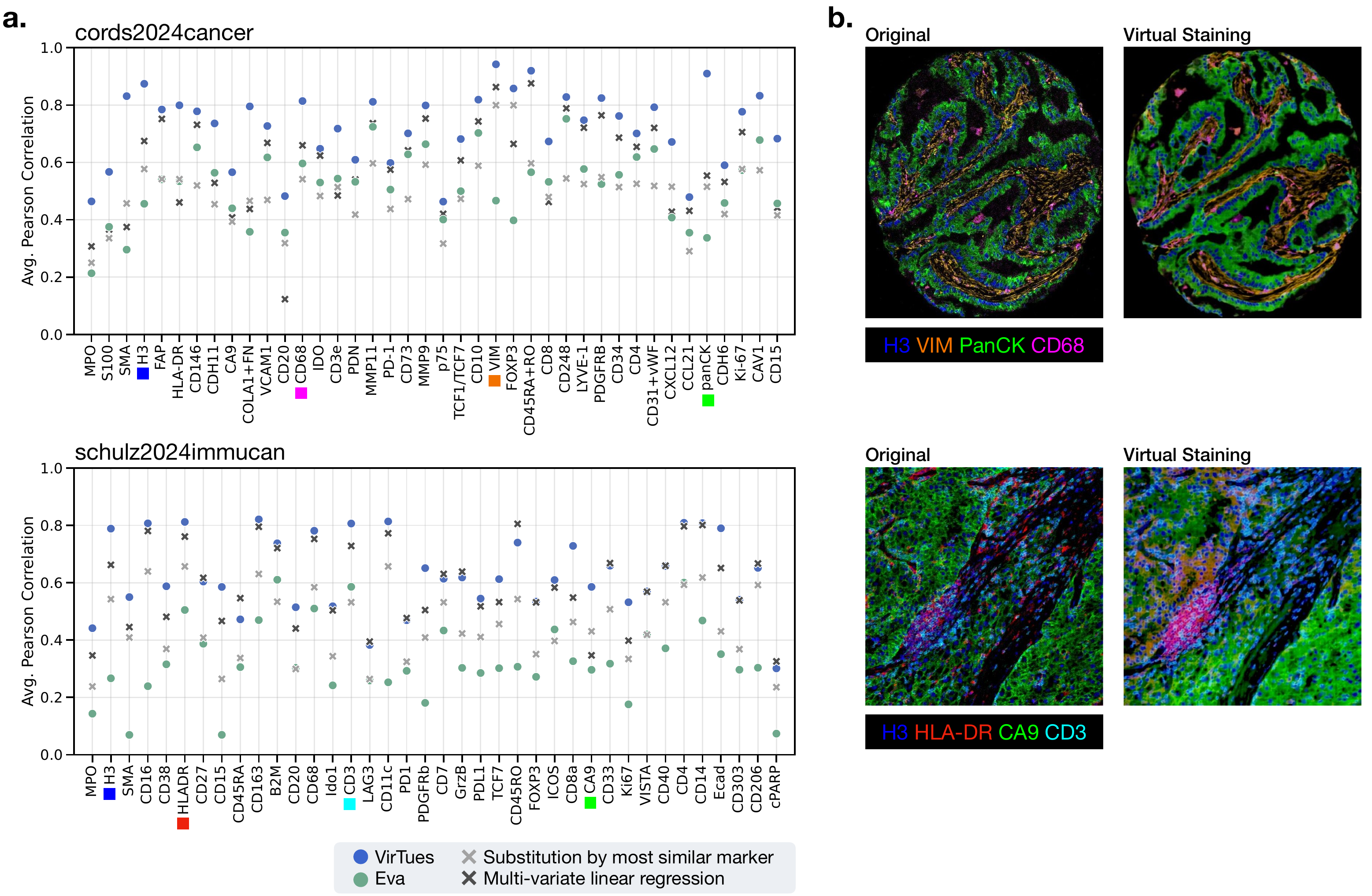}
  \caption{
    \textbf{Multiplex virtual staining results.}
    \textbf{a.} Average Pearson correlation between predicted and ground-truth markers across tissues; the baseline (light-gray cross) substitutes the closest available marker for the target.
    \textbf{b.} Representative multiplex images and their virtual staining predictions obtained from VirTues.
  }
  \label{fig:virtual_staining}
\end{figure}

\vspace{-5pt}

\paragraph{Virtual staining.}
Multiplexed SP resolves tissue composition with molecular specificity, but cost, turnaround time, and finite tissue limit its use, forcing investigators toward smaller panels or applying the assay to a subset of samples. Virtual staining offers a way to relax these constraints: given H\&E morphology or a reduced marker panel, a foundation model can impute unmeasured channels \emph{in silico}, extending an assay's effective size without additional tissue. 
Beyond its practical value, this task probes what a model has learned. 
Imputation requires exploiting dependencies among markers and between molecular signal and morphology, so performance measures how well spatial and cross-modal structure has been captured, not only how useful the output is.

\sporabench introduces two standardized virtual staining benchmarks: (i) \emph{multiplex-to-multiplex virtual staining}, and (ii) \emph{H\&E-to-multiplex virtual staining}. The benchmarks can be evaluated both in-distribution, using models trained on the datasets’ training splits, and out-of-distribution, by withholding the entire dataset from all model training.
We compare below methods evaluated in-distribution. Importantly, performance numbers from different evaluation settings are not directly comparable, and should be reported with the setting made explicit.

Multiplex-to-multiplex virtual staining benchmarks are provided for four cohorts \citep{cords2024cancer,eling2025multi,danenberg2022breast,rigamonti2024integrating}. Here, for each tissue, we iteratively hold out one channel at a time and predict it from the remaining channels. We report the per-marker Pearson correlation between predicted and ground-truth signal, averaged over all test tissues (\figref{fig:virtual_staining}, \figref{app:fig:virtual_staining_2}).
We compare two foundation models, VirTues and Eva \citep{liu2025modeling}, against two simpler references: a linear model fit per target marker on a pixel-wise basis, and a nearest-marker baseline that imputes for each marker the single most similar alternative marker.
Virtual stains predicted by VirTues correlate consistently well with the ground truth and outperform all other methods. Performance nevertheless varies substantially across markers and correlates with the nearest-marker baseline, suggesting some markers carry redundant information while others encode unique signal.

Likewise, we provide H\&E-to-multiplex virtual staining benchmarks for four cohorts \citep{lin2023high,hickey2023organization,phillips2021immune,schurch2020coordinated}. Leveraging the availability of registered modalities, the multiplexed marker panel is here predicted directly from an H\&E slide, which is fast and inexpensive to acquire. We benchmark three virtual staining models, GigaTIME \citep{valanarasu2025multimodal}, HistoPlexer \citep{andani2025histopathology} and ROSIE \citep{wu2025rosie}. All three were trained on closed-source and comparatively small corpora (21 WSIs, 336 ROIs and 1,342 tumor cores, respectively), all but ROSIE on a single cohort, and each predicts a fixed panel that overlaps only partly with the others. Accordingly, their performance on our four evaluation cohorts is limited (\tabref{tab:virtual_staining_he}, \tabref{app:tab:vs_per_marker_hickey2023organization}--\ref{app:tab:vs_per_marker_schurch2020coordinated}), with correlations near zero for most markers. 
This gap is what cross-cohort, cross-platform training and evaluation is meant to close, and what \spora is built to measure. 

\begin{table}[t]
\centering
\caption{\textbf{H\&E virtual staining results.} Pearson correlation and SSIM averaged over the markers predicted by all three models in a given cohort. Predicted marker set differs between models, indicated by \# markers in the order GigaTIME/HistoPlexer/ROSIE. Best performing model per dataset and metric is shown in bold.}\label{tab:virtual_staining_he}
\vspace{0.2cm}
\small
\begin{tabular}{lcSSSSSS}
\toprule
 & & \multicolumn{2}{c}{GigaTIME} & \multicolumn{2}{c}{HistoPlexer} & \multicolumn{2}{c}{ROSIE} \\
\cmidrule(lr){3-4} \cmidrule(lr){5-6} \cmidrule(lr){7-8}
Dataset & \# markers & {Pearson $\uparrow$} & {SSIM $\uparrow$} & {Pearson $\uparrow$} & {SSIM $\uparrow$} & {Pearson $\uparrow$} & {SSIM $\uparrow$} \\
\midrule
\citet{hickey2023organization} & 10/5/23 & {\bfseries 0.056} & 0.384 & 0.004 & {\bfseries 0.603} & 0.043 & 0.082 \\
\citet{lin2023high} & 10/3/15 & 0.046 & 0.266 & 0.004 & 0.056 & {\bfseries 0.066} & {\bfseries 0.301} \\
\citet{phillips2021immune} & 15/6/29 & {\bfseries 0.106} & 0.294 & 0.003 & {\bfseries 0.576} & 0.060 & 0.076 \\
\citet{schurch2020coordinated} & 14/7/33 & {\bfseries 0.072} & 0.245 & 0.004 & {\bfseries 0.408} & 0.038 & 0.098 \\
\bottomrule
\end{tabular}
\end{table}
\section{Conclusion}
\looseness -1 \spora is the first spatial proteomics resource large and standardized enough to support foundation model development across cohorts, panels, and tissue types. It combines harmonized data from 35 studies, a multiplex IO library, and a multiscale benchmark spanning cell-, tissue-, and patient-level tasks. Together, these components enable large-scale pretraining and systematic evaluation across heterogeneous biological contexts. They lay the data and computational foundations for spatial proteomics atlases in which evidence can accumulate across studies rather than remain confined to individual datasets. Like the public data it draws on, however, \spora is skewed towards cancer and covers only a subset of cancer types and organs, reflecting where the field has concentrated its efforts so far. \sporabench inherits these limitations, and the scope of our model evaluation is constrained by how few spatial proteomics foundation models are currently available. As continued data acquisition and community contributions broaden its coverage, \spora will support increasingly generalizable models that reveal how molecular and cellular organization shapes tissue function and disease.

\subsubsection*{Acknowledgments}
\vspace{5pt}
\begin{tcolorbox}[colback=boxgray, colframe=boxgray, boxrule=0pt, sharp corners]
\vspace{5pt}
\spora is officially hosted by EMBL-EBI. We are deeply grateful to the BioImage Archive team, in particular Matthew Hartley, Aybuke Kupcu Yoldas, Teresa Zulueta Coarasa, Christina Ernst, Tushar Mandloi, and Daniel Marks, for integrating \spora as an EMBL-EBI database. Their expertise in managing large-scale imaging data and their close and ongoing collaboration throughout the process were essential to establishing \spora as a sustainable, openly accessible resource. We hope it will continue to grow as a community resource, with more researchers contributing their own datasets over time.
\vspace{5pt}
\end{tcolorbox}
\vspace{5pt}

We thank the Research Computing Platform (RCP) at EPFL, especially Khadidja Malleck and Yoann Moulin, as well as Edward Andò and Florian Aymanns at the EPFL Center for Imaging for storing and visualizing \spora.
We further acknowledge support from John W. Hickey, Garry Nolan, Federica Marchesi, Marika Viatore, James Monkman, Lasse Meyer, Christian Sch\"urch, Jia-Ren Lin, Raza Ali, Magdalena Matusiak, and Gabriele Gut with accessing and analyzing the datasets.

This work was supported by the Swiss AI Initiative through a Swiss National Supercomputing Centre (CSCS) allocation under the Swiss AI Large Grants \texttt{a146} and \texttt{ab007} (“Virtual Patient Platform”), an AI2050 Early Career Fellowship, the EPFL AI Center PhD Fellowship program, and by the ISREC Foundation. 
\autoref{fig:overview} created in BioRender: Klein, L. (2026) \url{https://BioRender.com/j07qq2t}.

\bibliography{references}

\begin{thebibliography}{87}
\providecommand{\natexlab}[1]{#1}
\providecommand{\url}[1]{\texttt{#1}}
\expandafter\ifx\csname urlstyle\endcsname\relax
  \providecommand{\doi}[1]{doi: #1}\else
  \providecommand{\doi}{doi: \begingroup \urlstyle{rm}\Url}\fi

\bibitem[Giesen et~al.(2014)Giesen, Wang, Schapiro, Zivanovic, Jacobs, Hattendorf, Sch{\"u}ffler, Grolimund, Buhmann, Brandt, Varga, Wild, G{\"u}nther, and Bodenmiller]{giesen2014highly}
Charlotte Giesen, Hao A.~O. Wang, Denis Schapiro, Nevena Zivanovic, Andrea Jacobs, Bodo Hattendorf, Peter~J. Sch{\"u}ffler, Daniel Grolimund, Joachim~M. Buhmann, Simone Brandt, Zsuzsanna Varga, Peter~J. Wild, Detlef G{\"u}nther, and Bernd Bodenmiller.
\newblock {Highly multiplexed imaging of tumor tissues with subcellular resolution by mass cytometry}.
\newblock \emph{Nature Methods}, 11\penalty0 (4):\penalty0 417--422, 2014.

\bibitem[Black et~al.(2021)Black, Phillips, Hickey, Kennedy-Darling, Venkataraaman, Samusik, Goltsev, Sch{\"u}rch, and Nolan]{black2021codex}
Sarah Black, Darci Phillips, John~W. Hickey, Julia Kennedy-Darling, Vishal~G. Venkataraaman, Nikolay Samusik, Yury Goltsev, Christian~M. Sch{\"u}rch, and Garry~P. Nolan.
\newblock {CODEX multiplexed tissue imaging with DNA-conjugated antibodies}.
\newblock \emph{Nature Protocols}, 16\penalty0 (8):\penalty0 3802--3835, 2021.

\bibitem[Keren et~al.(2019)Keren, Bosse, Thompson, Risom, Vijayaragavan, McCaffrey, Marquez, Angoshtari, Greenwald, Fienberg, Wang, Kambham, Kirkwood, Nolan, Montine, Galli, West, Bendall, and Angelo]{keren_mibi-tof_2019}
Leeat Keren, Marc Bosse, Steve Thompson, Tyler Risom, Kausalia Vijayaragavan, Erin McCaffrey, Diana Marquez, Roshan Angoshtari, Noah~F. Greenwald, Harris Fienberg, Jennifer Wang, Neeraja Kambham, David Kirkwood, Garry Nolan, Thomas~J. Montine, Stephen~J. Galli, Robert West, Sean~C. Bendall, and Michael Angelo.
\newblock {MIBI-TOF: A multiplexed imaging platform relates cellular phenotypes and tissue structure}.
\newblock \emph{Science Advances}, 5\penalty0 (10):\penalty0 eaax5851, 2019.

\bibitem[Lin et~al.(2023{\natexlab{a}})Lin, Chen, Campton, Cooper, Coy, Yapp, Tefft, McCarty, Ligon, Rodig, et~al.]{lin2023high}
Jia-Ren Lin, Yu-An Chen, Daniel Campton, Jeremy Cooper, Shannon Coy, Clarence Yapp, Juliann~B. Tefft, Erin McCarty, Keith~L. Ligon, Scott~J. Rodig, et~al.
\newblock {High-plex immunofluorescence imaging and traditional histology of the same tissue section for discovering image-based biomarkers}.
\newblock \emph{Nature Cancer}, 4\penalty0 (7):\penalty0 1036--1052, 2023{\natexlab{a}}.

\bibitem[Wu et~al.(2024)Wu, Tao, Xu, Zheng, Wen, Wang, Peng, and Dai]{wu2024spatial}
Mengyao Wu, Huihui Tao, Tiantian Xu, Xuejia Zheng, Chunmei Wen, Guoying Wang, Yali Peng, and Yong Dai.
\newblock {Spatial proteomics: unveiling the multidimensional landscape of protein localization in human diseases}.
\newblock \emph{Proteome Science}, 22\penalty0 (1):\penalty0 7, 2024.

\bibitem[{Nature Eds.}(2024)]{karimi2024method}
{Nature Eds.}
\newblock {Method of the Year 2024: spatial proteomics}.
\newblock \emph{Nature Methods}, 21\penalty0 (12):\penalty0 2195--2196, 2024.

\bibitem[Cords et~al.(2023)Cords, Tietscher, Anzeneder, Langwieder, Rees, de~Souza, and Bodenmiller]{cords2023cancer}
Lena Cords, Sandra Tietscher, Tobias Anzeneder, Claus Langwieder, Martin Rees, Natalie de~Souza, and Bernd Bodenmiller.
\newblock {Cancer-associated fibroblast classification in single-cell and spatial proteomics data}.
\newblock \emph{Nature Communications}, 14\penalty0 (1):\penalty0 4294, 2023.

\bibitem[Sali{\'e} et~al.(2025)Sali{\'e}, Wischer, D'Alessio, Godbole, Suo, Otto-Mora, Beck, Neumann, Stenzinger, Schirmacher, et~al.]{salie2025spatial}
Henrike Sali{\'e}, Lara Wischer, Antonio D'Alessio, Ira Godbole, Yuan Suo, Patricia Otto-Mora, Juergen Beck, Olaf Neumann, Albrecht Stenzinger, Peter Schirmacher, et~al.
\newblock {Spatial single-cell profiling and neighbourhood analysis reveal the determinants of immune architecture connected to checkpoint inhibitor therapy outcome in hepatocellular carcinoma}.
\newblock \emph{Gut}, 74\penalty0 (3):\penalty0 451--466, 2025.

\bibitem[Xiao et~al.(2022)Xiao, Guo, Cui, Lin, Zhang, Ding, Li, Wang, Yang, Kong, et~al.]{xiao2022multiplexed}
Xu~Xiao, Qian Guo, Chuanliang Cui, Yating Lin, Lei Zhang, Xin Ding, Qiyuan Li, Minshu Wang, Wenxian Yang, Yan Kong, et~al.
\newblock {Multiplexed imaging mass cytometry reveals distinct tumor-immune microenvironments linked to immunotherapy responses in melanoma}.
\newblock \emph{Communications Medicine}, 2\penalty0 (1):\penalty0 131, 2022.

\bibitem[Quail and Walsh(2024)]{quail2024revolutionizing}
Daniela~F. Quail and Logan~A. Walsh.
\newblock {Revolutionizing cancer research with spatial proteomics and visual intelligence}.
\newblock \emph{Nature Methods}, 21\penalty0 (12):\penalty0 2216--2219, 2024.

\bibitem[Liu et~al.(2026)Liu, Dai, and Wang]{liu2025spatial}
Yunhe Liu, Yibo Dai, and Linghua Wang.
\newblock {Spatial omics at the forefront: emerging technologies, analytical innovations, and clinical applications}.
\newblock \emph{Cancer Cell}, 44\penalty0 (1):\penalty0 24--49, 2026.

\bibitem[Bommasani et~al.(2021)Bommasani, Hudson, Adeli, Altman, Arora, von Arx, Bernstein, Bohg, Bosselut, Brunskill, Brynjolfsson, Buch, Card, Castellon, Chatterji, Chen, Creel, Davis, Demszky, Donahue, Doumbouya, Durmus, Ermon, Etchemendy, Ethayarajh, Fei-Fei, Finn, Gale, Gillespie, Goel, Goodman, Grossman, Guha, Hashimoto, Henderson, Hewitt, Ho, Hong, Hsu, Huang, Icard, Jain, Jurafsky, Kalluri, Karamcheti, Keeling, Khani, Khattab, Koh, Krass, Krishna, Kuditipudi, et~al.]{bommasani2021opportunities}
Rishi Bommasani, Drew~A. Hudson, Ehsan Adeli, Russ~B. Altman, Simran Arora, Sydney von Arx, Michael~S. Bernstein, Jeannette Bohg, Antoine Bosselut, Emma Brunskill, Erik Brynjolfsson, Shyamal Buch, Dallas Card, Rodrigo Castellon, Niladri~S. Chatterji, Annie~S. Chen, Kathleen Creel, Jared~Quincy Davis, Dorottya Demszky, Chris Donahue, Moussa Doumbouya, Esin Durmus, Stefano Ermon, John Etchemendy, Kawin Ethayarajh, Li~Fei-Fei, Chelsea Finn, Trevor Gale, Lauren Gillespie, Karan Goel, Noah~D. Goodman, Shelby Grossman, Neel Guha, Tatsunori Hashimoto, Peter Henderson, John Hewitt, Daniel~E. Ho, Jenny Hong, Kyle Hsu, Jing Huang, Thomas Icard, Saahil Jain, Dan Jurafsky, Pratyusha Kalluri, Siddharth Karamcheti, Geoff Keeling, Fereshte Khani, Omar Khattab, Pang~Wei Koh, Mark~S. Krass, Ranjay Krishna, Rohith Kuditipudi, et~al.
\newblock {On the Opportunities and Risks of Foundation Models}.
\newblock \emph{arXiv preprint arXiv:2108.07258}, 2021.

\bibitem[Campanella et~al.(2025)Campanella, Chen, Singh, Verma, Muehlstedt, Zeng, Stock, Croken, Veremis, Elmas, et~al.]{campanella2025clinical}
Gabriele Campanella, Shengjia Chen, Manbir Singh, Ruchika Verma, Silke Muehlstedt, Jennifer Zeng, Aryeh Stock, Matt Croken, Brandon Veremis, Abdulkadir Elmas, et~al.
\newblock {A clinical benchmark of public self-supervised pathology foundation models}.
\newblock \emph{Nature Communications}, 16\penalty0 (1):\penalty0 3640, 2025.

\bibitem[Tomczak et~al.(2015)Tomczak, Czerwi{\'n}ska, and Wiznerowicz]{tomczak2015review}
Katarzyna Tomczak, Patrycja Czerwi{\'n}ska, and Maciej Wiznerowicz.
\newblock {The Cancer Genome Atlas (TCGA): an immeasurable source of knowledge}.
\newblock \emph{Contemporary Oncology/Wsp{\'o}{\l}czesna Onkologia}, 19\penalty0 (1A):\penalty0 A68--A77, 2015.

\bibitem[Nechaev et~al.(2025)Nechaev, Pchelnikov, and Ivanova]{nechaev2025histaiopensourcelargescaleslide}
Dmitry Nechaev, Alexey Pchelnikov, and Ekaterina Ivanova.
\newblock {HISTAI: An Open-Source, Large-Scale Whole Slide Image Dataset for Computational Pathology}.
\newblock \emph{arXiv preprint arXiv:2505.12120}, 2025.

\bibitem[Chen et~al.(2024)Chen, Ding, Lu, Williamson, Jaume, Song, Chen, Zhang, Shao, Shaban, Williams, Oldenburg, Weishaupt, Wang, Vaidya, Le, Gerber, Sahai, Williams, and Mahmood]{chen2024uni}
Richard~J. Chen, Tong Ding, Ming~Y. Lu, Drew F.~K. Williamson, Guillaume Jaume, Andrew~H. Song, Bowen Chen, Andrew Zhang, Daniel Shao, Muhammad Shaban, Mane Williams, Lukas Oldenburg, Luca~L. Weishaupt, Judy~J. Wang, Anurag Vaidya, Long~Phi Le, Georg Gerber, Sharifa Sahai, Walt Williams, and Faisal Mahmood.
\newblock {Towards a general-purpose foundation model for computational pathology}.
\newblock \emph{Nature Medicine}, 30\penalty0 (3):\penalty0 850--862, 2024.

\bibitem[Zimmermann et~al.(2024)Zimmermann, Vorontsov, Viret, Casson, Zelechowski, Shaikovski, Tenenholtz, Hall, Klimstra, Yousfi, et~al.]{zimmermann2024virchow2}
Eric Zimmermann, Eugene Vorontsov, Julian Viret, Adam Casson, Michal Zelechowski, George Shaikovski, Neil Tenenholtz, James Hall, David Klimstra, Razik Yousfi, et~al.
\newblock {Virchow2: Scaling Self-Supervised Mixed Magnification Models in Pathology}.
\newblock \emph{arXiv preprint arXiv:2408.00738}, 2024.

\bibitem[Saillard et~al.(2024)Saillard, Jenatton, Llinares-L{\'o}pez, Mariet, Cahan{\'e}, Durand, and Vert]{hoptimus0}
Charlie Saillard, Rodolphe Jenatton, Felipe Llinares-L{\'o}pez, Zelda Mariet, David Cahan{\'e}, Eric Durand, and Jean-Philippe Vert.
\newblock {H-optimus-0}, 2024.
\newblock URL \url{https://github.com/bioptimus/releases/tree/main/models/h-optimus/v0}.

\bibitem[Nechaev et~al.(2024)Nechaev, Pchelnikov, and Ivanova]{nechaev2024hibou}
Dmitry Nechaev, Alexey Pchelnikov, and Ekaterina Ivanova.
\newblock {Hibou: A Family of Foundational Vision Transformers for Pathology}.
\newblock \emph{arXiv preprint arXiv:2406.05074}, 2024.

\bibitem[Jaume et~al.(2024)Jaume, Doucet, Song, Lu, Almagro-Perez, Wagner, Vaidya, Chen, Williamson, Kim, and Mahmood]{jaume2024hest}
Guillaume Jaume, Paul Doucet, Andrew~H. Song, Ming~Y. Lu, Cristina Almagro-Perez, Sophia~J. Wagner, Anurag~J. Vaidya, Richard~J. Chen, Drew F.~K. Williamson, Ahrong Kim, and Faisal Mahmood.
\newblock {HEST-1k: A Dataset for Spatial Transcriptomics and Histology Image Analysis}.
\newblock In \emph{Advances in Neural Information Processing Systems (NeurIPS)}, volume~37, 2024.

\bibitem[Xu et~al.(2024{\natexlab{a}})Xu, Wang, Yang, Li, Ma, Chen, Wang, Huang, Gould, Lu, et~al.]{xu2024stomicsdb}
Zhicheng Xu, Weiwen Wang, Tao Yang, Ling Li, Xizheng Ma, Jing Chen, Jieyu Wang, Yan Huang, Joshua Gould, Huifang Lu, et~al.
\newblock {STOmicsDB: a comprehensive database for spatial transcriptomics data sharing, analysis and visualization}.
\newblock \emph{Nucleic Acids Research}, 52\penalty0 (D1):\penalty0 D1053--D1061, 2024{\natexlab{a}}.

\bibitem[{CZI Cell Science Program} et~al.(2025){CZI Cell Science Program}, Abdulla, Aevermann, Assis, Badajoz, Bell, Bezzi, Cakir, Chaffer, Chambers, Cherry, Chi, Chien, Dorman, Garcia-Nieto, Gloria, Hastie, Hegeman, Hilton, Huang, Infeld, Istrate, Jelic, Katsuya, Kim, Liang, Lin, Lombardo, Marshall, Martin, McDade, Megill, Patel, Predeus, Raymor, Robatmili, Rogers, Rutherford, Sadgat, Shin, Small, Smith, Sridharan, Tarashansky, Tavares, Thomas, Tolopko, Urisko, Yan, Yeretssian, Zamanian, Mani, Cool, and Carr]{program_cz_2025}
{CZI Cell Science Program}, Shibla Abdulla, Brian Aevermann, Pedro Assis, Seve Badajoz, Sidney~M. Bell, Emanuele Bezzi, Batuhan Cakir, Jim Chaffer, Signe Chambers, J.~Michael Cherry, Tiffany Chi, Jennifer Chien, Leah Dorman, Pablo Garcia-Nieto, Nayib Gloria, Mim Hastie, Daniel Hegeman, Jason Hilton, Timmy Huang, Amanda Infeld, Ana-Maria Istrate, Ivana Jelic, Kuni Katsuya, Yang~Joon Kim, Karen Liang, Mike Lin, Maximilian Lombardo, Bailey Marshall, Bruce Martin, Fran McDade, Colin Megill, Nikhil Patel, Alexander Predeus, Brian Raymor, Behnam Robatmili, Dave Rogers, Erica Rutherford, Dana Sadgat, Andrew Shin, Corinn Small, Trent Smith, Prathap Sridharan, Alexander Tarashansky, Norbert Tavares, Harley Thomas, Andrew Tolopko, Meghan Urisko, Joyce Yan, Garabet Yeretssian, Jennifer Zamanian, Arathi Mani, Jonah Cool, and Ambrose Carr.
\newblock {CZ CELLxGENE Discover: a single-cell data platform for scalable exploration, analysis and modeling of aggregated data}.
\newblock \emph{Nucleic Acids Research}, 53\penalty0 (D1):\penalty0 D886--D900, 2025.

\bibitem[Sorin et~al.(2023)Sorin, Rezanejad, Karimi, Fiset, Desharnais, Perus, Milette, Yu, Maritan, Dor{\'e}, et~al.]{sorin2023single}
Mark Sorin, Morteza Rezanejad, Elham Karimi, Benoit Fiset, Lysanne Desharnais, Lucas J.~M. Perus, Simon Milette, Miranda~W. Yu, Sarah~M. Maritan, Samuel Dor{\'e}, et~al.
\newblock {Single-cell spatial landscapes of the lung tumour immune microenvironment}.
\newblock \emph{Nature}, 614\penalty0 (7948):\penalty0 548--554, 2023.

\bibitem[Kraus et~al.(2024)Kraus, Kenyon-Dean, Saberian, Fallah, McLean, Leung, Sharma, Khan, Balakrishnan, Celik, et~al.]{kraus2024masked}
Oren Kraus, Kian Kenyon-Dean, Saber Saberian, Maryam Fallah, Peter McLean, Jess Leung, Vasudev Sharma, Ayla Khan, Jia Balakrishnan, Safiye Celik, et~al.
\newblock {Masked Autoencoders for Microscopy are Scalable Learners of Cellular Biology}.
\newblock In \emph{IEEE/CVF Conference on Computer Vision and Pattern Recognition (CVPR)}, pages 11757--11768, 2024.

\bibitem[Shaban et~al.(2025)Shaban, Chang, Qiu, Yeo, Song, Jaume, Wang, Weishaupt, Ding, Vaidya, et~al.]{shaban2025foundation}
Muhammad Shaban, Yuzhou Chang, Huaying Qiu, Yao~Yu Yeo, Andrew~H. Song, Guillaume Jaume, Yuchen Wang, Luca~L. Weishaupt, Tong Ding, Anurag Vaidya, et~al.
\newblock {A Foundation Model for Spatial Proteomics}.
\newblock \emph{arXiv preprint arXiv:2506.03373}, 2025.

\bibitem[Rigamonti et~al.(2024)Rigamonti, Viatore, Polidori, Rahal, Erreni, Fumagalli, Zanini, Doni, Putignano, Bossi, et~al.]{rigamonti2024integrating}
Alessandra Rigamonti, Marika Viatore, Rebecca Polidori, Daoud Rahal, Marco Erreni, Maria~Rita Fumagalli, Damiano Zanini, Andrea Doni, Anna~Rita Putignano, Paola Bossi, et~al.
\newblock {Integrating AI-Powered Digital Pathology and Imaging Mass Cytometry Identifies Key Classifiers of Tumor Cells, Stroma, and Immune Cells in Non--Small Cell Lung Cancer}.
\newblock \emph{Cancer Research}, 84\penalty0 (7):\penalty0 1165--1177, 2024.

\bibitem[Ruf et~al.(2023)Ruf, Bruhns, Babaei, Kedei, Ma, Revsine, Benmebarek, Ma, Heinrich, Subramanyam, et~al.]{ruf2023tumor}
Benjamin Ruf, Matthias Bruhns, Sepideh Babaei, Noemi Kedei, Lichun Ma, Mahler Revsine, Mohamed-Reda Benmebarek, Chi Ma, Bernd Heinrich, Varun Subramanyam, et~al.
\newblock {Tumor-associated macrophages trigger MAIT cell dysfunction at the HCC invasive margin}.
\newblock \emph{Cell}, 186\penalty0 (17):\penalty0 3686--3705, 2023.

\bibitem[Kundra et~al.(2021)Kundra, Zhang, Sheridan, Sirintrapun, Wang, Ochoa, Wilson, Gross, Sun, Madupuri, et~al.]{kundra2021oncotree}
Ritika Kundra, Hongxin Zhang, Robert Sheridan, Sahussapont~Joseph Sirintrapun, Avery Wang, Angelica Ochoa, Manda Wilson, Benjamin Gross, Yichao Sun, Ramyasree Madupuri, et~al.
\newblock {OncoTree: A Cancer Classification System for Precision Oncology}.
\newblock \emph{JCO Clinical Cancer Informatics}, 5:\penalty0 221--230, 2021.

\bibitem[Moore et~al.(2023)Moore, Basurto-Lozada, Besson, Bogovic, Bragantini, Brown, Burel, Casas~Moreno, de~Medeiros, Diel, et~al.]{moore2023ome}
Josh Moore, Daniela Basurto-Lozada, S{\'e}bastien Besson, John Bogovic, Jord{\~a}o Bragantini, Eva~M. Brown, Jean-Marie Burel, Xavier Casas~Moreno, Gustavo de~Medeiros, Erin~E. Diel, et~al.
\newblock {OME-Zarr: a cloud-optimized bioimaging file format with international community support}.
\newblock \emph{Histochemistry and Cell Biology}, 160\penalty0 (3):\penalty0 223--251, 2023.

\bibitem[Bankhead et~al.(2017)Bankhead, Loughrey, Fern{\'a}ndez, Dombrowski, McArt, Dunne, McQuaid, Gray, Murray, Coleman, et~al.]{bankhead2017qupath}
Peter Bankhead, Maurice~B Loughrey, Jos{\'e}~A Fern{\'a}ndez, Yvonne Dombrowski, Darragh~G McArt, Philip~D Dunne, Stephen McQuaid, Ronan~T Gray, Liam~J Murray, Helen~G Coleman, et~al.
\newblock Qupath: Open source software for digital pathology image analysis.
\newblock \emph{Scientific reports}, 7\penalty0 (1):\penalty0 16878, 2017.

\bibitem[Sofroniew et~al.(2022)Sofroniew, Lambert, Bokota, Nunez-Iglesias, Sobolewski, Sweet, Gaifas, Evans, Burt, Doncila~Pop, et~al.]{sofroniew2022napari}
Nicholas Sofroniew, Talley Lambert, Grzegorz Bokota, Juan Nunez-Iglesias, Peter Sobolewski, Andrew Sweet, Lorenzo Gaifas, Kira Evans, Alister Burt, Draga Doncila~Pop, et~al.
\newblock napari: a multi-dimensional image viewer for python.
\newblock \emph{Zenodo}, 2022.

\bibitem[Schindelin et~al.(2012)Schindelin, Arganda-Carreras, Frise, Kaynig, Longair, Pietzsch, Preibisch, Rueden, Saalfeld, Schmid, et~al.]{schindelin2012fiji}
Johannes Schindelin, Ignacio Arganda-Carreras, Erwin Frise, Verena Kaynig, Mark Longair, Tobias Pietzsch, Stephan Preibisch, Curtis Rueden, Stephan Saalfeld, Benjamin Schmid, et~al.
\newblock Fiji: an open-source platform for biological-image analysis.
\newblock \emph{Nature methods}, 9\penalty0 (7):\penalty0 676--682, 2012.

\bibitem[Wenckstern et~al.(2026)Wenckstern, Jain, von Querfurth, Cheng, Vasilev, Pariset, Cheng, Liakopoulos, Michielin, Wicki, et~al.]{wenckstern2026virtual}
Johann Wenckstern, Eeshaan Jain, Benedikt von Querfurth, Yexiang Cheng, Kiril Vasilev, Matteo Pariset, Phil~F. Cheng, Petros Liakopoulos, Olivier Michielin, Andreas Wicki, et~al.
\newblock {The Virtual Tissues foundation model resolves spatial proteomics across scales}.
\newblock \emph{Nature}, 657\penalty0 (8132):\penalty0 744--754, 2026.

\bibitem[Hochbaum and Pathria(1998)]{hochbaum1998analysis}
Dorit~S. Hochbaum and Anu Pathria.
\newblock {Analysis of the greedy approach in problems of maximum k-coverage}.
\newblock \emph{Naval Research Logistics}, 45\penalty0 (6):\penalty0 615--627, 1998.

\bibitem[Nemhauser et~al.(1978)Nemhauser, Wolsey, and Fisher]{nemhauser1978analysis}
George~L. Nemhauser, Laurence~A. Wolsey, and Marshall~L. Fisher.
\newblock {An analysis of approximations for maximizing submodular set functions---I}.
\newblock \emph{Mathematical Programming}, 14\penalty0 (1):\penalty0 265--294, 1978.

\bibitem[Cords et~al.(2024)Cords, Engler, Haberecker, R{\"u}schoff, Moch, de~Souza, and Bodenmiller]{cords2024cancer}
Lena Cords, Stefanie Engler, Martina Haberecker, Jan~Hendrik R{\"u}schoff, Holger Moch, Natalie de~Souza, and Bernd Bodenmiller.
\newblock {Cancer-associated fibroblast phenotypes are associated with patient outcome in non-small cell lung cancer}.
\newblock \emph{Cancer Cell}, 42\penalty0 (3):\penalty0 396--412, 2024.

\bibitem[Danenberg et~al.(2022)Danenberg, Bardwell, Zanotelli, Provenzano, Chin, Rueda, Green, Rakha, Aparicio, Ellis, et~al.]{danenberg2022breast}
Esther Danenberg, Helen Bardwell, Vito R.~T. Zanotelli, Elena Provenzano, Suet-Feung Chin, Oscar~M. Rueda, Andrew Green, Emad Rakha, Samuel Aparicio, Ian~O. Ellis, et~al.
\newblock {Breast tumor microenvironment structures are associated with genomic features and clinical outcome}.
\newblock \emph{Nature Genetics}, 54\penalty0 (5):\penalty0 660--669, 2022.

\bibitem[Hoch et~al.(2022)Hoch, Schulz, Eling, G{\'o}mez, Levesque, and Bodenmiller]{hoch2022multiplexed}
Tobias Hoch, Daniel Schulz, Nils Eling, Julia~Mart{\'\i}nez G{\'o}mez, Mitchell~P. Levesque, and Bernd Bodenmiller.
\newblock {Multiplexed imaging mass cytometry of the chemokine milieus in melanoma characterizes features of the response to immunotherapy}.
\newblock \emph{Science Immunology}, 7\penalty0 (70):\penalty0 eabk1692, 2022.

\bibitem[Meyer et~al.(2025)Meyer, Jackson, Eling, Zhao, Usui, Dakhli, Schraml, Dettwiler, Elfgen, Varga, et~al.]{meyer2025stratification}
Lasse Meyer, Hartland~W. Jackson, Nils Eling, Shan Zhao, Genki Usui, Haithem Dakhli, Peter Schraml, Susanne Dettwiler, Constanze Elfgen, Zsuzsanna Varga, et~al.
\newblock {A stratification system for breast cancer based on basoluminal tumor cells and spatial tumor architecture}.
\newblock \emph{Cancer Cell}, 43\penalty0 (9):\penalty0 1637--1655, 2025.

\bibitem[Moldoveanu et~al.(2022)Moldoveanu, Ramsay, Lajoie, Anderson-Trocme, Lingrand, Berry, Perus, Wei, Moraes, Alkallas, et~al.]{moldoveanu2022spatially}
Dan Moldoveanu, LeeAnn Ramsay, Mathieu Lajoie, Luke Anderson-Trocme, Marine Lingrand, Diana Berry, Lucas J.~M. Perus, Yuhong Wei, Cleber Moraes, Rached Alkallas, et~al.
\newblock {Spatially mapping the immune landscape of melanoma using imaging mass cytometry}.
\newblock \emph{Science Immunology}, 7\penalty0 (70):\penalty0 eabi5072, 2022.

\bibitem[Phillips et~al.(2021)Phillips, Matusiak, Gutierrez, Bhate, Barlow, Jiang, Demeter, Smythe, Pierce, Fling, et~al.]{phillips2021immune}
Darci Phillips, Magdalena Matusiak, Bel{\'e}n~Rivero Gutierrez, Salil~S. Bhate, Graham~L. Barlow, Sizun Jiang, Janos Demeter, Kimberly~S. Smythe, Robert~H. Pierce, Steven~P. Fling, et~al.
\newblock {Immune cell topography predicts response to PD-1 blockade in cutaneous T cell lymphoma}.
\newblock \emph{Nature Communications}, 12\penalty0 (1):\penalty0 6726, 2021.

\bibitem[Eling et~al.(2025)Eling, Dorier, Rusakiewicz, Liechti, Devanand, Daniel, Windhager, Fernandez, D{\'e}glise, Despland, Benyagoub, Mo{\.z}ejko, Uchal, Szczurek, Loboda, Sandkuijl, Parsotam, Hong, Morfouace, Guex, Coukos, Bodenmiller, Tissot, and Schulz]{eling2025multi}
Nils Eling, Julien Dorier, Sylvie Rusakiewicz, Robin Liechti, Preethi Devanand, Michelle Daniel, Jonas Windhager, Bruno~Palau Fernandez, Sophie D{\'e}glise, Lucie Despland, Abdelkader Benyagoub, Marcin Mo{\.z}ejko, Dawid Uchal, Ewa Szczurek, Alexander Loboda, Daaf Sandkuijl, Nikesh Parsotam, Henoch~S. Hong, Marie Morfouace, Nicolas Guex, George Coukos, Bernd Bodenmiller, Stephanie Tissot, and Daniel Schulz.
\newblock {Multi-modal image analysis for large-scale cancer tissue studies within IMMUcan}.
\newblock \emph{Cell Reports Methods}, 5\penalty0 (9), 2025.

\bibitem[Geuenich et~al.(2021)Geuenich, Hou, Lee, Ayub, Jackson, and Campbell]{geuenich2021automated}
Michael~J. Geuenich, Jinyu Hou, Sunyun Lee, Shanza Ayub, Hartland~W. Jackson, and Kieran~R. Campbell.
\newblock {Automated assignment of cell identity from single-cell multiplexed imaging and proteomic data}.
\newblock \emph{Cell Systems}, 12\penalty0 (12):\penalty0 1173--1186, 2021.

\bibitem[Shaban et~al.(2024)Shaban, Bai, Qiu, Mao, Yeung, Yeo, Shanmugam, Chen, Zhu, Weirather, et~al.]{shaban2024maps}
Muhammad Shaban, Yunhao Bai, Huaying Qiu, Shulin Mao, Jason Yeung, Yao~Yu Yeo, Vignesh Shanmugam, Han Chen, Bokai Zhu, Jason~L. Weirather, et~al.
\newblock {MAPS: pathologist-level cell type annotation from tissue images through machine learning}.
\newblock \emph{Nature Communications}, 15\penalty0 (1):\penalty0 28, 2024.

\bibitem[Pachitariu et~al.(2025)Pachitariu, Rariden, and Stringer]{pachitariu2025cellpose}
Marius Pachitariu, Michael Rariden, and Carsen Stringer.
\newblock {Cellpose-SAM: superhuman generalization for cellular segmentation}.
\newblock \emph{bioRxiv preprint doi:10.1101/2025.04.28.651001}, 2025.

\bibitem[Goldsborough et~al.(2024)Goldsborough, Philps, O'Callaghan, Inglis, Leplat, Filby, Bilen, and Bankhead]{goldsborough2024instanseg}
Thibaut Goldsborough, Ben Philps, Alan O'Callaghan, Fiona Inglis, Leo Leplat, Andrew Filby, Hakan Bilen, and Peter Bankhead.
\newblock {InstanSeg: an embedding-based instance segmentation algorithm optimized for accurate, efficient and portable cell segmentation}.
\newblock \emph{arXiv preprint arXiv:2408.15954}, 2024.

\bibitem[Weigert and Schmidt(2022)]{weigert2022nuclei}
Martin Weigert and Uwe Schmidt.
\newblock {Nuclei Instance Segmentation and Classification in Histopathology Images with Stardist}.
\newblock In \emph{IEEE International Symposium on Biomedical Imaging Challenges (ISBIC)}, pages 1--4, 2022.

\bibitem[Klein et~al.(2026)Klein, Ziegler, Gerst, Morgenroth, Gotkowski, Sch{\"o}niger, Heni, Kipke, Friedland, Seiler, et~al.]{klein2026explainable}
Lukas Klein, Sebastian Ziegler, Felicia Gerst, Yanni Morgenroth, Karol Gotkowski, Eyke Sch{\"o}niger, Martin Heni, Nicole Kipke, Daniela Friedland, Annika Seiler, et~al.
\newblock {Explainable AI-based analysis of human pancreas sections identifies traits of type 2 diabetes}.
\newblock \emph{Nature Communications}, 17\penalty0 (1):\penalty0 1558, 2026.

\bibitem[Ilse et~al.(2018)Ilse, Tomczak, and Welling]{ilse2018attention}
Maximilian Ilse, Jakub Tomczak, and Max Welling.
\newblock {Attention-based Deep Multiple Instance Learning}.
\newblock In \emph{International Conference on Machine Learning (ICML)}, pages 2127--2136, 2018.

\bibitem[Liu et~al.(2025)Liu, Sharma, Bieniosek, Kang, Wu, Chou, Li, Rahim, Bauer, Ji, et~al.]{liu2025modeling}
Yufan Liu, Rishabh Sharma, Matthew Bieniosek, Amy Kang, Eric Wu, Peter Chou, Irene Li, Maha Rahim, Erica Bauer, Ran Ji, et~al.
\newblock {Modeling patient tissues at molecular resolution with Eva}.
\newblock \emph{bioRxiv preprint doi:10.64898/2025.12.10.693553}, 2025.

\bibitem[Hickey et~al.(2023)Hickey, Becker, Nevins, Horning, Perez, Zhu, Zhu, Wei, Chiu, Chen, et~al.]{hickey2023organization}
John~W. Hickey, Winston~R. Becker, Stephanie~A. Nevins, Aaron Horning, Almudena~Espin Perez, Chenchen Zhu, Bokai Zhu, Bei Wei, Roxanne Chiu, Derek~C. Chen, et~al.
\newblock {Organization of the human intestine at single-cell resolution}.
\newblock \emph{Nature}, 619\penalty0 (7970):\penalty0 572--584, 2023.

\bibitem[Sch{\"u}rch et~al.(2020)Sch{\"u}rch, Bhate, Barlow, Phillips, Noti, Zlobec, Chu, Black, Demeter, McIlwain, et~al.]{schurch2020coordinated}
Christian~M. Sch{\"u}rch, Salil~S. Bhate, Graham~L. Barlow, Darci~J. Phillips, Luca Noti, Inti Zlobec, Pauline Chu, Sarah Black, Janos Demeter, David~R. McIlwain, et~al.
\newblock {Coordinated cellular neighborhoods orchestrate antitumoral immunity at the colorectal cancer invasive front}.
\newblock \emph{Cell}, 182\penalty0 (5):\penalty0 1341--1359, 2020.

\bibitem[Valanarasu et~al.(2025)Valanarasu, Xu, Usuyama, Kim, Wong, Argaw, Shimol, Crabtree, Matlock, Bartlett, et~al.]{valanarasu2025multimodal}
Jeya Maria~Jose Valanarasu, Hanwen Xu, Naoto Usuyama, Chanwoo Kim, Cliff Wong, Peniel Argaw, Racheli~Ben Shimol, Angela Crabtree, Kevin Matlock, Alexandra~Q. Bartlett, et~al.
\newblock {Multimodal AI generates virtual population for tumor microenvironment modeling}.
\newblock \emph{Cell}, 2025.

\bibitem[Andani et~al.(2025)Andani, Chen, Ficek-Pascual, Heinke, Casanova, Hild, Sobottka, Bodenmiller, {Tumor Profiler Consortium}, Koelzer, and R{\"a}tsch]{andani2025histopathology}
Sonali Andani, Boqi Chen, Joanna Ficek-Pascual, Simon Heinke, Ruben Casanova, Bernard~Friedrich Hild, Bettina Sobottka, Bernd Bodenmiller, {Tumor Profiler Consortium}, Viktor~H. Koelzer, and Gunnar R{\"a}tsch.
\newblock {Histopathology-based protein multiplex generation using deep learning}.
\newblock \emph{Nature Machine Intelligence}, 7\penalty0 (8):\penalty0 1292--1307, 2025.

\bibitem[Wu et~al.(2025)Wu, Bieniosek, Wu, Thakkar, Charville, Makky, Sch{\"u}rch, Huyghe, Peters, Li, et~al.]{wu2025rosie}
Eric Wu, Matthew Bieniosek, Zhenqin Wu, Nitya Thakkar, Gregory~W. Charville, Ahmad Makky, Christian~M. Sch{\"u}rch, Jeroen~R. Huyghe, Ulrike Peters, Christopher~I. Li, et~al.
\newblock {ROSIE: AI generation of multiplex immunofluorescence staining from histopathology images}.
\newblock \emph{Nature Communications}, 16\penalty0 (1):\penalty0 7633, 2025.

\bibitem[Li et~al.(2023)Li, Dou, Leprevost, Geffen, Calinawan, Aguet, Akiyama, Anand, Birger, Cao, et~al.]{li2023proteogenomic}
Yize Li, Yongchao Dou, Felipe Da~Veiga Leprevost, Yifat Geffen, Anna~P. Calinawan, Fran{\c{c}}ois Aguet, Yo~Akiyama, Shankara Anand, Chet Birger, Song Cao, et~al.
\newblock {Proteogenomic data and resources for pan-cancer analysis}.
\newblock \emph{Cancer Cell}, 41\penalty0 (8):\penalty0 1397--1406, 2023.

\bibitem[Weitz et~al.(2023)Weitz, Valkonen, Solorzano, Carr, Kartasalo, Boissin, Koivukoski, Kuusela, Rasic, Feng, et~al.]{weitz2023multi}
Philippe Weitz, Masi Valkonen, Leslie Solorzano, Circe Carr, Kimmo Kartasalo, Constance Boissin, Sonja Koivukoski, Aino Kuusela, Dusan Rasic, Yanbo Feng, et~al.
\newblock {A multi-stain breast cancer histological whole-slide-image data set from routine diagnostics}.
\newblock \emph{Scientific Data}, 10\penalty0 (1):\penalty0 562, 2023.

\bibitem[Xu et~al.(2024{\natexlab{b}})Xu, Usuyama, Bagga, Zhang, Rao, Naumann, Wong, Gero, Gonz{\'a}lez, Gu, Xu, Wei, Wang, Ma, Wei, Yang, Li, Gao, Rosemon, Bower, Lee, Weerasinghe, Wright, Robicsek, Piening, Bifulco, Wang, and Poon]{xu_whole-slide_2024}
Hanwen Xu, Naoto Usuyama, Jaspreet Bagga, Sheng Zhang, Rajesh Rao, Tristan Naumann, Cliff Wong, Zelalem Gero, Javier Gonz{\'a}lez, Yu~Gu, Yanbo Xu, Mu~Wei, Wenhui Wang, Shuming Ma, Furu Wei, Jianwei Yang, Chunyuan Li, Jianfeng Gao, Jaylen Rosemon, Tucker Bower, Soohee Lee, Roshanthi Weerasinghe, Bill~J. Wright, Ari Robicsek, Brian Piening, Carlo Bifulco, Sheng Wang, and Hoifung Poon.
\newblock {A whole-slide foundation model for digital pathology from real-world data}.
\newblock \emph{Nature}, 630\penalty0 (8015):\penalty0 181--188, 2024{\natexlab{b}}.

\bibitem[Ehteshami~Bejnordi et~al.(2017)Ehteshami~Bejnordi, Veta, Johannes~van Diest, van Ginneken, Karssemeijer, Litjens, van~der Laak, and {the CAMELYON16 Consortium}]{bejnordi2017diagnostic}
Babak Ehteshami~Bejnordi, Mitko Veta, Paul Johannes~van Diest, Bram van Ginneken, Nico Karssemeijer, Geert Litjens, Jeroen A. W.~M. van~der Laak, and {the CAMELYON16 Consortium}.
\newblock {Diagnostic Assessment of Deep Learning Algorithms for Detection of Lymph Node Metastases in Women With Breast Cancer}.
\newblock \emph{JAMA}, 318\penalty0 (22):\penalty0 2199--2210, 2017.

\bibitem[Bandi et~al.(2019)Bandi, Geessink, Manson, van Dijk, Balkenhol, Hermsen, Ehteshami~Bejnordi, Lee, Paeng, Zhong, et~al.]{bandi2018detection}
Peter Bandi, Oscar Geessink, Quirine Manson, Marcory van Dijk, Maschenka Balkenhol, Meyke Hermsen, Babak Ehteshami~Bejnordi, Byungjae Lee, Kyunghyun Paeng, Aoxiao Zhong, et~al.
\newblock {From Detection of Individual Metastases to Classification of Lymph Node Status at the Patient Level: The CAMELYON17 Challenge}.
\newblock \emph{IEEE Transactions on Medical Imaging}, 38\penalty0 (2):\penalty0 550--560, 2019.

\bibitem[Bulten et~al.(2022)Bulten, Kartasalo, Chen, Str{\"o}m, Pinckaers, Nagpal, Cai, Steiner, van Boven, Vink, et~al.]{bulten2022artificial}
Wouter Bulten, Kimmo Kartasalo, Po-Hsuan~Cameron Chen, Peter Str{\"o}m, Hans Pinckaers, Kunal Nagpal, Yuannan Cai, David~F. Steiner, Hester van Boven, Robert Vink, et~al.
\newblock {Artificial intelligence for diagnosis and Gleason grading of prostate cancer: the PANDA challenge}.
\newblock \emph{Nature Medicine}, 28\penalty0 (1):\penalty0 154--163, 2022.

\bibitem[Huo et~al.(2024)Huo, Ong, Lau, Gole, Young, Tan, Zhu, Zhang, Zhang, Li, et~al.]{huo2024comprehensive}
Xinmi Huo, Kok~Haur Ong, Kah~Weng Lau, Laurent Gole, David~M. Young, Char~Loo Tan, Xiaohui Zhu, Chongchong Zhang, Yonghui Zhang, Longjie Li, et~al.
\newblock {A comprehensive AI model development framework for consistent Gleason grading}.
\newblock \emph{Communications Medicine}, 4\penalty0 (1):\penalty0 84, 2024.

\bibitem[Han et~al.(2022)Han, Pan, Yan, Lin, Li, Yao, Lv, Shi, Mai, Lin, Zhao, Xu, Wang, Wang, Zhang, Wang, Zhu, Lin, Mao, Wu, Duan, Zhu, Hu, Fang, Chen, Zhang, Li, Zou, Yu, Li, Li, Cui, Han, Xu, Xu, Yang, Li, Liu, Lu, Chen, Liang, Zhang, and Liu]{han2022wsss4luad}
Chu Han, Xipeng Pan, Lixu Yan, Huan Lin, Bingbing Li, Su~Yao, Shanshan Lv, Zhenwei Shi, Jinhai Mai, Jiatai Lin, Bingchao Zhao, Zeyan Xu, Zhizhen Wang, Yumeng Wang, Yuan Zhang, Huihui Wang, Chao Zhu, Chunhui Lin, Lijian Mao, Min Wu, Luwen Duan, Jingsong Zhu, Dong Hu, Zijie Fang, Yang Chen, Yongbing Zhang, Yi~Li, Yiwen Zou, Yiduo Yu, Xiaomeng Li, Haiming Li, Yanfen Cui, Guoqiang Han, Yan Xu, Jun Xu, Huihua Yang, Chunming Li, Zhenbing Liu, Cheng Lu, Xin Chen, Changhong Liang, Qingling Zhang, and Zaiyi Liu.
\newblock {WSSS4LUAD: Grand Challenge on Weakly-supervised Tissue Semantic Segmentation for Lung Adenocarcinoma}.
\newblock \emph{arXiv preprint arXiv:2204.06455}, 2022.

\bibitem[Luecken et~al.(2022)Luecken, B{\"u}ttner, Chaichoompu, Danese, Interlandi, M{\"u}ller, Strobl, Zappia, Dugas, Colom{\'e}-Tatch{\'e}, and Theis]{luecken2022benchmarking}
Malte~D. Luecken, Maren B{\"u}ttner, Kridsadakorn Chaichoompu, Anna Danese, Marta Interlandi, Michaela~F. M{\"u}ller, Daniel~C. Strobl, Luke Zappia, Martin Dugas, Maria Colom{\'e}-Tatch{\'e}, and Fabian~J. Theis.
\newblock {Benchmarking atlas-level data integration in single-cell genomics}.
\newblock \emph{Nature Methods}, 19\penalty0 (1):\penalty0 41--50, 2022.

\bibitem[Luecken et~al.(2021)Luecken, Burkhardt, Cannoodt, Lance, Agrawal, Aliee, Chen, Deconinck, Detweiler, Granados, et~al.]{luecken2021sandbox}
Malte~D. Luecken, Daniel~Bernard Burkhardt, Robrecht Cannoodt, Christopher Lance, Aditi Agrawal, Hananeh Aliee, Ann~T. Chen, Louise Deconinck, Angela~M. Detweiler, Alejandro~A. Granados, et~al.
\newblock {A sandbox for prediction and integration of DNA, RNA, and proteins in single cells}.
\newblock In \emph{Proceedings of the Neural Information Processing Systems Track on Datasets and Benchmarks}, 2021.

\bibitem[Wang et~al.(2023)Wang, Danenberg, Huang, Egle, Callari, Bermejo, Dugo, Zamagni, Thill, Anton, et~al.]{wang2023spatial}
Xiao~Qian Wang, Esther Danenberg, Chiun-Sheng Huang, Daniel Egle, Maurizio Callari, Bego{\~n}a Bermejo, Matteo Dugo, Claudio Zamagni, Marc Thill, Anton Anton, et~al.
\newblock {Spatial predictors of immunotherapy response in triple-negative breast cancer}.
\newblock \emph{Nature}, 621\penalty0 (7980):\penalty0 868--876, 2023.

\bibitem[Fischer et~al.(2023)Fischer, Jackson, de~Souza, Varga, Schraml, Moch, and Bodenmiller]{fischer2023multiplex}
Jana~Raja Fischer, Hartland~Warren Jackson, Natalie de~Souza, Zsuzsanna Varga, Peter Schraml, Holger Moch, and Bernd Bodenmiller.
\newblock {Multiplex imaging of breast cancer lymph node metastases identifies prognostic single-cell populations independent of clinical classifiers}.
\newblock \emph{Cell Reports Medicine}, 4\penalty0 (3), 2023.

\bibitem[Li et~al.(2026)Li, Li, Xiang, Wang, Yang, Zhang, Eweje, Chen, Luo, Li, et~al.]{li2026ai}
Zhe Li, Yuchen Li, Jinxi Xiang, Xiyue Wang, Sen Yang, Xiaoming Zhang, Feyisope Eweje, Yijiang Chen, Xiangde Luo, Yuanyuan Li, et~al.
\newblock {AI-enabled virtual spatial proteomics from histopathology for interpretable biomarker discovery in lung cancer}.
\newblock \emph{Nature Medicine}, 32\penalty0 (1):\penalty0 231--244, 2026.

\bibitem[Lee et~al.(2021)Lee, Yeong, Lee, Loo, and Dong]{lee2021627}
Jia Ying~Joey Lee, Joe Yeong, Li~Wen Justina~Nadia Lee, Lit-Hsin Loo, and Jiahui Dong.
\newblock {627 ImmunoAtlas: an online public portal for sharing, visualizing, and referencing multiplex immunohistochemistry/immunofluorescence (mIHC/IF) images and results for immuno-oncology}.
\newblock \emph{Journal for ImmunoTherapy of Cancer}, 9\penalty0 (Suppl 2):\penalty0 A657, 2021.

\bibitem[Allam et~al.(2022)Allam, Hu, Lee, Aldrich, Badve, G{\"o}kmen-Polar, Bhave, Ramalingam, Schneider, and Coskun]{allam2022spatially}
Mayar Allam, Thomas Hu, Jeongjin Lee, Jeffrey Aldrich, Sunil~S. Badve, Yesim G{\"o}kmen-Polar, Manali Bhave, Suresh~S. Ramalingam, Frank Schneider, and Ahmet~F. Coskun.
\newblock {Spatially variant immune infiltration scoring in human cancer tissues}.
\newblock \emph{npj Precision Oncology}, 6\penalty0 (1):\penalty0 60, 2022.

\bibitem[Bowen et~al.(2023)Bowen, Deng, Reyes-Uribe, Parra, Rocha, Solis, Wistuba, Sepeda, Vornik, Perloff, Szabo, Umar, Sinha, Brown, and Vilar]{bowen2023naproxen}
Charles~M. Bowen, Nan Deng, Laura Reyes-Uribe, Edwin~Roger Parra, Pedro Rocha, Luisa~M. Solis, Ignacio~I. Wistuba, Valerie~O. Sepeda, Lana Vornik, Marjorie Perloff, Eva Szabo, Asad Umar, Krishna~M. Sinha, Powel~H. Brown, and Eduardo Vilar.
\newblock {Naproxen chemoprevention induces proliferation of cytotoxic lymphocytes in Lynch Syndrome colorectal mucosa}.
\newblock \emph{Frontiers in Immunology}, 14, 2023.

\bibitem[Cho et~al.(2026)Cho, Lee, Shin, Hernandez, Yuan, Schneider, Hooper, Wood, Jaffee, Deshpande, and Ho]{cho2025modeling}
Yeonju Cho, Jae~W. Lee, Sarah~M. Shin, Alexei~G. Hernandez, Xuan Yuan, Jowaly Schneider, Jody~E. Hooper, Laura~D. Wood, Elizabeth~M. Jaffee, Atul Deshpande, and Won~Jin Ho.
\newblock {Computational Modeling of Cellular Influence Delineates Functionally Relevant and Distinct Cellular Neighborhoods in Primary and Metastatic Pancreatic Ductal Adenocarcinoma}.
\newblock \emph{Cancer Immunology Research}, 14\penalty0 (4):\penalty0 571--584, 2026.

\bibitem[Damond et~al.(2019)Damond, Engler, Zanotelli, Schapiro, Wasserfall, Kusmartseva, Nick, Thorel, Herrera, Atkinson, et~al.]{damond2019map}
Nicolas Damond, Stefanie Engler, Vito R.~T. Zanotelli, Denis Schapiro, Clive~H. Wasserfall, Irina Kusmartseva, Harry~S. Nick, Fabrizio Thorel, Pedro~L. Herrera, Mark~A. Atkinson, et~al.
\newblock {A map of human type 1 diabetes progression by imaging mass cytometry}.
\newblock \emph{Cell Metabolism}, 29\penalty0 (3):\penalty0 755--768, 2019.

\bibitem[Ehret(2025)]{ehret2025shaking}
Benjamin Ehret.
\newblock {IMC data associated with the publication `Shaking organoid co-cultures: human-relevant models of epithelial, stromal and immune interactions for cancer research and immunotherapy development'}, 2025.
\newblock URL \url{https://doi.org/10.5281/zenodo.15198804}.

\bibitem[Greenwald et~al.(2026)Greenwald, Nederlof, Sowers, Ding, Park, Kong, Houlahan, Varra, de~Graaf, Geurts, et~al.]{greenwald2026temporal}
Noah~F. Greenwald, Iris Nederlof, Cameron Sowers, Daisy~Yi Ding, Seongyeol Park, Alex Kong, Kathleen~E. Houlahan, Sricharan~Reddy Varra, Manon de~Graaf, Veerle Geurts, et~al.
\newblock {Temporal and spatial composition of the tumor microenvironment predicts response to immune checkpoint inhibition in metastatic TNBC}.
\newblock \emph{Nature Cancer}, 7\penalty0 (3):\penalty0 435--450, 2026.

\bibitem[Gupta et~al.(2026)Gupta, Lee, Masqu{\'e}~Soler, Schrader, Wang, Mayer, Flores, Beatty, Roth, Aparicio, et~al.]{gupta2026single}
Pulkit Gupta, Eric Lee, Neus Masqu{\'e}~Soler, Ellen Schrader, Xiao~Qian Wang, Shimrit Mayer, Cristina Flores, Sean Beatty, Andrew Roth, Samuel Aparicio, et~al.
\newblock {Single-cell spatial atlas of the aging human breast}.
\newblock \emph{Nature Aging}, 6\penalty0 (4):\penalty0 916--931, 2026.

\bibitem[Haist et~al.(2026)Haist, Baertsch, Reticker-Flynn, Lu, Kempchen, Chu, Vazquez, Chen, Sunwoo, Zhang, et~al.]{haist2026lymph}
Maximilian Haist, Marc-A. Baertsch, Nathan~E. Reticker-Flynn, Guolan Lu, Tim~N. Kempchen, Pauline Chu, Gustavo Vazquez, Han Chen, John~B. Sunwoo, Weiruo Zhang, et~al.
\newblock {Lymph node colonization induces tissue remodeling via immunosuppressive fibroblast-myeloid cell niches supporting metastatic tolerance}.
\newblock \emph{Cancer Cell}, 44\penalty0 (3):\penalty0 604--623, 2026.

\bibitem[{HuBMAP Consortium}(2019)]{hubmap2019human}
{HuBMAP Consortium}.
\newblock {The human body at cellular resolution: the NIH Human Biomolecular Atlas Program}.
\newblock \emph{Nature}, 574\penalty0 (7777):\penalty0 187--192, 2019.

\bibitem[Hu et~al.(2023)Hu, Allam, Cai, Henderson, Yueh, Garipcan, Ievlev, Afkarian, Beyaz, and Coskun]{hu2023single}
Thomas Hu, Mayar Allam, Shuangyi Cai, Walter Henderson, Brian Yueh, Aybuke Garipcan, Anton~V. Ievlev, Maryam Afkarian, Semir Beyaz, and Ahmet~F. Coskun.
\newblock {Single-cell spatial metabolomics with cell-type specific protein profiling for tissue systems biology}.
\newblock \emph{Nature Communications}, 14\penalty0 (1):\penalty0 8260, 2023.

\bibitem[Jackson et~al.(2020)Jackson, Fischer, Zanotelli, Ali, Mechera, Soysal, Moch, Muenst, Varga, Weber, et~al.]{jackson2020single}
Hartland~W. Jackson, Jana~R. Fischer, Vito R.~T. Zanotelli, H.~Raza Ali, Robert Mechera, Savas~D. Soysal, Holger Moch, Simone Muenst, Zsuzsanna Varga, Walter~P. Weber, et~al.
\newblock {The single-cell pathology landscape of breast cancer}.
\newblock \emph{Nature}, 578\penalty0 (7796):\penalty0 615--620, 2020.

\bibitem[Lin et~al.(2023{\natexlab{b}})Lin, Wang, Coy, Chen, Yapp, Tyler, Nariya, Heiser, Lau, Santagata, et~al.]{lin2023multiplexed}
Jia-Ren Lin, Shu Wang, Shannon Coy, Yu-An Chen, Clarence Yapp, Madison Tyler, Maulik~K. Nariya, Cody~N. Heiser, Ken~S. Lau, Sandro Santagata, et~al.
\newblock {Multiplexed 3D atlas of state transitions and immune interaction in colorectal cancer}.
\newblock \emph{Cell}, 186\penalty0 (2):\penalty0 363--381, 2023{\natexlab{b}}.

\bibitem[Liu et~al.(2022)Liu, Bosse, Kong, Kagel, Kinders, Hewitt, Varma, van~de Rijn, Nowak, Bendall, et~al.]{liu2022reproducible}
Candace~C. Liu, Marc Bosse, Alex Kong, Adam Kagel, Robert Kinders, Stephen~M. Hewitt, Sushama Varma, Matt van~de Rijn, Stanis{\l}aw~H. Nowak, Sean~C. Bendall, et~al.
\newblock {Reproducible, high-dimensional imaging in archival human tissue by multiplexed ion beam imaging by time-of-flight (MIBI-TOF)}.
\newblock \emph{Laboratory Investigation}, 102\penalty0 (7):\penalty0 762--770, 2022.

\bibitem[Malchiodi et~al.(2024)Malchiodi, Suter, Deshpande, Peran, Harris, Duttargi, Chien, Hariharan, Wetherill, Jablonski, et~al.]{malchiodi2024natural}
Zoe~X. Malchiodi, Robert~K. Suter, Atul Deshpande, Ivana Peran, Brent~T. Harris, Anju Duttargi, Min-Ju Chien, Samika Hariharan, Lucia Wetherill, Sandra~A. Jablonski, et~al.
\newblock {Natural killer cells associate with epithelial cells in the pancreatic ductal adenocarcinoma tumor microenvironment}.
\newblock \emph{bioRxiv preprint doi:10.1101/2024.05.23.593868}, 2024.

\bibitem[Monkman et~al.(2024)Monkman, Moradi, Yunis, Ivison, Mayer, Ladwa, O'Byrne, and Kulasinghe]{monkman2024spatial}
James Monkman, Afshin Moradi, Joseph Yunis, Geoff Ivison, Aaron Mayer, Rahul Ladwa, Ken O'Byrne, and Arutha Kulasinghe.
\newblock {Spatial insights into immunotherapy response in non-small cell lung cancer (NSCLC) by multiplexed tissue imaging}.
\newblock \emph{Journal of Translational Medicine}, 22\penalty0 (1):\penalty0 239, 2024.

\bibitem[Schulz et~al.(2018)Schulz, Zanotelli, Fischer, Schapiro, Engler, Lun, Jackson, and Bodenmiller]{schulz2018simultaneous}
Daniel Schulz, Vito Riccardo~Tomaso Zanotelli, Jana~Raja Fischer, Denis Schapiro, Stefanie Engler, Xiao-Kang Lun, Hartland~Warren Jackson, and Bernd Bodenmiller.
\newblock {Simultaneous multiplexed imaging of mRNA and proteins with subcellular resolution in breast cancer tissue samples by mass cytometry}.
\newblock \emph{Cell Systems}, 6\penalty0 (1):\penalty0 25--36, 2018.

\bibitem[Zhu et~al.(2025)Zhu, Chen, Aminu, Li, Fujimoto, Tian, Hong, Chen, Hu, Li, et~al.]{zhu2025spatial}
Bo~Zhu, Pingjun Chen, Muhammad Aminu, Jian-Rong Li, Junya Fujimoto, Yanhua Tian, Lingzhi Hong, Hong Chen, Xin Hu, Chenyang Li, et~al.
\newblock {Spatial and multiomics analysis of human and mouse lung adenocarcinoma precursors reveals TIM-3 as a putative target for precancer interception}.
\newblock \emph{Cancer Cell}, 43\penalty0 (6):\penalty0 1125--1140, 2025.

\bibitem[Kingma and Ba(2015)]{kingma2014adam}
Diederik~P. Kingma and Jimmy Ba.
\newblock {Adam: A Method for Stochastic Optimization}.
\newblock In \emph{International Conference on Learning Representations (ICLR)}, 2015.

\end{thebibliography}

\clearpage
\appendix
\setcounter{figure}{0}
\setcounter{table}{0}
\renewcommand{\thefigure}{A\arabic{figure}}
\renewcommand{\thetable}{A\arabic{table}}

\section*{\Large Appendix}

\etocdepthtag.toc{mtappendix}
\etocsettagdepth{mtmain}{none}
\etocsettagdepth{mtappendix}{subsection}
\etocsettocstyle{}{}
\tableofcontents
\newpage

\setcounter{table}{0}
\setcounter{figure}{0}

\section{Related Work}
\label{sec:related_work}

\paragraph{Histopathology atlases and benchmarks.}
Histopathology has been the testbed for the data-then-models trajectory in medical imaging. TCGA \citep{tomczak2015review}, CPTAC \citep{li2023proteogenomic}, ACROBAT \citep{weitz2023multi}, and HistAI \citep{nechaev2025histaiopensourcelargescaleslide} together provided H\&E corpora at the scale of hundreds of thousands of WSIs, behind the current generation of histology foundation models \citep{chen2024uni, zimmermann2024virchow2, xu_whole-slide_2024}. Standardized benchmarks consolidated alongside the data: CAMELYON16 \citep{bejnordi2017diagnostic} and CAMELYON17 \citep{bandi2018detection} for metastasis detection, PANDA \citep{bulten2022artificial} for prostate cancer grading, AGGC \citep{huo2024comprehensive} for gleason grading, or WSSS4LUAD \citep{han2022wsss4luad} for lung adenocarcinoma tissue differentiation.

\paragraph{Spatial transcriptomics atlases and benchmarks.}
Spatial transcriptomics is consolidating along the same trajectory on a shorter timeline. HEST-1k \citep{jaume2024hest} and STOmicsDB \citep{xu2024stomicsdb} aggregated paired ST and H\&E samples in unified formats, enabling the first wave of FM pretraining for the modality. HEST-Bench, released with HEST-1k, evaluates ST FMs on gene expression prediction from H\&E and serves as the principal benchmark for the modality.

\paragraph{Single-cell RNA-seq atlases and benchmarks.}
Single-cell RNA-seq consolidated earlier than the spatial modalities and offers the clearest precedent. CELLxGENE \citep{program_cz_2025} aggregates more than fifty million cells from thousands of studies into a single queryable resource with harmonized ontologies and metadata, supporting cross-study atlases and foundation models. scIB \citep{luecken2022benchmarking} provided the first standardized comparison framework for atlas-level integration, and Open Problems in Single-Cell Analysis \citep{luecken2021sandbox} extended this to label transfer, denoising, and modality prediction. The trajectory illustrates how atlas-scale data and shared benchmarks compound, accelerating the field's progress toward foundation models.

\paragraph{Spatial proteomics datasets and benchmarks.}
SP has not yet undergone this consolidation. Public SP cohorts are released study by study, each typically restricted to a single cancer type, technology, and institution \citep{lin2023high, wang2023spatial, fischer2023multiplex, cords2024cancer}, and no resource of comparable scale to TCGA, HEST-1k, or CELLxGENE exists. The set of foundation models trained on SP is correspondingly limited and is described in \aref{app:sec:sp_fms}. A complementary line of work synthesizes SP from H\&E through virtual staining \citep{wu2025rosie, andani2025histopathology, valanarasu2025multimodal, li2026ai}, but prediction quality varies across organs, cancer types, and markers, and cannot be validated without ground-truth SP on the target cohorts. On the evaluation side, no standardized large-scale SP benchmark exists: existing tasks are extracted from individual biological studies \citep{wang2023spatial, cords2024cancer, lee2021627}, each confined to a single cohort and heterogeneous in organ, cancer type, technology, and target label, with baselines from classical computational biology rather than from comparable FM evaluation. 

\spora closes both gaps. 
It aggregates 35 cohorts under a single format with paired H\&E where the source studies allow, and \sporabench provides a unified cell- and tissue-level evaluation across cohorts, technologies, and cancer types, enabling the first head-to-head comparison of SP foundation models on a common protocol.

\section{\spora Dataset Metadata Overview} \label{app:sec:1}

\begin{sidewaystable}[!htbp]
\centering
\scriptsize
\setlength{\tabcolsep}{2.2pt}
\renewcommand{\arraystretch}{1.2}
\caption{Overview of cohorts included in \textsc{spora}. mx = multiplex spatial proteomics images. Alignment indicates whether MX and H\&E modalities are registered at the pixel level or unregistered but from matched samples. Specimen type: WSI = whole-slide image, TC = tumor core. Demographics lists the patient-level demographic fields available in the released metadata; $^\dagger$age provided as ranges only. Additional metadata might be available in the referenced publications.}
\label{tab:cohort_overview}

\newcolumntype{L}[1]{>{\raggedright\arraybackslash}p{#1}}

\begin{adjustbox}{max width=\textheight, max totalheight=\textwidth, keepaspectratio}
\begin{tabular}{@{}%
L{6.0cm}  
L{1.7cm}  
L{1.8cm}  
r         
L{0.8cm}  
L{2.35cm} 
L{2.45cm} 
L{2.0cm}  
r         
r         
r         
r         
L{2.5cm}  
@{}}
\toprule
\textbf{Dataset} &
\textbf{Modalities} &
\textbf{Alignment} &
\textbf{\# Patients} &
\textbf{\makecell[l]{Image\\Type}} &
\textbf{Organ} &
\textbf{\makecell[l]{Primary\\Cancer}} &
\textbf{Demographics} &
\textbf{\makecell{\#MX\\imgs}} &
\textbf{\makecell{\#MX\\crops}} &
\textbf{\makecell{\#H\&E\\imgs}} &
\textbf{\makecell{\#H\&E\\crops}} &
\textbf{License} \\
\midrule
\texttt{allam2022spatially} \citep{allam2022spatially} & IMC & -- & 26 & TC & Lung, Lymphnode & Lung & -- & 26 & 2,592 & -- & -- & CC BY 4.0 \\
\texttt{bowen2023naproxen} \citep{bowen2023naproxen} & IMC & -- & 18 & TC & Colon & Other disease & -- & 58 & 721 & -- & -- & CC BY 4.0 \\
\texttt{cho2025modeling} \citep{cho2025modeling} & IMC & -- & 12 & TC & Pancreas, Liver & Pancreatic & Age$^\dagger$, Sex & 61 & 3,664 & -- & -- & CC BY 4.0 \\
\texttt{cords2023cancer} \citep{cords2023cancer} & IMC & -- & 12 & TC & Breast & Breast & Age$^\dagger$, Sex & 109 & 6,820 & -- & -- & CC BY 4.0 \\
\texttt{cords2024cancer} \citep{cords2024cancer} & IMC, H\&E & matched & 1,361 & TC & Lung & Lung & Age, Sex, Smoking & 2,040 & 76,796 & 2,328 & 87,637 & CC BY 4.0 \\
\texttt{damond2019map} \citep{damond2019map} & IMC & -- & 12 & TC & Pancreas & Other disease & -- & 845 & 19,156 & -- & -- & CC BY 4.0 \\
\texttt{danenberg2022breast} \citep{danenberg2022breast} & IMC & -- & 718 & TC & Breast & Breast & -- & 718 & 26,012 & -- & -- & CC BY 4.0 \\
\texttt{ehret2025shaking} \citep{ehret2025shaking} & IMC & -- & 55 & TC & Colon & Colorectal & Age, Sex & 108 & 5,832 & -- & -- & CC BY 4.0 \\
\texttt{fischer2023multiplex} \citep{fischer2023multiplex} & IMC & matched & 892 & TC & Breast & Breast & Age, Sex & 1,252 & 43,864 & -- & -- & CC BY 4.0 \\
\texttt{greenwald2026temporal} \citep{greenwald2026temporal} & MIBI & -- & 113 & TC & Breast & Breast & -- & 1,258 & 20,128 & -- & -- & CC SA 4.0 \\
\texttt{gupta2026single} \citep{gupta2026single} & IMC & -- & 527 & TC & Breast & No disease & Age & 1,492 & 197,568 & -- & -- & CC BY 4.0 \\
\texttt{haist2026lymph} \citep{haist2026lymph} & CODEX, H\&E & registered & 79 & TC & Lymphnode & Head and Neck & Age, Sex, Smoking & 389 & 3,388 & 389 & 3,388 & CC BY 4.0 \\
\texttt{hickey2023organization} \citep{hickey2023organization} & CODEX, H\&E & registered & 8 & WSI & Colon & Other disease & -- & 64 & 22,544 & 64 & 14,300 & CC0 1.0 \\
\texttt{hoch2022multiplexed} \citep{hoch2022multiplexed} & IMC & -- & 35 & TC & Skin, Lymphnode & Skin (melanoma) & Age$^\dagger$, Sex & 50 & 3,408 & -- & -- & CC BY 4.0 \\
\texttt{hoefflin2026spatial} \citep{klein2026explainable} & CODEX, H\&E & registered & 25 & WSI & Brain & Brain & Age, Sex & 30 & 22,364 & 30 & 22,364 & CC BY 4.0 \\
\texttt{hubmap2019human} \citep{hubmap2019human} & CODEX & -- & 15 & WSI & Spleen, Lymphnode & Other disease & -- & 80 & 481,156 & -- & -- & CC BY 4.0 \\
\texttt{hu2023single} \citep{hu2023single} & IMC & -- & 12 & TC & Lung, Head and Neck & Lung & -- & 51 & 4,732 & -- & -- & CC BY 4.0 \\
\texttt{jackson2020single} \citep{jackson2020single} & IMC & -- & 357 & TC & Breast & Breast & Age, Sex & 723 & 23,928 & -- & -- & CC BY 4.0 \\
\texttt{klein2026explainable} \citep{klein2026explainable} & IF & -- & 100 & WSI & Pancreas & Other disease & Age$^\dagger$, Sex, BMI & 400 & 316,707 & -- & -- & CC BY-NC 4.0 \\ 
\texttt{lin2023multiplexed} \citep{lin2023multiplexed} & CODEX & -- & 6 & WSI & Colon, Head and Neck & Colorectal & -- & 13 & 34,456 & -- & -- & CC BY 4.0 \\
\texttt{lin2023high} \citep{lin2023high} & Orion, H\&E & registered & 40 & WSI & Colon & Colorectal & Age, Sex & 41 & 623,196 & 41 & 623,196 & CC BY 4.0 \\
\texttt{liu2022reproducible} \citep{liu2022reproducible} & MIBI & -- & 28 & WSI & Lymphnode, Tonsil, Soft Tissue, Bladder & Breast, Colorectal & -- & 164 & 1,312 & -- & -- & CC BY 4.0\\
\texttt{malchiodi2024natural} \citep{malchiodi2024natural} & IMC & -- & 63 & TC & Pancreas & Pancreatic, Other disease & Age, Sex, Race, Ethnicity & 164 & 16,468 & -- & -- & CC BY 4.0 \\
\texttt{meyer2025stratification} \citep{meyer2025stratification} & IMC & -- & 215 & TC & Breast & Breast & Age & 450 & 17,304 & -- & -- & CC BY 4.0 \\
\texttt{moldoveanu2022spatially} \citep{moldoveanu2022spatially} & IMC & -- & 90 & TC & Lymphnode, Skin & Skin (melanoma) & Age, Sex & 90 & 5,616 & -- & -- & CC BY 4.0 \\
\texttt{monkman2024spatial} \citep{monkman2024spatial} & CODEX & - & 42 & TC & Lung & Lung & -- & 42 & 20,916 & - & - & CC BY 4.0 \\
\texttt{phillips2021immune} \citep{phillips2021immune} & CODEX & -- & 14 & TC & Skin & Lymphoma & Age & 70 & 4,804 & -- & -- & CC BY 4.0 \\
\texttt{rigamonti2024integrating} \citep{rigamonti2024integrating} & IMC, H\&E & matched & 108 & TC & Lung & Lung & Age, Sex & 158 & 17,108 & 210 & 24,992 & By author agreement \\
\texttt{ruf2023tumor} \citep{ruf2023tumor} & CODEX & -- & 17 & WSI & Liver & Liver & -- & 17 & 52,190 & -- & -- & CC BY 4.0 \\
\texttt{salie2025spatial} \citep{salie2025spatial} & IMC & -- & 96 & WSI & Liver & Liver & Age, Sex & 194 & 22,708 & -- & -- & CC BY 4.0 \\
\texttt{schulz2018simultaneous} \citep{schulz2018simultaneous} & IMC & -- & 79 & TC & Breast & Breast & -- & 79 & 2,744 & -- & -- & CC BY 4.0 \\
\texttt{schulz2024immucan} \citep{eling2025multi} & IMC & -- & 179 & TC & Lung, Head and Neck & Colorectal, Lung, Head \& neck, Breast & -- & 179 & 6,372 & -- & -- & CC BY 4.0 \\
\texttt{schurch2020coordinated} \citep{schurch2020coordinated} & CODEX, H\&E & registered & 35  & TC & Colon, Other & Colorectal & Age, Sex & 251 & 8,064 & 251 & 7,784 & CC0 1.0 \\
\texttt{xiao2022multiplexed} \citep{xiao2022multiplexed} & IMC & -- & 26 & WSI & Lymphnode, Soft Tissue, Other & Skin (melanoma) & Age$^\dagger$, Sex & 158 & 6,152 & -- & -- & CC BY 4.0 \\
\texttt{zhuhuman2025spatial} \citep{zhu2025spatial} & IMC & -- & 62 & TC & Lung & Lung & -- & 1,649 & 109,372 & -- & -- & CC BY 4.0 \\

\bottomrule
\end{tabular}
\end{adjustbox}
\end{sidewaystable}

\begin{figure}[H] 
  \centering
  \includegraphics[trim=0cm 0cm 30cm 0cm,clip,width=\textwidth, page = 9]{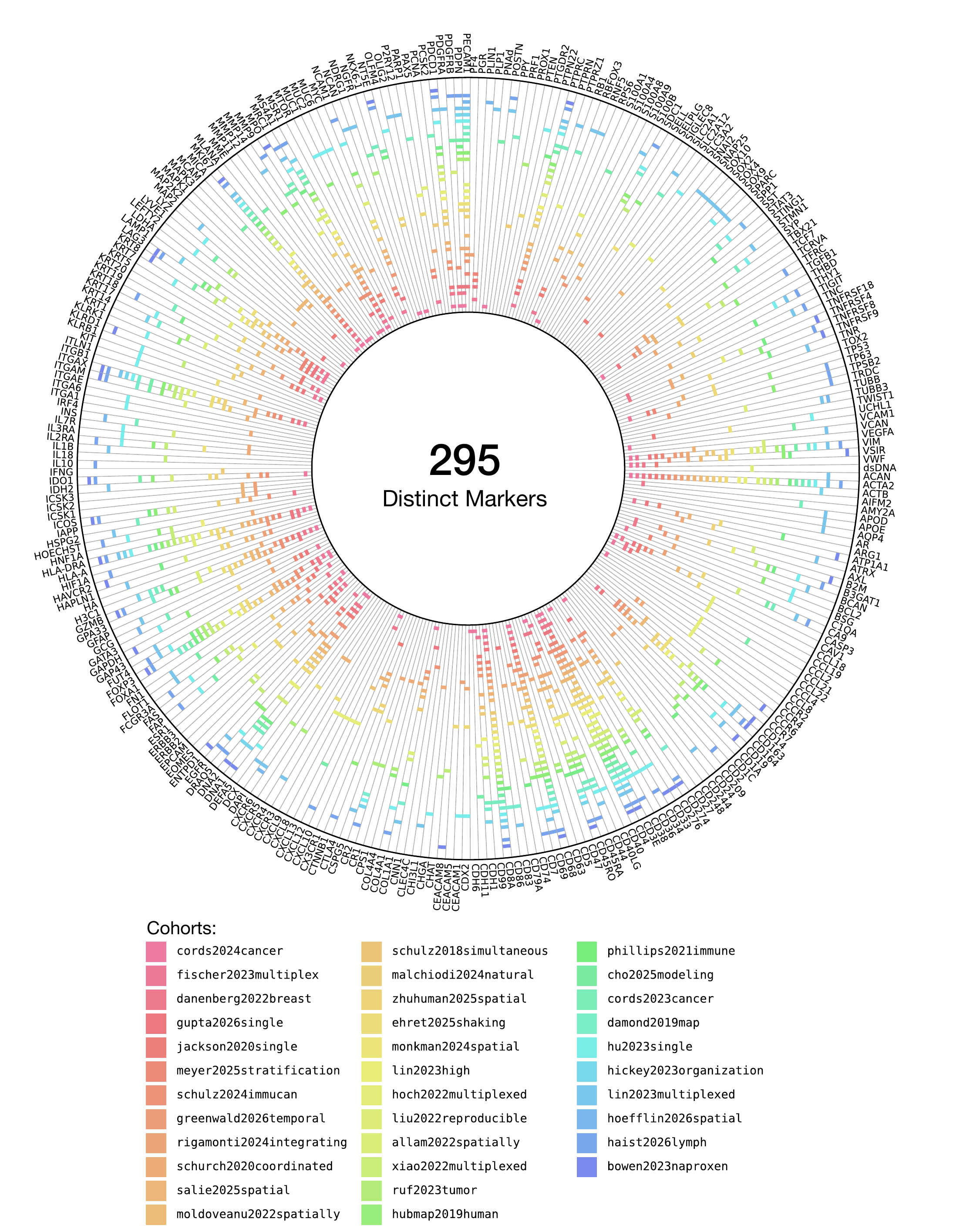}
  \caption{Circle plot indicating, for each protein (HUGO gene nomenclature) and non-protein nuclear marker in the dataset, the cohorts in which it is represented.}
  \label{fig:marker_overview}
\end{figure}

\begin{figure}[H] 
  \centering
  \includegraphics[trim=0cm 12cm 7cm 0cm,clip,width=\textwidth, page =8]{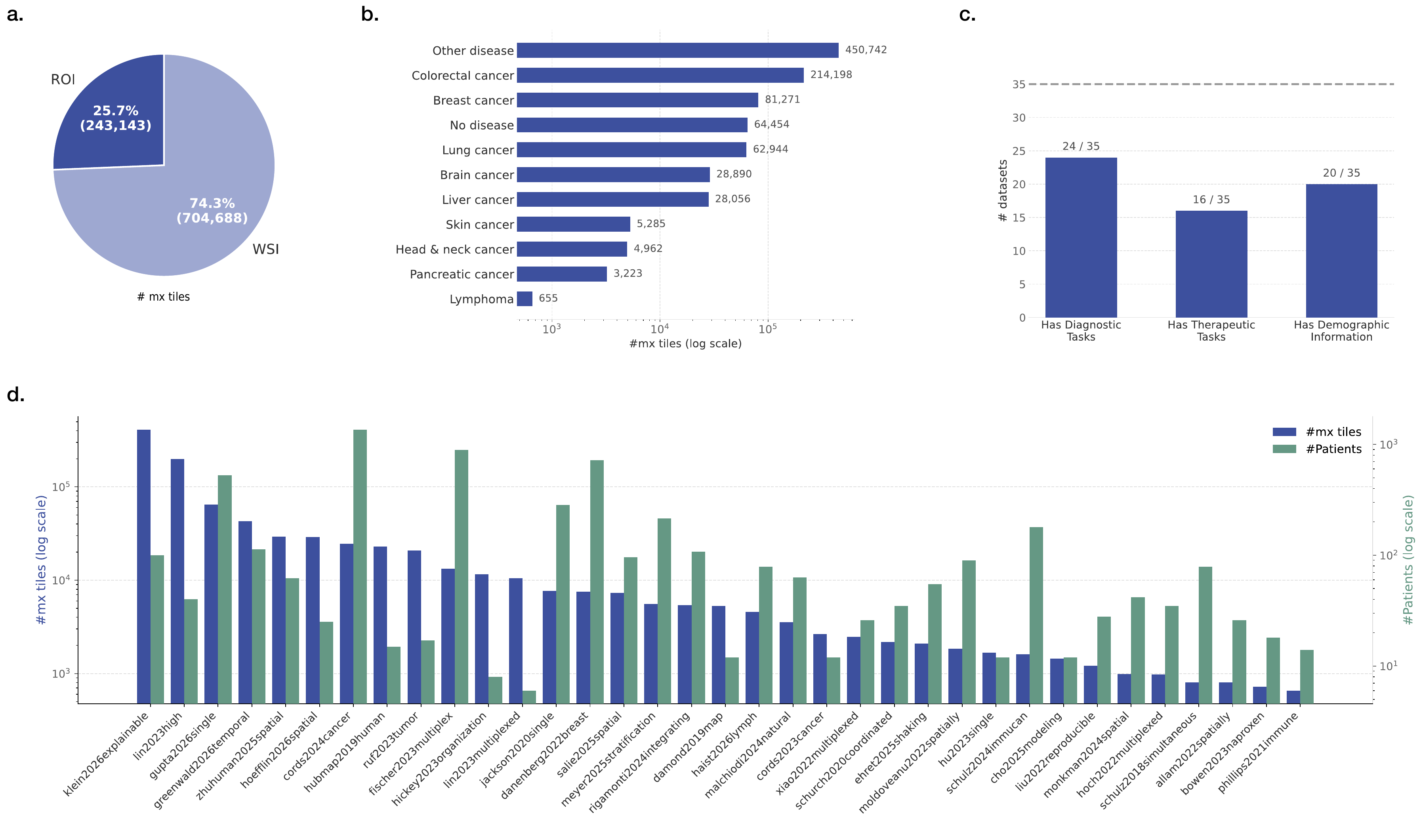}
  \caption{\textbf{Overview of the dataset.}
    \textbf{a.} Distribution of multiplexed tiles across specimen types, split between tumor cores (TC) and whole-slide images.
    \textbf{b.} Number of mx tiles per organ (log scale), spanning 13 tissue types.
    \textbf{c.} Number of datasets containing diagnostic and therapeutic annotations that can be used ofr downstream tasks.
    \textbf{d.} Per-cohort breakdown of mx tiles (blue, left axis) and number of patients (green, right axis), both on a log scale, sorted by decreasing tile count.}
  \label{fig:descriptive_stats}
\end{figure}

\begin{table}[h]
\centering
\caption{Cohorts in \spora with original and derived cell type annotations, and the methods the original authors used to derive them.}
\label{tab:cell_types}
\begin{adjustbox}{max width=\textwidth}
\begin{tabular}{l >{\raggedright\arraybackslash}p{4.2cm} l l}
\toprule
\textbf{Dataset} & \textbf{Original cell-type fields} & \textbf{\makecell[l]{Derived cell-type\\fields}} & \textbf{Method name} \\
\midrule
\texttt{cords2023cancer}          & \texttt{celltype}, \texttt{cell\_subtype}                            & \texttt{cell\_type\_atlas} & PhenoGraph, FlowSOM \\
\texttt{cords2024cancer}          & \texttt{cell\_category}, \texttt{cell\_subtype}, \texttt{cell\_type} & \texttt{cell\_type\_atlas} & PhenoGraph, FlowSOM \\
\texttt{damond2019map}            & \texttt{cell\_category}, \texttt{cell\_type}                         &                            & ilastik \\
\texttt{danenberg2022breast}      & \texttt{cell\_category}, \texttt{cell\_type}, \texttt{cell\_subtype} & \texttt{cell\_type\_atlas} & FlowSOM, PhenoGraph \\
\texttt{gupta2026single}          & \texttt{cluster\_label}                                              &                            & PhenoGraph \\
\texttt{hoch2022multiplexed}      & \texttt{cell\_category}, \texttt{cell\_type}               & \texttt{cell\_type\_atlas} & cytomapper \\
\texttt{jackson2020single}        & \texttt{cell\_category}, \texttt{cell\_type}               & \texttt{cell\_type\_atlas} & PhenoGraph \\
\texttt{lin2023high}              & \texttt{cell\_type}                                                  & \texttt{cell\_type\_atlas} & Custom RFC gating \\
\texttt{liu2022reproducible}      & \texttt{categories}                                                  &                            & FlowSOM \\
\texttt{meyer2025stratification}  & \texttt{metacluster\_combined}, \texttt{metacluster\_tumor\_combined} & \texttt{cell\_type\_atlas} & PhenoGraph, Astir \\
\texttt{moldoveanu2022spatially} & \texttt{cell\_category}                                          & \texttt{cell\_type\_atlas} & PhenoGraph, ilastik \\
\texttt{phillips2021immune}       & \texttt{cell\_category}                                    &                            & X-shift \\
\texttt{rigamonti2024integrating} & \texttt{cell\_category}, \texttt{cell\_type}               & \texttt{cell\_type\_atlas} & CytoMAP, QuPath \\
\texttt{schulz2024immucan}        & \texttt{celltypes}                                                   & \texttt{cell\_type\_atlas} & cytomapper \\
\texttt{schurch2020coordinated}   & \texttt{cell\_category}, \texttt{cell\_type}               &                            & X-shift \\
\texttt{xiao2022multiplexed}      & \texttt{celltype20}, \texttt{celltype4}                              &                            & FlowSOM, PhenoGraph \\
\bottomrule
\end{tabular}
\end{adjustbox}
\end{table}

\section{Example MX Images across Technologies} \label{app:sec:2}

\begin{figure}[H] 
  \centering
  \includegraphics[trim=0cm 9cm 5cm 0cm,clip,width=\textwidth, page = 4]{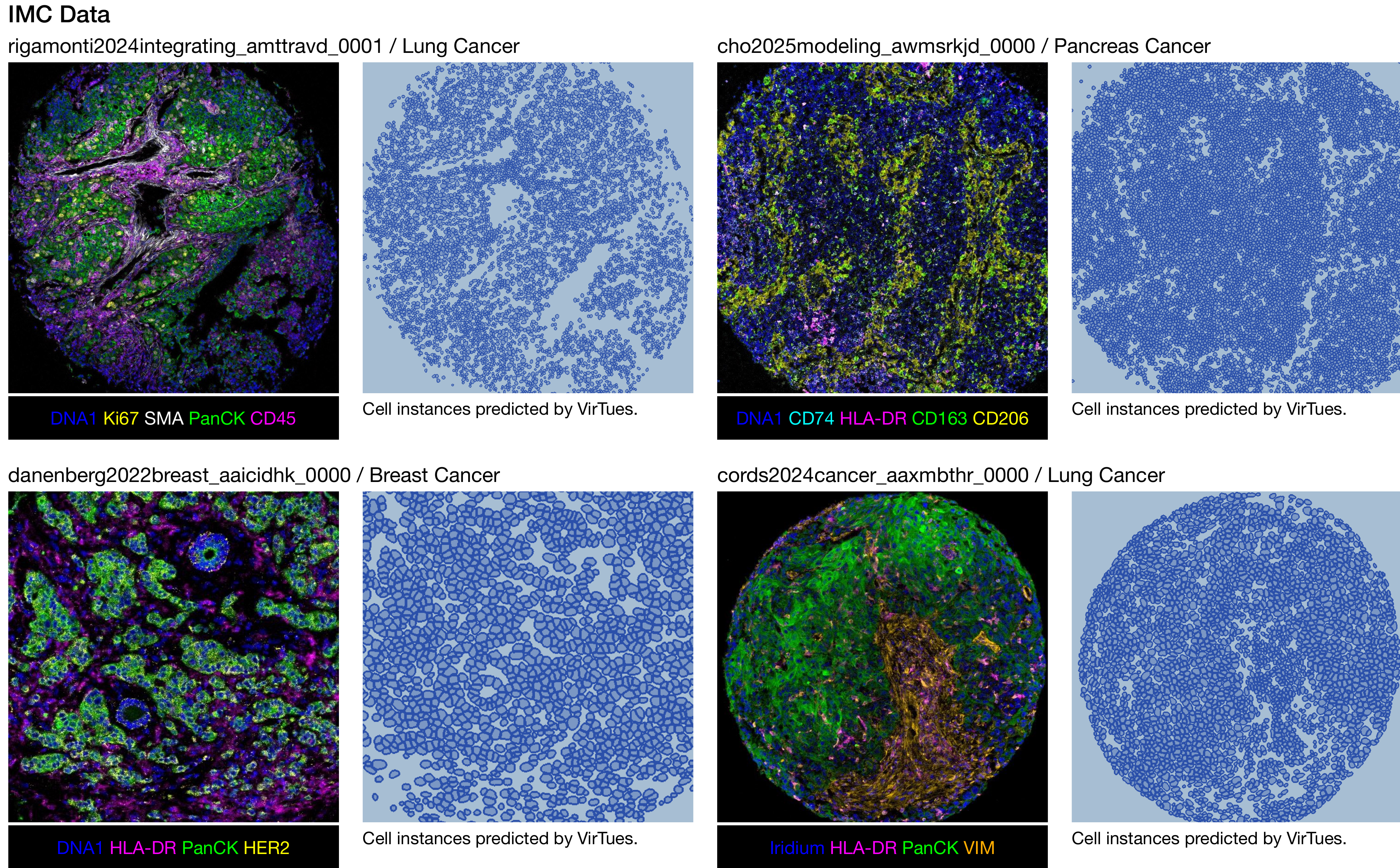}
  \caption{Example mx images for four IMC datasets with instance cell masks.}
  \label{fig:example_mx_imc}
\end{figure}

\begin{figure}[H] 
  \centering
  \includegraphics[trim=0cm 9cm 5cm 0cm,clip,width=\textwidth, page = 5]{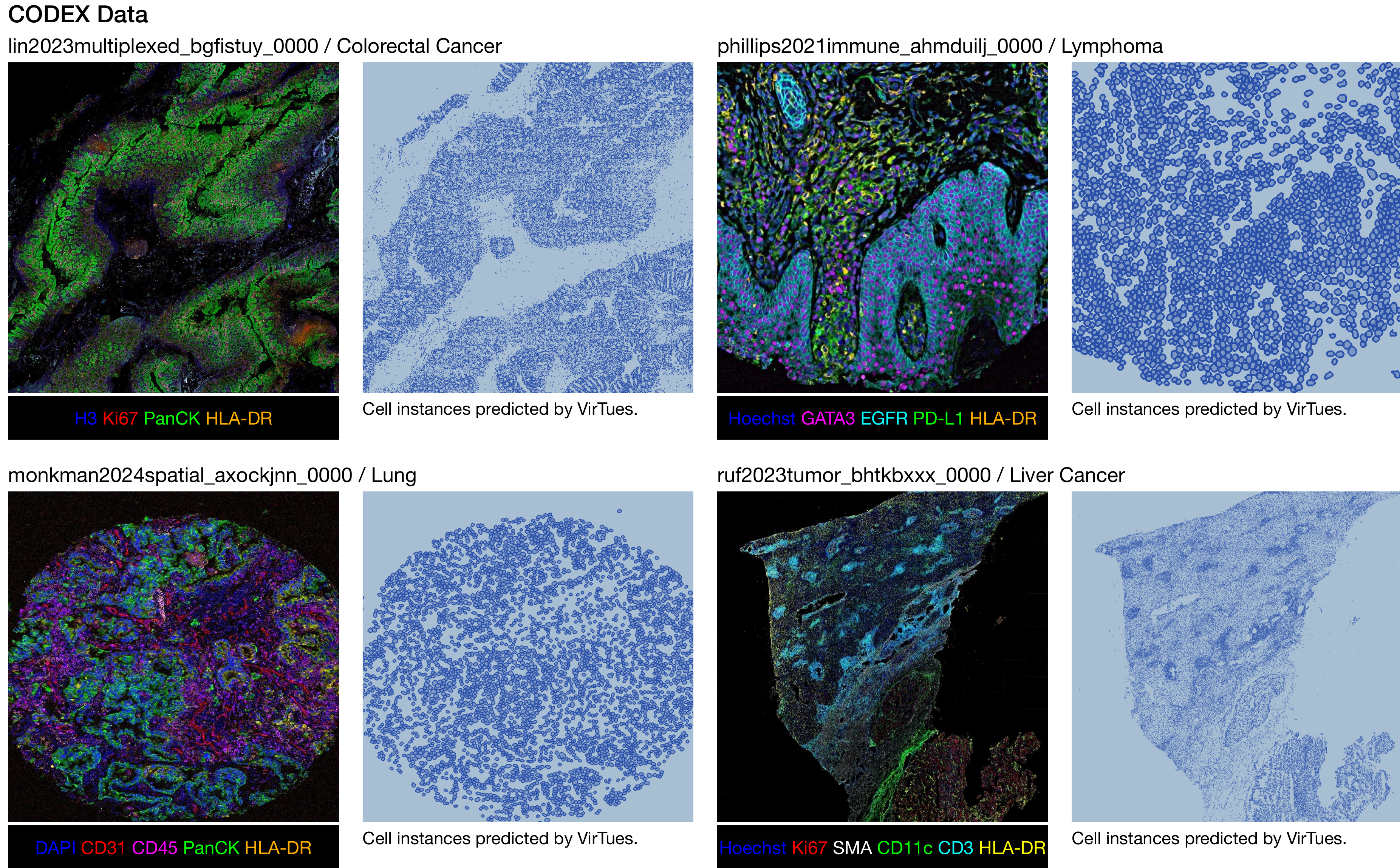}
  \caption{Example mx images for four CODEX datasets with VirTues predicted instance cell masks.}
  \label{fig:example_mx_codex}
\end{figure}

\begin{figure}[H] 
  \centering
  \includegraphics[trim=0cm 28cm 5cm 0cm,clip,width=\textwidth, page = 6]{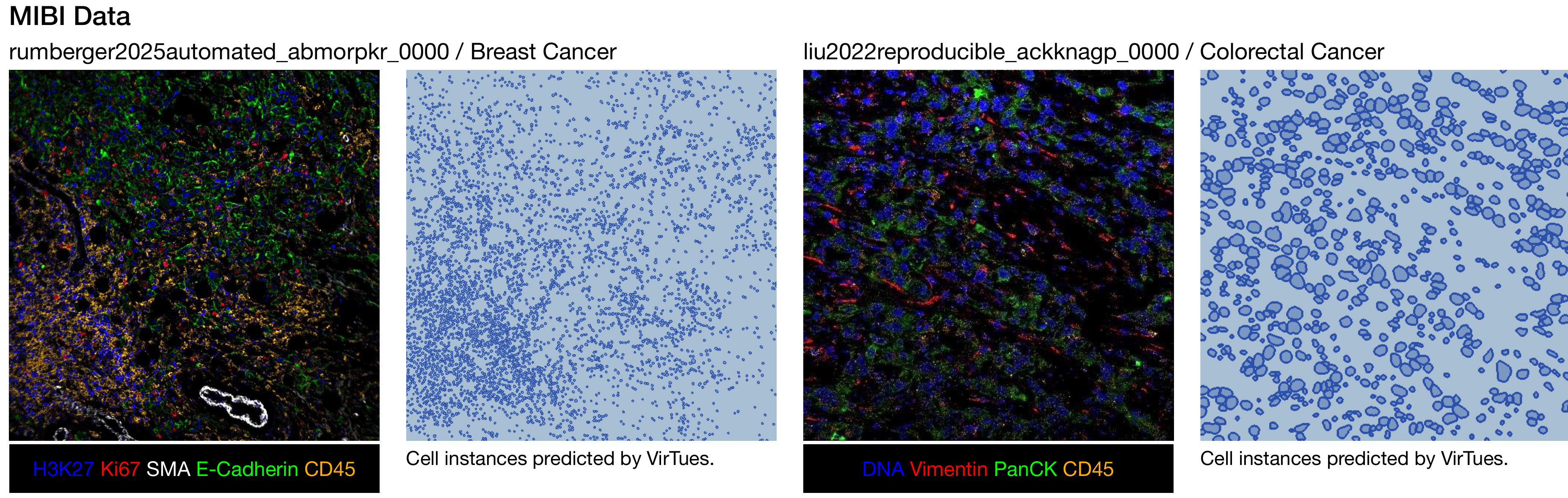}
  \caption{Example mx images for two MIBI datasets with VirTues predicted instance cell masks.}
  \label{fig:example_mx_mibi}
\end{figure}

\begin{figure}[H] 
  \centering
  \includegraphics[trim=0cm 28cm 37cm 0cm,clip,width=0.5\textwidth, page = 7]{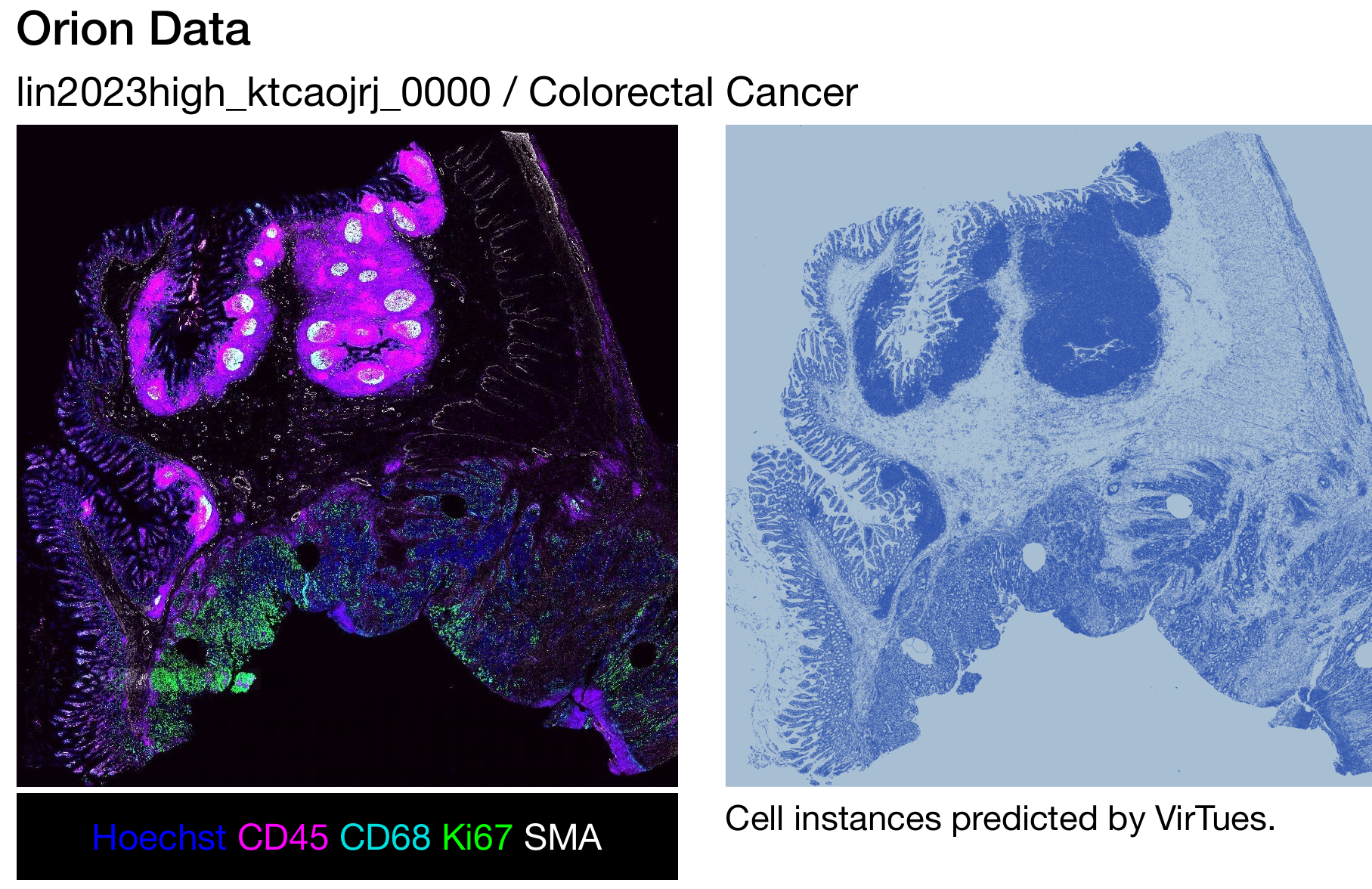}
  \caption{Example mx image for the Orion dataset with VirTues predicted instance cell masks.}
  \label{fig:example_mx_orion}
\end{figure}

\section{Qualitative VirTues Segmentation Evaluation} \label{app:sec:virtues_seg_eval}

\begin{figure}[H] 
  \centering
  \includegraphics[trim=0cm 6cm 2cm 0cm,clip,width=\textwidth, page = 11]{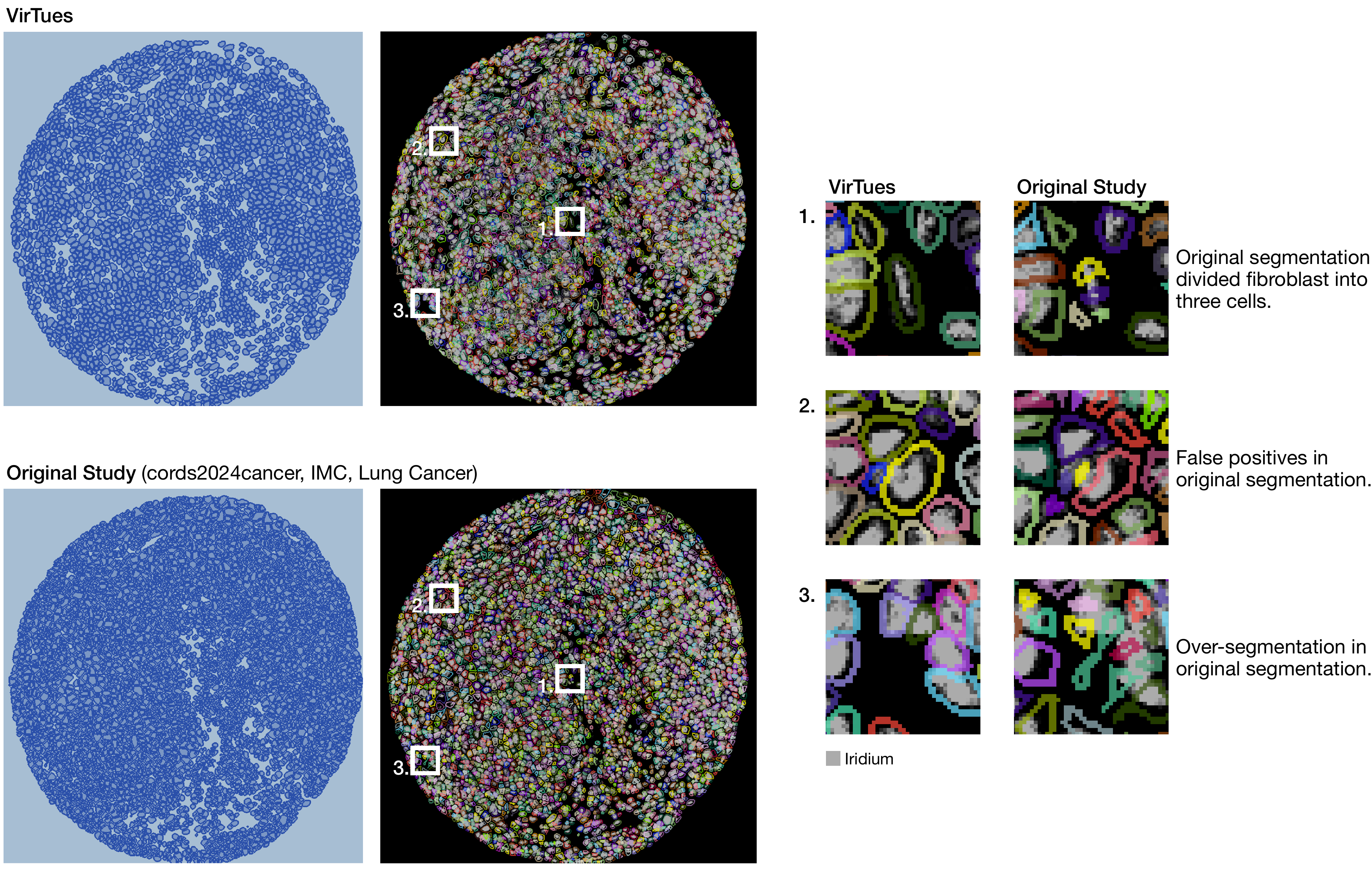}
  \caption{Comparison of cell instance segmentations obtained from VirTues and provided with the original study by \citet{cords2024cancer}.}
  \label{fig:virtues_segmentations_cords}
\end{figure}

\begin{figure}[H] 
  \centering
  \includegraphics[trim=0cm 6cm 2cm 0cm,clip,width=\textwidth, page = 12]{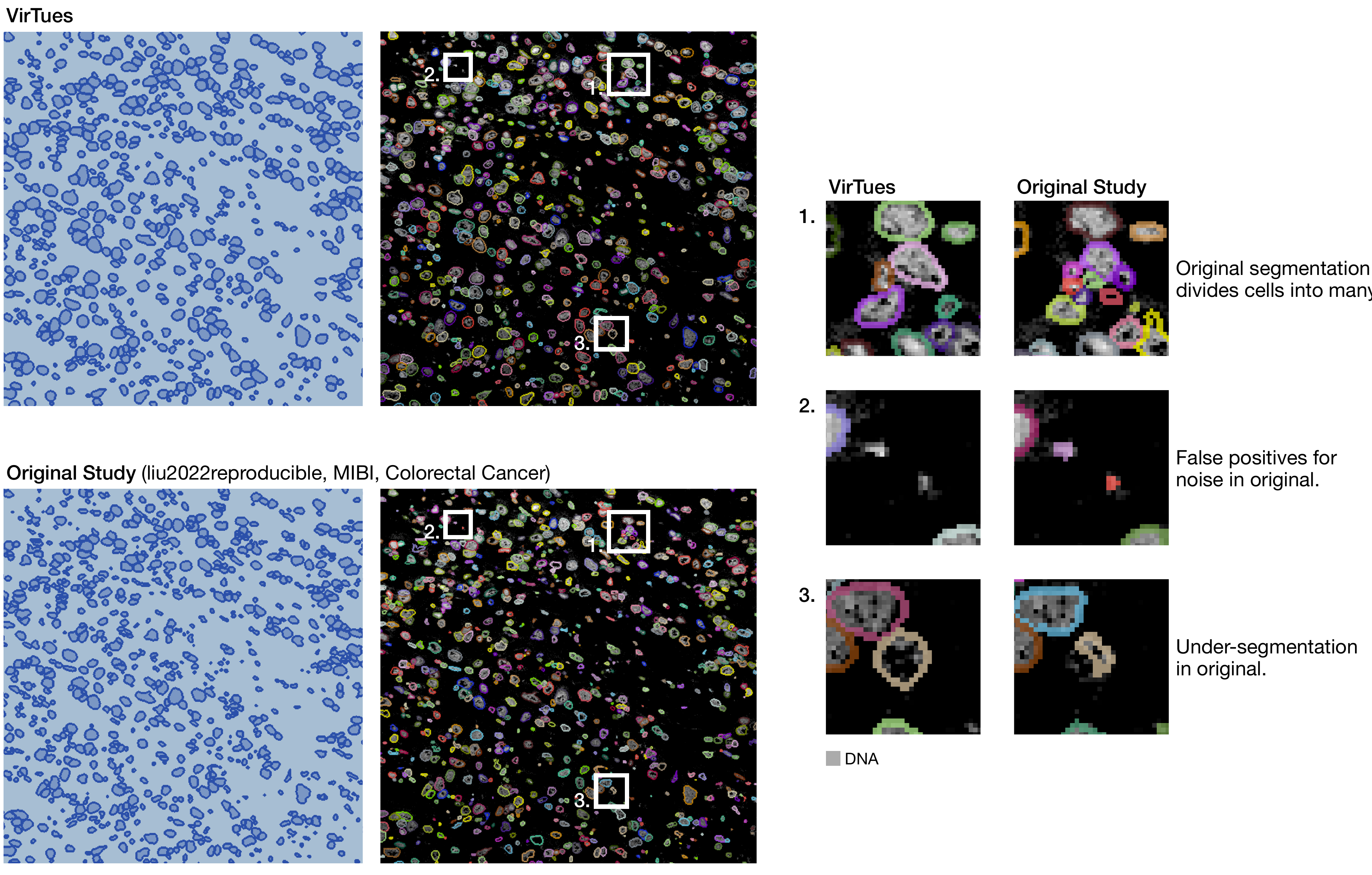}
  \caption{Comparison of cell instance segmentations obtained from VirTues and provided with the original study by \citet{liu2022reproducible}.}
  \label{fig:virtues_segmentations_liu}
\end{figure}

We compared the VirTues foundation model–based segmentations against the classical instance segmentation algorithms used in the original studies and found the VirTues-based segmentations to be preferable in the large majority of cases. 
\figref{fig:virtues_segmentations_cords} and \figref{fig:virtues_segmentations_liu} provide qualitative comparisons across two cohorts that differ in both image acquisition technique and tissue of origin. Neither of the two original studies were included in the training of segmentation model.

For classical cell instance segmentation algorithms such as watershed- or star-convex polygon–based methods (e.g., StarDist), we observe a substantial number of failure modes. 
False positives are common, with single cells frequently fragmented into multiple instances—particularly for elongated morphologies such as fibroblasts. 
We further observe pronounced over- and under-segmentation for cells that deviate from a roughly round shape, as well as spurious segmentation of noise and imaging artifacts, which are especially prevalent in IMC and MIBI acquisitions. 
Under-segmentation arises in large part because these methods rely solely on a nuclear marker channel and do not exploit complementary markers that delineate cell shape and membrane boundaries. 
Conversely, over-segmentation occurs more frequently in CODEX images, where markers tend to appear more diffuse.

Because VirTues is a foundation model trained across diverse tissues and acquisition modalities, it is inherently more robust to noise and to the wide variability of cell morphologies encountered in highly multiplexed imaging. 
Moreover, VirTues leverages the full set of available markers rather than the nuclear channel alone, providing richer evidence for boundary delineation. 
These properties together yield consistently more accurate instance segmentations, which motivated our choice to include additionally VirTues-based segmentation masks in the data corpus.

\section{\sporaio Library}

\subsection{Adaptive Greedy Tiling Algorithm}\label{app:sec:tiling}

The following section details the candidate enumeration, purity filtering, and mask update steps that complete the tiling algorithm summarized in \aref{sec:spora_io}.

\paragraph{Candidate enumeration.}
Given a stride $\Delta$, tile positions are enumerated along each axis as
\begin{equation}
    \Ycal = \{0, \Delta, 2\Delta, \dots\} \cup \{H-c\},
    \qquad
    \Xcal = \{0, \Delta, 2\Delta, \dots\} \cup \{W-c\},
\end{equation}
yielding the candidate grid $\Ccal = \Ycal \times \Xcal$. The boundary positions $H-c$ and $W-c$ are added to ensure that tissue near image edges is not systematically excluded by the stride.

\paragraph{Purity filtering.}
For each $(y,x) \in \Ccal$, let $r(y,x)$ denote the number of valid (foreground) pixels the tile contains. The candidate set passed to the greedy procedure is
\begin{equation}
    \Ccal_\tau = \left\{ (y,x) \in \Ccal : r(y,x)/c^2 \geq 1 - \tau \right\},
\end{equation}
where $\tau \in [0,1)$ is the purity tolerance. Setting $\tau$ small forces tiles to be nearly all tissue; larger $\tau$ admits boundary tiles that may contain biologically informative tissue margins at the cost of background pixels.

\paragraph{Mask update.}
At iteration $t$, after selecting the tile of maximum marginal gain $(y_t, x_t)$, the uncovered mask is updated as
\begin{equation}
    U_{t+1} = U_t \odot \mathbb{I}\{(i,j) \notin T(y_t, x_t)\},
\end{equation}
where $\odot$ denotes element-wise product and $\mathbb{I}\{\cdot\}$ is the indicator over pixel positions. Since $T(y_t, x_t)$ is a contiguous rectangle, the update is performed in $O(c^2)$ time, independent of image size.

\paragraph{Hyperparameters.}
The algorithm exposes four hyperparameters: stride $\Delta$, purity tolerance $\tau$, target coverage $\gamma$, and marginal-gain threshold $\alpha$. Defaults used in \spora are exposed through \sporaio and can be overridden per cohort. The tiles in \spora are generated with: $\Delta=c/2$, $\tau=0.95$, $\gamma=0.95$, and $\alpha=0.1$.

\begin{figure}[H] 
  \centering
  \includegraphics[trim=0cm 10.5cm 7cm 0cm,clip,width=\textwidth, page = 10]{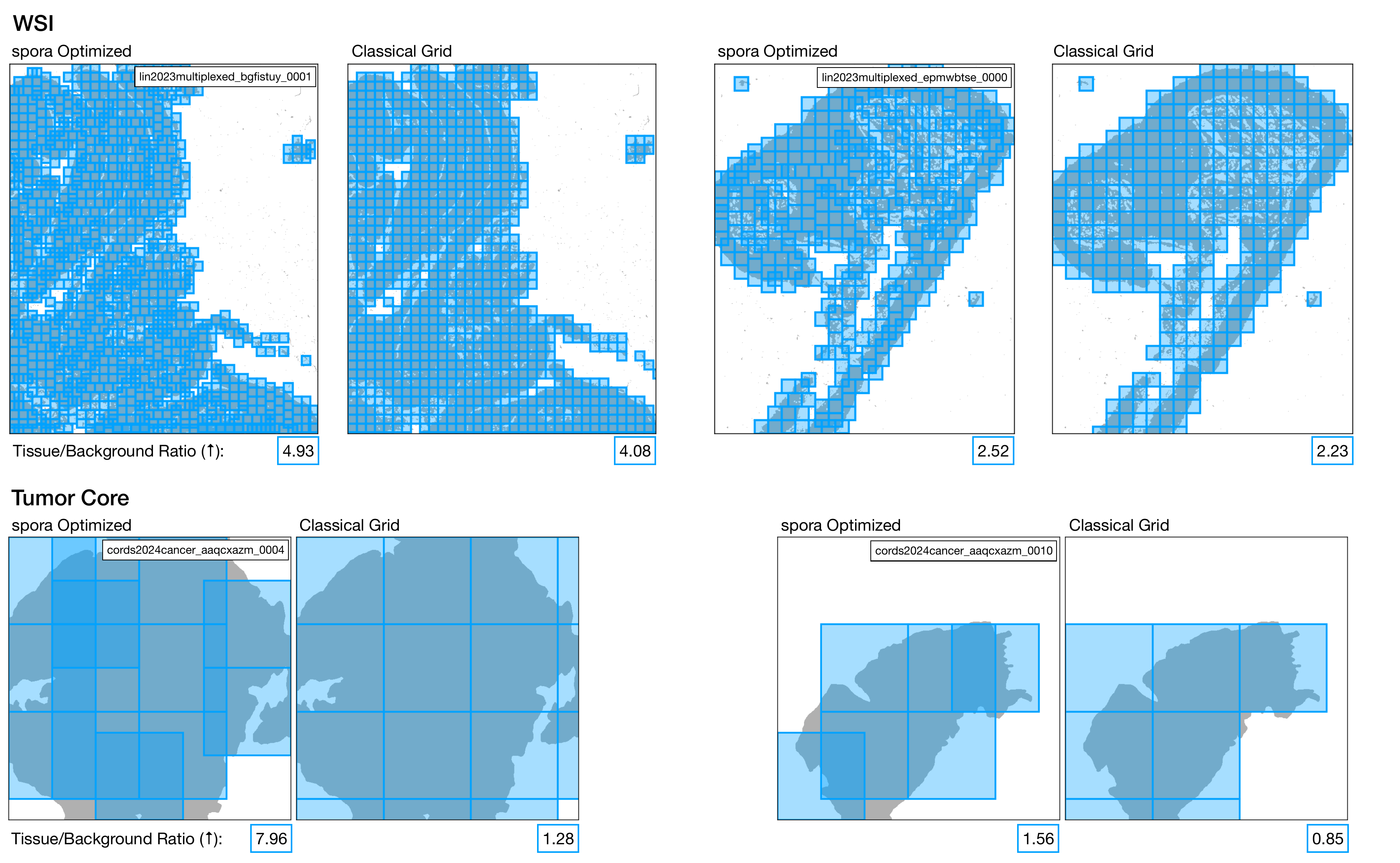}
  \caption{Tile distributions under classical grid tiling ($\geq$10\% tissue inclusion) compared with our coverage-optimized tiling ($\Delta=c/2$, $\tau=0.95$, $\gamma=0.95$, $\alpha=0.1$). Tissue-to-background ratios show improved coverage under our method, particularly near image borders where grid tiling discards partial tiles or includes large out-of-image portions.}
  \label{app:fig:9}
\end{figure}

\subsection{Multiplex Standardization}\label{app:sec:standardization}

The following section details the standardization tables, default transformations, and extension routines summarized in \aref{sec:spora_io}.

\paragraph{Standardization tables.}
For each selected tissue $i$ and channel $c$, \sporaio retrieves an upper quantile $q^\mathrm{up}_{ic}$, lower quantile $q^\mathrm{low}_{ic}$, mean $\mu_{ic}$, and standard deviation $\sigma_{ic}$ from precomputed tables, and performs standardization in an optimized fashion. Channels with invalid standardization metadata are filtered out by a validity mask before preprocessing, so they do not corrupt the input.

\paragraph{Default transformations.}
\sporaio ships two default transformations. The first is quantile clipping followed by optional mean-standard-deviation normalization,
\begin{equation}
    \tilde{x}_{icuv}
    =
    \frac{
    \mathrm{clip}\!\left(x_{icuv}, q^\mathrm{low}_{ic}, q^\mathrm{up}_{ic}\right)
    - q^\mathrm{low}_{ic}}
    {q^\mathrm{up}_{ic} - q^\mathrm{low}_{ic} + \epsilon},
    \qquad
    z_{icuv}
    =
    \frac{\tilde{x}_{icuv} - \mu_{ic}}{\sigma_{ic} + \epsilon}.
\end{equation}
This maps each marker channel to a bounded dynamic range while reducing the influence of rare saturated pixels or acquisition artifacts.

The second is a log-compressed variant,
\begin{equation}
    \tilde{x}_{icuv}
    =
    \log\!\left(
    1 +
    \mathrm{clip}\!\left(x_{icuv}, q^\mathrm{low}_{ic}, q^\mathrm{up}_{ic}\right)
    \right),
    \qquad
    z_{icuv}
    =
    \frac{\tilde{x}_{icuv} - \mu_{ic}}{\sigma_{ic} + \epsilon},
\end{equation}
which is useful for heavy-tailed marker distributions commonly observed in multiplex proteomics.

\paragraph{Extending to new datasets.}
\sporaio includes routines for computing tile coordinates under the optimized tiling strategy (\aref{app:sec:tiling}) and tissue- and dataset-wise standardization statistics, so cohorts beyond those currently in \spora can be integrated under the same protocol.

\clearpage

\begin{figure}[H] 
  \centering
  \includegraphics[trim=0cm 10.5cm 24cm 0cm,clip,width=\textwidth, page = 13]{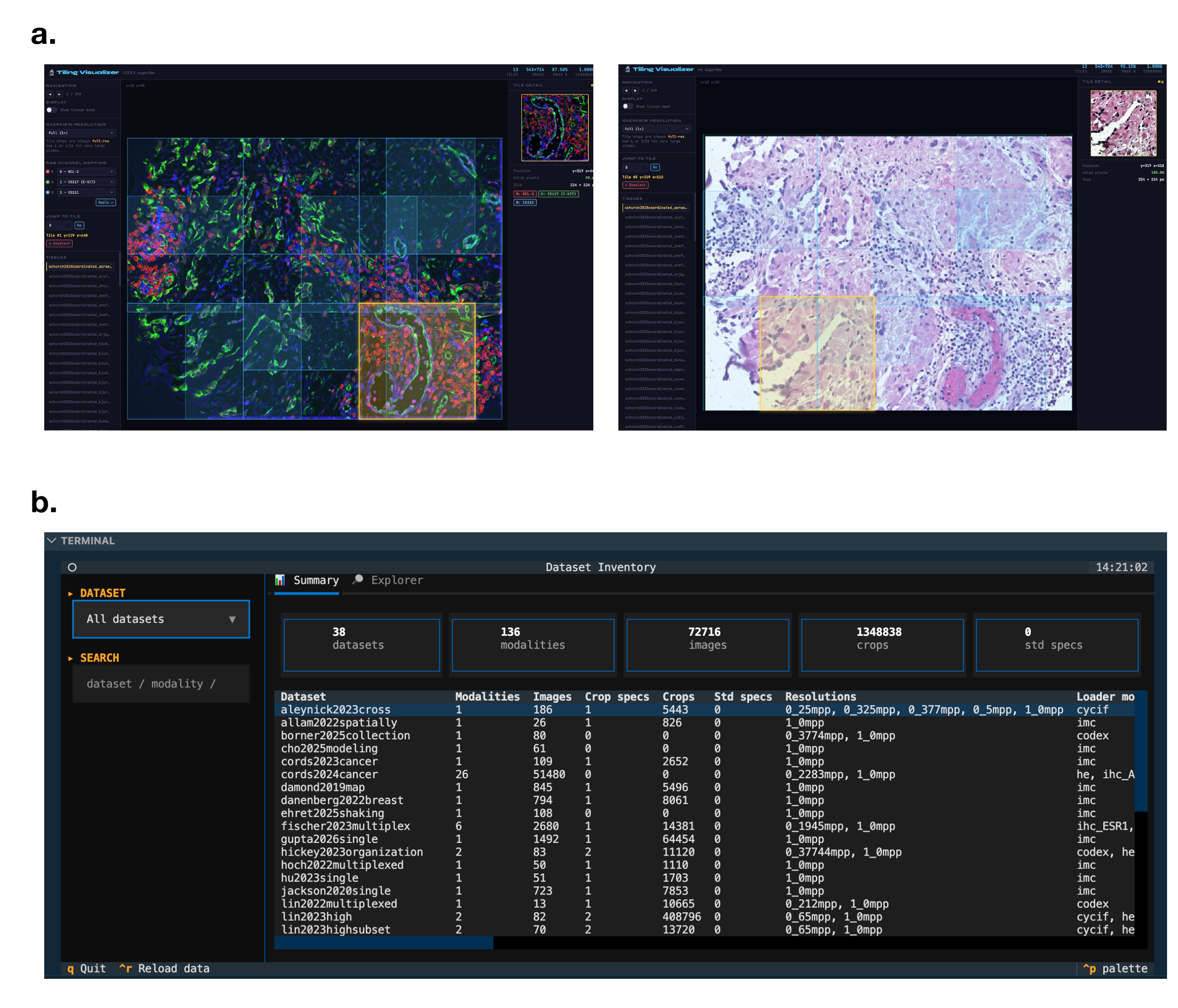}
  \caption{
  \textbf{a.} Tile visualizer tool to compare different tiling hyperparameter visually.
  \textbf{b.} Dataset explorer command line tool that is part of \sporaio to explore and combine different cohorts based on meta data.
  }
  \label{app:fig:10}
\end{figure}

\section{\spora Benchmark} \label{app:sec:spora_benchmark}

\subsection{Foundation Models in Spatial Proteomics}\label{app:sec:sp_fms}
Several foundation models for SP have appeared in recent literature. They differ primarily in how they handle high channel dimensionality and panel heterogeneity across studies.

\paragraph{VirTues \citep{wenckstern2026virtual}} disentangles marker and spatial attention and tokenizes channels with protein language model embeddings, allowing the model to generalize to markers not seen during pretraining. The dual-axis attention design decouples spatial context from marker identity, so arbitrary panels can in principle be processed at inference time. Generalization to unseen markers, however, is mediated by the protein language model and is therefore bounded by its coverage and the quality of its embeddings for less-studied proteins. For all benchmarks, we use the model instance \texttt{virtues-sp32} of VirTues.

\paragraph{KRONOS \citep{shaban2025foundation}} adopts a panel-agnostic single-marker formulation with sinusoidal marker-identity encodings. Its pretraining corpus is partially non-redistributable: 44\% of the data used for training is not publicly available. By treating each channel independently and identifying it through a fixed positional code, KRONOS sidesteps the need for any biological grounding of marker identity, but for the same reason cannot directly transfer semantic information across related markers. The non-public portion of its pretraining data further complicates reproduction and clean head-to-head comparison against models trained on fully open corpora.

\subsection{Disclosure of Pretraining Exposure}

\begin{table}[H]
\centering
\caption{Disclosure of pretraining exposure. We indicate pretraining occurrence for each foundation model and benchmark cohort of \sporabench. We use the model instance \texttt{virtues-sp32} of VirTues. Pretraining data of Eva is not clearly disclosed.}
\label{tab:disclosure_pretraining_exposure}
\begin{tabular}{llll}
\toprule
Model & VirTues & KRONOS & Eva \\
Dataset &  &  &  \\
\midrule
\citet{cords2023cancer} & \checkmark & \ding{55} & ? \\
\citet{cords2024cancer} & \checkmark & \ding{55} & ? \\
\citet{danenberg2022breast} & \checkmark & \ding{55} & ? \\
\citet{eling2025multi} & \checkmark & \ding{55} & ? \\
\citet{hickey2023organization} & \checkmark & \ding{55} & ? \\
\citet{hoch2022multiplexed} & \checkmark & \ding{55} & ? \\
\citet{lin2023high} & \checkmark & \ding{55} & ? \\
\citet{meyer2025stratification} & \checkmark & \ding{55} & ? \\
\citet{moldoveanu2022spatially} & \checkmark & \ding{55} & ? \\
\citet{phillips2021immune} & \checkmark & \ding{55} & ? \\
\citet{rigamonti2024integrating} & \checkmark & \ding{55} & ? \\
\citet{schurch2020coordinated} & \checkmark & \ding{55} & ? \\
\bottomrule
\end{tabular}
\end{table}

\subsection{Details on Cell Phenotyping Benchmarks}

\paragraph{Benchmark procedure.}
For linear probing of the foundation-model-based cell representations, we follow each method's own procedure to compute representations of the benchmark dataset. We discard all cells that are unlabeled or not assigned to a concrete cell phenotype from both training and evaluation. The remaining representations are standardized feature-wise using statistics computed on the train split and used to fit a logistic regression classifier with $\ell_2$ regularization (inverse regularization strength $C = 1.0$), which we evaluate on the test split.

For cross-cohort cell annotation, we likewise follow the annotation procedure of each benchmarked method. In contrast to the linear probe, we place no constraints on the prediction architecture or its training, with one crucial exception: no labels from the target cohort may be used for training, neither those of the train split nor those of the test split. As above, cells without a label or with an unspecific one are excluded from evaluation, and we report classification metrics on the test split of the target benchmark dataset.

\paragraph{Methods and baselines.}

For linear-probing, we compute cell tokens for KRONOS following the proposed approach for cell phenotyping: we extract a $64\times64$ pixel window centered on the cell and, using the instance segmentation mask, set all pixels outside the cell to the marker-wise background value. The final cell representation is the concatenation of the individual marker embeddings. To match the training resolution of KRONOS, we resample all tissue images to 0.5 \textmu m/px before passing them to the model. VirTues-based cell tokens are generated with the authors' original implementation \citep{wenckstern2026virtual}, using a tile size of $128\times128$ pixels, a stride of $42$ pixels and a patch size of $8\times8$ pixels.

For cross-cohort cell annotation, VirTues provides a semantic segmentation head, whose pixel-level predictions we aggregate by majority vote over the ground-truth instance segmentation masks to obtain single-cell predictions. Using the cell type annotations harmonized across all eight benchmark datasets, we train a separate segmentation head for each target dataset in a leave-one-dataset-out fashion, i.e. on the remaining seven datasets, following the training protocol of \citet{wenckstern2026virtual}. KRONOS is omitted from this setting, as it does not provide a procedure for cross-cohort, cross-panel cell annotation.

Beyond spatial proteomics foundation models, we report cell annotation results for two specialized tools: astir \citep{geuenich2021automated}, a probabilistic framework for label-free cell type annotation, and MAPS \citep{shaban2024maps}, an MLP-based classifier that we use for label transfer.

For astir, we curated a list of canonical markers for each annotation task and cell type, validated by biomedical experts. These marker sets are provided as configuration files in the \sporabench repository. We apply astir to arcsinh-transformed mean single-cell marker intensities, $\mathrm{arcsinh}(x / c)$. Unless stated otherwise, we follow the hyperparameter recommendations of the official documentation. The only exception is the arcsinh cofactor $c$, for which a dataset-specific value improved performance: $c = 1280$ for \texttt{lin2023high} \citep{lin2023high} and $c = 1$ for all other datasets. By default, astir predicts two additional categories beyond the target cell types ("Unknown" and "Other"). We disable these to keep the label space comparable across methods. As astir is label-free, it uses no annotations from any cohort.

For MAPS, we select one source cohort per target benchmark cohort, preferring cohorts of the same tissue origin and, among those, the one sharing the largest number of markers with the target. The MLP is trained on normalized mean single-cell intensities of the markers shared between source and target cohort, with batch size $b = 128$, a minimum of $100$ and a maximum of $300$ epochs, and early stopping with a patience of $20$ epochs. The trained model is then applied to the target benchmark cohort for evaluation. The source cohort selected for each target is listed in ~\tabref{tab:maps-cohort-mapping}.

\begin{table}[t]
\centering
\caption{Source cohort selected for training the MAPS label-transfer model for each target benchmark cohort. Source cohorts were chosen to match the tissue origin of the target where possible and to maximize the number of shared markers.}
\label{tab:maps-cohort-mapping}
\begin{tabular}{ll}
\toprule
Target benchmark cohort & Selected source cohort \\
\midrule
\citet{cords2023cancer}          & \citet{meyer2025stratification} \\
\citet{cords2024cancer}          & \citet{rigamonti2024integrating} \\
\citet{danenberg2022breast}      & \citet{meyer2025stratification} \\
\citet{lin2023high}        & \citet{rigamonti2024integrating} \\
\citet{meyer2025stratification}  & \citet{cords2023cancer} \\
\citet{moldoveanu2022spatially}  & \citet{danenberg2022breast} \\
\citet{rigamonti2024integrating} & \citet{cords2024cancer} \\
\citet{eling2025multi}        & \citet{danenberg2022breast} \\
\bottomrule
\end{tabular}
\end{table}

\paragraph{Further results.} We report further benchmark results in \tabref{tab:cross_cohort_cell_annotation_macro_f1}-\tabref{tab:schulz2024immucan_cell_types}. 

\begin{table}[ht]
\centering
\caption{Cross-cohort cell annotation results. Macro average F1-score per dataset and model. Best performing model per dataset is highlighted in bold, the second-best is underlined. The 95\% confidence intervals calculated via bootstrapping ($n=1000$ repeats) are shown in brackets.}
\label{tab:cross_cohort_cell_annotation_macro_f1}
\begin{tabular}{llll}
\toprule
Model & VirTues & astir & MAPS \\
Dataset &  &  &  \\
\midrule
\citet{cords2023cancer} & \textbf{0.717} {\footnotesize [.713, .721]} & 0.532 {\footnotesize [.530, .535]} & \underline{0.614} {\footnotesize [.611, .617]} \\
\citet{cords2024cancer} & \textbf{0.652} {\footnotesize [.650, .653]} & \underline{0.592} {\footnotesize [.591, .594]} & 0.502 {\footnotesize [.500, .503]} \\
\citet{danenberg2022breast} & \textbf{0.553} {\footnotesize [.551, .556]} & \underline{0.407} {\footnotesize [.404, .410]} & 0.390 {\footnotesize [.387, .392]} \\
\citet{lin2023high} & 0.478 {\footnotesize [.475, .480]} & \underline{0.520} {\footnotesize [.518, .521]} & \textbf{0.637} {\footnotesize [.634, .640]} \\
\citet{meyer2025stratification} & \underline{0.605} {\footnotesize [.602, .608]} & 0.596 {\footnotesize [.594, .599]} & \textbf{0.693} {\footnotesize [.691, .696]} \\
\citet{moldoveanu2022spatially} & \textbf{0.611} {\footnotesize [.596, .624]} & \underline{0.561} {\footnotesize [.555, .566]} & 0.314 {\footnotesize [.305, .324]} \\
\citet{rigamonti2024integrating} & \underline{0.605} {\footnotesize [.600, .609]} & \textbf{0.623} {\footnotesize [.619, .626]} & 0.497 {\footnotesize [.494, .501]} \\
\citet{eling2025multi} & \underline{0.620} {\footnotesize [.617, .624]} & \textbf{0.621} {\footnotesize [.618, .624]} & 0.455 {\footnotesize [.451, .459]} \\
\bottomrule
\end{tabular}
\end{table}

\begin{table}[h!]
\centering
\caption{Detailed cell phenotyping results for \texttt{cords2023cancer}~\citep{cords2023cancer}. F1-scores and bootstrapped 95\% confidence intervals per cell phenotype are shown.}
\label{tab:cords2023cancer_cell_types}
\begin{tabular}{lll}
\toprule
Model & KRONOS & VirTues \\
Class &  &  \\
\midrule
CAF & 0.627 {\footnotesize [.622, .632]} & \textbf{0.791} {\footnotesize [.788, .794]} \\
Endothelial & 0.420 {\footnotesize [.408, .433]} & \textbf{0.585} {\footnotesize [.573, .596]} \\
Immune & 0.591 {\footnotesize [.587, .596]} & \textbf{0.819} {\footnotesize [.816, .822]} \\
Tumour & 0.765 {\footnotesize [.762, .768]} & \textbf{0.932} {\footnotesize [.930, .933]} \\
\bottomrule
\end{tabular}
\end{table}

\begin{table}[h!]
\centering
\caption{Detailed cell phenotyping results for \texttt{cords2024cancer}~\citep{cords2024cancer}. F1-scores and bootstrapped 95\% confidence intervals per cell phenotype are shown. Results for KRONOS are omitted because the high dimensionality of its cell embeddings, combined with the large number of cells, made logistic regression computationally infeasible.}
\label{tab:cords2024cancer_cell_types}
\begin{tabular}{lll}
\toprule
Model & KRONOS & VirTues \\
Class &  &  \\
\midrule
Fibroblast & - & \textbf{0.780} {\footnotesize [.778, .781]} \\
Immune cell & - & \textbf{0.739} {\footnotesize [.738, .741]} \\
T cell & - & \textbf{0.772} {\footnotesize [.770, .775]} \\
Tumor cell & - & \textbf{0.943} {\footnotesize [.942, .943]} \\
Vascular cell & - & \textbf{0.549} {\footnotesize [.543, .555]} \\
\bottomrule
\end{tabular}
\end{table}

\begin{table}[h!]
\centering
\caption{Detailed cell phenotyping results for \texttt{danenberg2022breast}~\citep{danenberg2022breast}. F1-scores and bootstrapped 95\% confidence intervals per cell phenotype are shown.}
\label{tab:danenberg2022breast_cell_types}
\begin{tabular}{lll}
\toprule
Model & KRONOS & VirTues \\
Class &  &  \\
\midrule
Antigen-presenting cell & 0.099 {\footnotesize [.081, .117]} & \textbf{0.251} {\footnotesize [.230, .272]} \\
B/plasma cell & 0.427 {\footnotesize [.417, .437]} & \textbf{0.798} {\footnotesize [.792, .804]} \\
ER- epithelial cell & 0.512 {\footnotesize [.510, .515]} & \textbf{0.674} {\footnotesize [.671, .677]} \\
ER+ epithelial cell & 0.342 {\footnotesize [.337, .347]} & \textbf{0.582} {\footnotesize [.578, .586]} \\
Myeloid cell & 0.348 {\footnotesize [.341, .354]} & \textbf{0.596} {\footnotesize [.590, .602]} \\
NK cell & 0.023 {\footnotesize [.016, .030]} & \textbf{0.487} {\footnotesize [.470, .502]} \\
Stromal cell & 0.645 {\footnotesize [.642, .649]} & \textbf{0.834} {\footnotesize [.832, .836]} \\
T cell & 0.444 {\footnotesize [.436, .452]} & \textbf{0.665} {\footnotesize [.658, .671]} \\
\bottomrule
\end{tabular}
\end{table}

\begin{table}[h!]
\centering
\caption{Detailed cell phenotyping results for \texttt{hoch2022multiplexed}~\citep{hoch2022multiplexed}. F1-scores and bootstrapped 95\% confidence intervals per cell phenotype are shown.}
\label{tab:hoch2022multiplexed_cell_types}
\begin{tabular}{lll}
\toprule
Model & KRONOS & VirTues \\
Class &  &  \\
\midrule
Lymphocyte & 0.500 {\footnotesize [.467, .530]} & \textbf{0.838} {\footnotesize [.820, .856]} \\
Macrophage & 0.643 {\footnotesize [.630, .656]} & \textbf{0.829} {\footnotesize [.819, .837]} \\
Stromal cell & 0.325 {\footnotesize [.309, .341]} & \textbf{0.626} {\footnotesize [.613, .639]} \\
T cell & 0.618 {\footnotesize [.609, .627]} & \textbf{0.949} {\footnotesize [.945, .952]} \\
Tumor cell & 0.785 {\footnotesize [.781, .788]} & \textbf{0.937} {\footnotesize [.935, .939]} \\
\bottomrule
\end{tabular}
\end{table}

\begin{table}[h!]
\centering
\caption{Detailed cell phenotyping results for \texttt{lin2023highsubset}~\citep{lin2023high}. F1-scores and bootstrapped 95\% confidence intervals per cell phenotype are shown.}
\label{tab:lin2023highsubset_cell_types}
\begin{tabular}{lll}
\toprule
Model & KRONOS & VirTues \\
Class &  &  \\
\midrule
B cell & 0.557 {\footnotesize [.545, .568]} & \textbf{0.762} {\footnotesize [.753, .770]} \\
CD8 T cell & 0.495 {\footnotesize [.485, .504]} & \textbf{0.672} {\footnotesize [.664, .679]} \\
Endothelial cell & 0.396 {\footnotesize [.387, .405]} & \textbf{0.815} {\footnotesize [.809, .820]} \\
Exhausted CD8 T cell & 0.067 {\footnotesize [.058, .076]} & \textbf{0.309} {\footnotesize [.275, .343]} \\
M2-like macrophage & 0.483 {\footnotesize [.476, .490]} & \textbf{0.704} {\footnotesize [.698, .710]} \\
Macrophage & 0.523 {\footnotesize [.519, .526]} & \textbf{0.668} {\footnotesize [.665, .671]} \\
Memory T cell & 0.207 {\footnotesize [.198, .216]} & \textbf{0.547} {\footnotesize [.534, .561]} \\
PD-L1+ tumor cell & 0.195 {\footnotesize [.181, .210]} & \textbf{0.275} {\footnotesize [.260, .290]} \\
Stroma & 0.358 {\footnotesize [.355, .361]} & \textbf{0.788} {\footnotesize [.786, .790]} \\
T cell & 0.145 {\footnotesize [.141, .149]} & \textbf{0.433} {\footnotesize [.423, .442]} \\
CD4 T cell & 0.486 {\footnotesize [.478, .495]} & \textbf{0.670} {\footnotesize [.663, .677]} \\
Regulatory T cell & 0.553 {\footnotesize [.540, .566]} & \textbf{0.650} {\footnotesize [.638, .661]} \\
Tumor & 0.759 {\footnotesize [.758, .761]} & \textbf{0.958} {\footnotesize [.957, .958]} \\
Vascular smooth muscle & 0.373 {\footnotesize [.367, .379]} & \textbf{0.735} {\footnotesize [.731, .740]} \\
\bottomrule
\end{tabular}
\end{table}

\begin{table}[h!]
\centering
\caption{Detailed cell phenotyping results for \texttt{meyer2025stratification}~\citep{meyer2025stratification}. F1-scores and bootstrapped 95\% confidence intervals per cell phenotype are shown.}
\label{tab:meyer2025stratification_cell_types}
\begin{tabular}{lll}
\toprule
Model & KRONOS & VirTues \\
Class &  &  \\
\midrule
B cell & 0.541 {\footnotesize [.530, .553]} & \textbf{0.851} {\footnotesize [.845, .858]} \\
CD4 T cell & 0.557 {\footnotesize [.548, .566]} & \textbf{0.834} {\footnotesize [.828, .839]} \\
CD8 T cell & 0.596 {\footnotesize [.587, .605]} & \textbf{0.831} {\footnotesize [.825, .836]} \\
Endothelial cell & 0.435 {\footnotesize [.427, .444]} & \textbf{0.758} {\footnotesize [.752, .763]} \\
HLA-DR+ cell & 0.107 {\footnotesize [.090, .125]} & \textbf{0.312} {\footnotesize [.289, .333]} \\
Immune-low cell & 0.370 {\footnotesize [.350, .389]} & \textbf{0.598} {\footnotesize [.581, .616]} \\
Monocyte & 0.511 {\footnotesize [.504, .518]} & \textbf{0.761} {\footnotesize [.756, .765]} \\
Neutrophil & 0.280 {\footnotesize [.262, .297]} & \textbf{0.476} {\footnotesize [.457, .494]} \\
Stromal cell & 0.578 {\footnotesize [.572, .585]} & \textbf{0.789} {\footnotesize [.784, .793]} \\
Tumor cell & 0.821 {\footnotesize [.819, .822]} & \textbf{0.968} {\footnotesize [.967, .968]} \\
\bottomrule
\end{tabular}
\end{table}

\begin{table}[h!]
\centering
\caption{Detailed cell phenotyping results for \texttt{moldoveanu2022spatially}~\citep{moldoveanu2022spatially}. F1-scores and bootstrapped 95\% confidence intervals per cell phenotype are shown.}
\label{tab:moldoveanu2022spatially_cell_types}
\begin{tabular}{lll}
\toprule
Model & KRONOS & VirTues \\
Class &  &  \\
\midrule
B cell & 0.218 {\footnotesize [.154, .282]} & \textbf{0.626} {\footnotesize [.556, .691]} \\
Endothelial cell & 0.750 {\footnotesize [.727, .772]} & \textbf{0.940} {\footnotesize [.931, .950]} \\
Activated CD8 T cell & 0.625 {\footnotesize [.596, .654]} & \textbf{0.798} {\footnotesize [.778, .818]} \\
Naive CD8 T cell & 0.347 {\footnotesize [.296, .394]} & \textbf{0.525} {\footnotesize [.474, .576]} \\
Activated CD4 T cell & 0.519 {\footnotesize [.475, .565]} & \textbf{0.715} {\footnotesize [.677, .750]} \\
Naive CD4 T cell & 0.279 {\footnotesize [.211, .349]} & \textbf{0.338} {\footnotesize [.257, .423]} \\
Regulatory T cell & 0.412 {\footnotesize [.345, .476]} & \textbf{0.662} {\footnotesize [.604, .719]} \\
Macrophage/monocyte & 0.648 {\footnotesize [.638, .658]} & \textbf{0.868} {\footnotesize [.862, .875]} \\
Melanocyte & 0.911 {\footnotesize [.909, .914]} & \textbf{0.974} {\footnotesize [.972, .975]} \\
\bottomrule
\end{tabular}
\end{table}

\begin{table}[h!]
\centering
\caption{Detailed cell phenotyping results for \texttt{phillips2021immune}~\citep{phillips2021immune}. F1-scores and bootstrapped 95\% confidence intervals per cell phenotype are shown.}
\label{tab:phillips2021immune_cell_types}
\begin{tabular}{lll}
\toprule
Model & KRONOS & VirTues \\
Class &  &  \\
\midrule
B cell & 0.330 {\footnotesize [.284, .382]} & \textbf{0.720} {\footnotesize [.685, .754]} \\
Epithelial cell & 0.602 {\footnotesize [.590, .614]} & \textbf{0.945} {\footnotesize [.941, .949]} \\
Langerhans cell & 0.210 {\footnotesize [.152, .266]} & \textbf{0.328} {\footnotesize [.262, .395]} \\
Mast cell & 0.553 {\footnotesize [.456, .639]} & \textbf{0.754} {\footnotesize [.692, .813]} \\
Myeloid cell & 0.464 {\footnotesize [.448, .480]} & \textbf{0.757} {\footnotesize [.747, .768]} \\
Nerve cell & 0.230 {\footnotesize [.138, .317]} & \textbf{0.621} {\footnotesize [.528, .714]} \\
Neutrophil & 0.502 {\footnotesize [.388, .607]} & \textbf{0.826} {\footnotesize [.757, .883]} \\
Stromal cell & 0.475 {\footnotesize [.455, .494]} & \textbf{0.716} {\footnotesize [.701, .730]} \\
T cell & 0.286 {\footnotesize [.265, .306]} & \textbf{0.532} {\footnotesize [.515, .549]} \\
Tumor & 0.681 {\footnotesize [.675, .687]} & \textbf{0.841} {\footnotesize [.836, .846]} \\
Vascular cell & 0.475 {\footnotesize [.450, .502]} & \textbf{0.826} {\footnotesize [.808, .841]} \\
\bottomrule
\end{tabular}
\end{table}

\begin{table}[h!]
\centering
\caption{Detailed cell phenotyping results for \texttt{rigamonti2024integrating}~\citep{rigamonti2024integrating}. F1-scores and bootstrapped 95\% confidence intervals per cell phenotype are shown.}
\label{tab:rigamonti2024integrating_cell_types}
\begin{tabular}{lll}
\toprule
Model & KRONOS & VirTues \\
Class &  &  \\
\midrule
B cell & 0.666 {\footnotesize [.648, .683]} & \textbf{0.795} {\footnotesize [.782, .807]} \\
CD4 T cell & 0.572 {\footnotesize [.556, .588]} & \textbf{0.712} {\footnotesize [.700, .725]} \\
CD8 T cell & 0.673 {\footnotesize [.662, .684]} & \textbf{0.807} {\footnotesize [.798, .816]} \\
Ki67+ cell & 0.114 {\footnotesize [.057, .180]} & \textbf{0.224} {\footnotesize [.150, .295]} \\
Macrophage & 0.613 {\footnotesize [.602, .622]} & \textbf{0.728} {\footnotesize [.719, .736]} \\
Regulatory T cell & 0.472 {\footnotesize [.440, .500]} & \textbf{0.614} {\footnotesize [.588, .639]} \\
Tumor cell & 0.926 {\footnotesize [.925, .928]} & \textbf{0.966} {\footnotesize [.965, .967]} \\
Vimentin+ cell & 0.390 {\footnotesize [.375, .404]} & \textbf{0.573} {\footnotesize [.561, .585]} \\
aSMA+ cell & 0.615 {\footnotesize [.604, .625]} & \textbf{0.760} {\footnotesize [.752, .768]} \\
Vessel & 0.729 {\footnotesize [.720, .737]} & \textbf{0.810} {\footnotesize [.803, .818]} \\
\bottomrule
\end{tabular}
\end{table}

\begin{table}[h!]
\centering
\caption{Detailed cell phenotyping results for \texttt{schulz2024immucan}~\citep{eling2025multi}. F1-scores and bootstrapped 95\% confidence intervals per cell phenotype are shown.}
\label{tab:schulz2024immucan_cell_types}
\begin{tabular}{lll}
\toprule
Model & KRONOS & VirTues \\
Class &  &  \\
\midrule
B cell & 0.546 {\footnotesize [.527, .563]} & \textbf{0.835} {\footnotesize [.824, .846]} \\
CD4 T cell & 0.492 {\footnotesize [.470, .514]} & \textbf{0.766} {\footnotesize [.751, .781]} \\
CD8 T cell & 0.570 {\footnotesize [.554, .585]} & \textbf{0.784} {\footnotesize [.775, .795]} \\
Ki67+ cell & 0.630 {\footnotesize [.619, .641]} & \textbf{0.873} {\footnotesize [.866, .879]} \\
Dendritic cell & 0.346 {\footnotesize [.329, .366]} & \textbf{0.519} {\footnotesize [.501, .537]} \\
HLA-DR+ cell & 0.322 {\footnotesize [.295, .347]} & \textbf{0.413} {\footnotesize [.386, .438]} \\
CD163+ macrophage & 0.543 {\footnotesize [.529, .555]} & \textbf{0.735} {\footnotesize [.726, .744]} \\
Mural cell & 0.648 {\footnotesize [.637, .659]} & \textbf{0.886} {\footnotesize [.880, .892]} \\
NK cell & 0.454 {\footnotesize [.396, .506]} & \textbf{0.614} {\footnotesize [.564, .658]} \\
Neutrophil & 0.595 {\footnotesize [.579, .610]} & \textbf{0.919} {\footnotesize [.912, .925]} \\
Treg & 0.657 {\footnotesize [.635, .678]} & \textbf{0.884} {\footnotesize [.871, .897]} \\
Tumor cell & 0.771 {\footnotesize [.768, .774]} & \textbf{0.948} {\footnotesize [.947, .950]} \\
Plasmacytoid DC & 0.517 {\footnotesize [.478, .558]} & \textbf{0.627} {\footnotesize [.591, .663]} \\
Plasma cell & 0.554 {\footnotesize [.536, .570]} & \textbf{0.858} {\footnotesize [.849, .867]} \\
\bottomrule
\end{tabular}
\end{table}

\clearpage

\subsection{Details on Instance Segmentation Benchmarks}

\paragraph{Benchmark procedure.}
The instance segmentation benchmark is evaluated zero-shot, i.e., no method may be trained on segmentation masks of the target benchmark cohort. For each tissue image in the test split, we predict a cell instance segmentation mask following the method-specific procedure. To compute the instance F1-score at an IoU threshold $\tau$, predicted and ground-truth instances are matched as follows. We compute the IoU for every pair of predicted and ground-truth instances, discard all pairs with $\mathrm{IoU} < \tau$, and determine a maximum-weight bipartite matching on the remaining pairs, with the IoU as edge weight. Matched pairs are counted as true positives, unmatched predictions as false positives and unmatched ground-truth instances as false negatives. For large tissues, matchings and corresponding counts are calculated chunk-wise using tiles of size $1024\times1024$ pixels. Finally, we compute precision, recall and the instance F1-scores by pooling counts across tiles and tissues.

\paragraph{Methods and baselines.}
VirTues provides an instance segmentation head. To satisfy the zero-shot constraint of the benchmark, we train a separate head for each of the eight benchmark datasets in a leave-one-dataset-out fashion, i.e. on the remaining seven datasets, following the training protocol of \citet{wenckstern2026virtual}. The resulting head is then applied to the held-out target dataset without any further adaptation. 

In addition to VirTues, we benchmark three widely used, general-purpose cell segmentation tools. All three are applied with their publicly available pretrained weights and without any fine-tuning, so the zero-shot constraint is satisfied by construction. Since these methods expect one or two grayscale channels rather than a full protein panel, we curate a panel-independent mapping from UniProt identifiers to nuclear and cytoplasmic/membrane markers; the marker lists are provided as configuration files in the \sporabench repository. 

We use the pretrained \texttt{cpsam\_v2} model of Cellpose~\citep{pachitariu2025cellpose} in a two-channel configuration. The cytoplasm channel is formed by normalizing each cytoplasmic and membrane marker to its $1$st--$99$th intensity percentile, clipping to $[0, 1]$ and averaging the resulting channels; the nuclear channel is given by the nuclear marker (Histone H3). For images without a nuclear marker in the panel, we fall back to single-channel cytoplasm segmentation. All remaining parameters are left at their defaults.

For StarDist~\citep{weigert2022nuclei} we utilize  the pretrained \texttt{2D\_versatile\_fluo} model instance, applied to the percentile-normalized nuclear marker channel (Histone H3) with default prediction parameters. As StarDist operates on nuclei only, datasets whose panel contains no nuclear marker cannot be processed and are excluded from its evaluation.

Lastly, for InstanSeg~\citep{goldsborough2024instanseg} we employ the pretrained \texttt{fluorescence\_nuclei\_and\_cells} model instance and apply it to the full marker stack with the built-in normalization enabled, and retain the predicted whole-cell masks.

\paragraph{Results.} We report benchmark results per dataset in terms of instance F1-score at $\tau=0.5$ IoU threshold in \tabref{tab:segmentation_f1_tau_05}.

\begin{table}[ht]
\centering
\caption{Cross-cohort cell segmentation results. Instance F1-score at $\tau=0.5$ IoU threshold per dataset and model. Best performing model per dataset is highlighted in bold, the second-best is underlined.}
\label{tab:segmentation_f1_tau_05}
\begin{tabular}{lllll}
\toprule
Model & VirTues & Cellpose & InstanSeg & StarDist \\
Dataset &  &  &  &  \\
\midrule
\citet{cords2023cancer} & \textbf{0.801} & \underline{0.312} & 0.251 & 0.120 \\
\citet{cords2024cancer} & \textbf{0.742} & 0.326 & \underline{0.369} & 0.128 \\
\citet{danenberg2022breast} & \textbf{0.537} & 0.299 & \underline{0.340} & 0.210 \\
\citet{lin2023high} & 0.340 & \textbf{0.495} & \underline{0.454} & 0.300 \\
\citet{meyer2025stratification} & \underline{0.597} & \textbf{0.694} & 0.157 & 0.408 \\
\citet{moldoveanu2022spatially} & \textbf{0.622} & \underline{0.564} & 0.485 & 0.473 \\
\citet{rigamonti2024integrating} & \textbf{0.498} & \underline{0.384} & 0.287 & 0.089 \\
\citet{eling2025multi} & \textbf{0.700} & \underline{0.595} & 0.418 & 0.368 \\
\bottomrule
\end{tabular}
\end{table}

\subsection{Details on Diagnostics and Therapeutic Prediction Benchmarks} \label{app:sec:benchmark_tissue}

\paragraph{Benchmark procedure.}
Tissue-level tasks are evaluated with a common multiple-instance learning pipeline, so that methods differ only in how tissue representations are computed. For each tissue image, we sample non-overlapping crops from a regular grid and retain those covering at least $30\%$ tissue. Each method then encodes the retained crops into a bag of embeddings, and the bag is aggregated and classified by a gated multi-head attention-based MIL (ABMIL) model~\citep{ilse2018attention} with $8$ attention heads and a hidden dimension of $256$. We train ABMIL with Adam~\citep{kingma2014adam}, a batch size of 24, a learning rate of $5 \times 10^{-4}$, a weight decay of $10^{-5}$ and cross-entropy loss, for at most $100$ epochs with early stopping on the validation loss and a patience of $10$ epochs. For every dataset, we use a patient-level $80/20$ train-test split and hold out a stratified $20\%$ subset of the training data for validation.

Several label spaces are coarsened prior to training. For cancer subtyping on \texttt{cords2024cancer}, we restrict the task to the two major subtypes, adenocarcinoma and squamous cell carcinoma. For cancer grading on \texttt{cords2024cancer} and \texttt{danenberg2022breast}, we group the three grades $\{1, 2, 3\}$ into $\{1\text{--}2, 3\}$, and for risk stratification on \texttt{meyer2025stratification} we group the provided risk levels into $\{\text{low--medium}, \text{high}\}$, in both cases to reduce class imbalance.

\paragraph{Methods and baselines.}
We evaluate three tissue encoders: KRONOS~\citep{shaban2025foundation}, VirTues~\citep{wenckstern2026virtual} and a ResNet-based encoding approach~\citep{sorin2023single}. In all three cases, image preprocessing uses dataset-wise, per-channel statistics. 

For KRONOS, we resample all images to the pre-training resolution of $0.5\,\mu$m/px and map the channel names provided by the \sporaio tissue loader to the KRONOS channel index. Using a tile size of $224\times224$ pixels, each tile is represented by the concatenation of its per-marker embeddings. Following \citet{shaban2025foundation}, we reduce the resulting tile-level tokens with a PCA to $256$ principal components. Images are preprocessed by $99$th-percentile clipping followed by z-score normalization.

Following the original implementation of VirTues, we compute bags of $8\times8$ px patch tokens using a tile size of $128\times128$ px at a resolution of $1\,\mu$m/px, and take the bag of final-layer patch summary tokens as tile representations. Images are preprocessed by $99$th-percentile clipping, a $\log(1+x)$ transformation and z-score normalization.

We initialize a ResNet-50 with ImageNet-1k pre-trained weights and encode each marker channel of a $224\times224$ px tile separately, replicating the channel three times to imitate an RGB input. For each channel we extract the penultimate-layer features and concatenate them across channels. Following \citet{sorin2023single}, the concatenated embeddings are reduced with an sparse mini-batch PCA to the top $9$ principal components. Images are preprocessed by $99$th-percentile clipping, a $\log(1+x)$ transformation and z-score normalization.

\clearpage

\subsection{Additional Virtual Staining Results}

\begin{figure}[H] 
  \centering
  \includegraphics[trim=0.0cm 11.5cm 11.5cm 0.0cm,clip,width=\textwidth, page = 15]{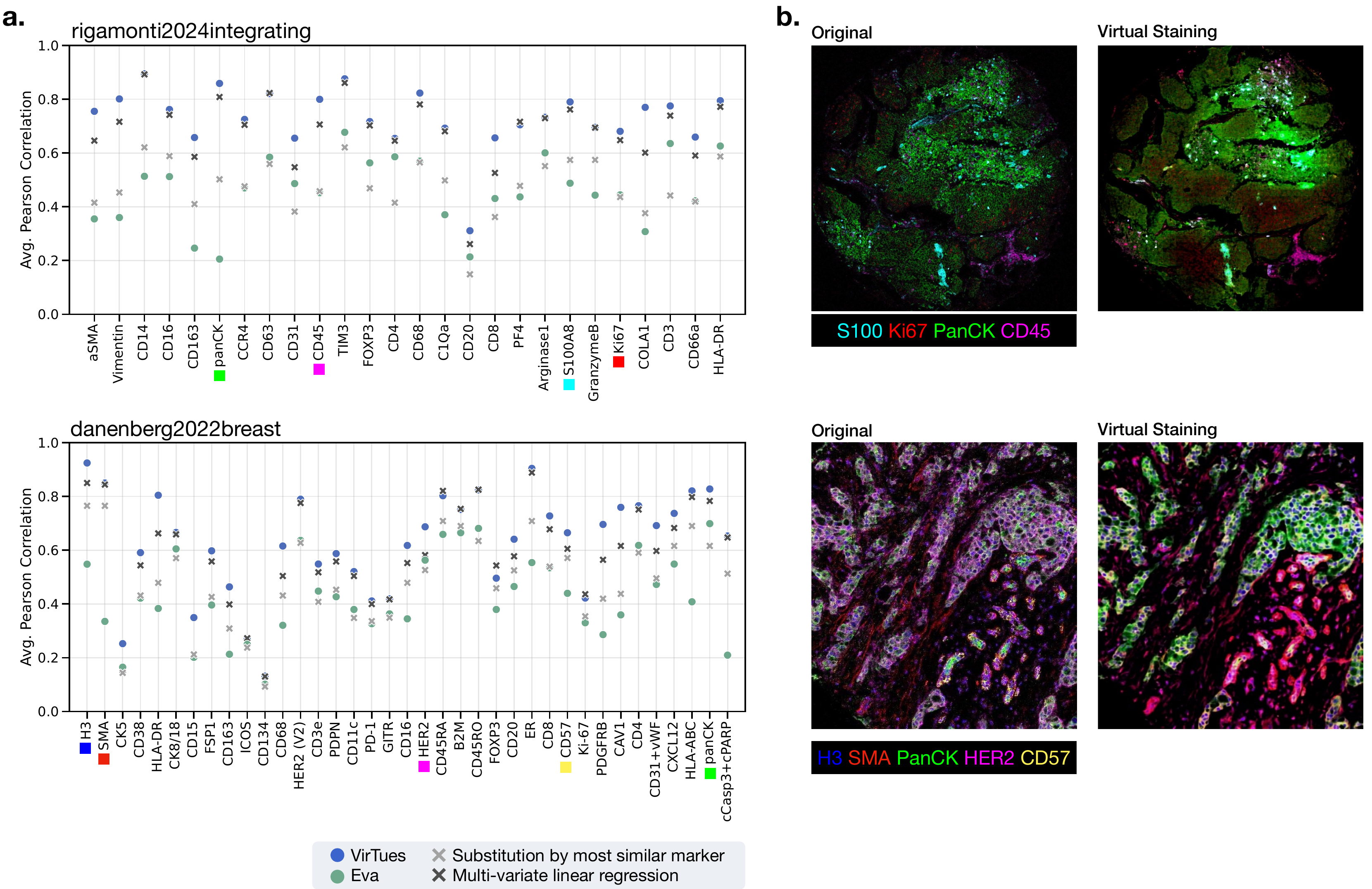}
  \caption{
    \textbf{Virtual staining results.}
    \textbf{a.} Average Pearson correlation between predicted and ground-truth markers across multiplex images; the baseline (light-gray cross) substitutes the closest available marker for the target.
    \textbf{b.} Representative multiplex images and their virtual staining predictions obtained from VirTues.
  }
  \label{app:fig:virtual_staining_2}
\end{figure}

\begin{table}[ht]
\centering
\caption{\textbf{Per-marker virtual staining results on \citet{hickey2023organization}.} Pearson correlation and SSIM for every marker predicted by at least one model. Dashes mark markers absent from a model's output panel (GigaTIME: 10, HistoPlexer: 5, ROSIE: 23). Best performing model per marker and metric is shown in bold.}\label{app:tab:vs_per_marker_hickey2023organization}
\vspace{0.2cm}
\small
\begin{tabular}{lSSSSSS}
\toprule
 & \multicolumn{2}{c}{GigaTIME} & \multicolumn{2}{c}{HistoPlexer} & \multicolumn{2}{c}{ROSIE} \\
\cmidrule(lr){2-3} \cmidrule(lr){4-5} \cmidrule(lr){6-7}
Marker & {Pearson $\uparrow$} & {SSIM $\uparrow$} & {Pearson $\uparrow$} & {SSIM $\uparrow$} & {Pearson $\uparrow$} & {SSIM $\uparrow$} \\
\midrule
BCL2 & {--} & {--} & {--} & {--} & {\bfseries 0.060} & {\bfseries 0.081} \\
CD11C & {\bfseries 0.034} & {\bfseries 0.379} & {--} & {--} & 0.002 & 0.135 \\
CD138 & {\bfseries 0.022} & {\bfseries 0.364} & {--} & {--} & {--} & {--} \\
CD16 & {\bfseries 0.050} & 0.439 & 0.011 & {\bfseries 0.505} & {--} & {--} \\
CD163 & {--} & {--} & {--} & {--} & {\bfseries 0.057} & {\bfseries 0.027} \\
CD21 & {--} & {--} & {--} & {--} & {\bfseries 0.038} & {\bfseries 0.108} \\
CD3 & {\bfseries 0.072} & 0.395 & 0.007 & {\bfseries 0.624} & 0.042 & 0.123 \\
CD31 & {--} & {--} & 0.013 & {\bfseries 0.722} & {\bfseries 0.074} & 0.040 \\
CD34 & 0.052 & {\bfseries 0.497} & {--} & {--} & {\bfseries 0.082} & 0.081 \\
CD38 & {--} & {--} & {--} & {--} & {\bfseries 0.043} & {\bfseries 0.028} \\
CD4 & {\bfseries 0.065} & {\bfseries 0.424} & {--} & {--} & 0.048 & 0.074 \\
CD44 & {--} & {--} & {--} & {--} & {\bfseries 0.091} & {\bfseries 0.084} \\
CD45 & {--} & {--} & {--} & {--} & {\bfseries 0.083} & {\bfseries 0.076} \\
CD45RO & {--} & {--} & {--} & {--} & {\bfseries 0.078} & {\bfseries 0.093} \\
CD66 & {--} & {--} & {--} & {--} & {\bfseries 0.052} & {\bfseries 0.120} \\
CD68 & 0.047 & {\bfseries 0.460} & {--} & {--} & {\bfseries 0.071} & 0.109 \\
CD8 & 0.040 & 0.373 & 0.002 & {\bfseries 0.582} & {\bfseries 0.043} & 0.041 \\
COLLAGENIV & {--} & {--} & {--} & {--} & {\bfseries 0.147} & {\bfseries 0.081} \\
GATA3 & {--} & {--} & {--} & {--} & {\bfseries -0.001} & {\bfseries 0.072} \\
HLADR & {--} & {--} & 0.010 & {\bfseries 0.586} & {\bfseries 0.058} & 0.112 \\
KI67 & {\bfseries 0.058} & {\bfseries 0.442} & {--} & {--} & 0.048 & 0.083 \\
PANCK & 0.062 & {\bfseries 0.417} & {--} & {--} & {\bfseries 0.110} & 0.114 \\
PODOPLANIN & {--} & {--} & {--} & {--} & {\bfseries 0.051} & {\bfseries 0.034} \\
SMA & {--} & {--} & {--} & {--} & {\bfseries 0.089} & {\bfseries 0.088} \\
VIMENTIN & {--} & {--} & {--} & {--} & {\bfseries 0.109} & {\bfseries 0.110} \\
\bottomrule
\vspace{-0.4cm}
\end{tabular}
\end{table}

\begin{table}[ht]
\centering
\caption{\textbf{Per-marker virtual staining results on \citet{lin2023high}.} Pearson correlation and SSIM for every marker predicted by at least one model. Dashes mark markers absent from a model's output panel (GigaTIME: 10, HistoPlexer: 3, ROSIE: 15). Best performing model per marker and metric is shown in bold.}\label{app:tab:vs_per_marker_lin2023high}
\vspace{0.2cm}
\small
\begin{tabular}{lSSSSSS}
\toprule
 & \multicolumn{2}{c}{GigaTIME} & \multicolumn{2}{c}{HistoPlexer} & \multicolumn{2}{c}{ROSIE} \\
\cmidrule(lr){2-3} \cmidrule(lr){4-5} \cmidrule(lr){6-7}
Marker & {Pearson $\uparrow$} & {SSIM $\uparrow$} & {Pearson $\uparrow$} & {SSIM $\uparrow$} & {Pearson $\uparrow$} & {SSIM $\uparrow$} \\
\midrule
CD163 & {--} & {--} & {--} & {--} & {\bfseries 0.030} & {\bfseries 0.228} \\
CD20 & 0.039 & 0.164 & -0.004 & 0.053 & {\bfseries 0.069} & {\bfseries 0.254} \\
CD3 & 0.062 & 0.240 & 0.009 & 0.017 & {\bfseries 0.069} & {\bfseries 0.399} \\
CD4 & {\bfseries 0.079} & 0.304 & {--} & {--} & 0.075 & {\bfseries 0.316} \\
CD45 & {--} & {--} & {--} & {--} & {\bfseries 0.119} & {\bfseries 0.265} \\
CD45RO & {--} & {--} & {--} & {--} & {\bfseries 0.130} & {\bfseries 0.320} \\
CD68 & 0.026 & {\bfseries 0.403} & {--} & {--} & {\bfseries 0.053} & 0.284 \\
CD8 & 0.038 & {\bfseries 0.394} & 0.007 & 0.099 & {\bfseries 0.060} & 0.251 \\
DAPI & 0.210 & {\bfseries 0.298} & {--} & {--} & {\bfseries 0.250} & 0.189 \\
ECAD & {--} & {--} & {--} & {--} & {\bfseries 0.058} & {\bfseries 0.208} \\
KI67 & {\bfseries 0.087} & {\bfseries 0.267} & {--} & {--} & 0.075 & 0.244 \\
PANCK & {\bfseries 0.059} & {\bfseries 0.180} & {--} & {--} & 0.040 & 0.173 \\
PD1 & {\bfseries 0.093} & 0.156 & {--} & {--} & 0.059 & {\bfseries 0.236} \\
PDL1 & {\bfseries 0.082} & 0.230 & {--} & {--} & 0.069 & {\bfseries 0.303} \\
SMA & {--} & {--} & {--} & {--} & {\bfseries 0.052} & {\bfseries 0.300} \\
\bottomrule
\vspace{-0.4cm}
\end{tabular}
\end{table}

\begin{table}[ht]
\centering
\caption{\textbf{Per-marker virtual staining results: \citet{phillips2021immune}.} Pearson correlation and SSIM for every marker predicted by at least one model. Dashes mark markers absent from a model's output panel (GigaTIME: 15, HistoPlexer: 6, ROSIE: 29). Best performing model per marker and metric is shown in bold.}\label{app:tab:vs_per_marker_phillips2021immune}
\vspace{0.2cm}
\small
\begin{tabular}{lSSSSSS}
\toprule
 & \multicolumn{2}{c}{GigaTIME} & \multicolumn{2}{c}{HistoPlexer} & \multicolumn{2}{c}{ROSIE} \\
\cmidrule(lr){2-3} \cmidrule(lr){4-5} \cmidrule(lr){6-7}
Marker & {Pearson $\uparrow$} & {SSIM $\uparrow$} & {Pearson $\uparrow$} & {SSIM $\uparrow$} & {Pearson $\uparrow$} & {SSIM $\uparrow$} \\
\midrule
BCL2 & {--} & {--} & {--} & {--} & {\bfseries 0.138} & {\bfseries 0.096} \\
CD11C & {\bfseries 0.069} & {\bfseries 0.312} & {--} & {--} & 0.028 & 0.106 \\
CD138 & {\bfseries 0.085} & {\bfseries 0.223} & {--} & {--} & {--} & {--} \\
CD16 & {\bfseries 0.125} & {\bfseries 0.417} & 0.014 & 0.342 & {--} & {--} \\
CD163 & {--} & {--} & {--} & {--} & {\bfseries 0.043} & {\bfseries 0.038} \\
CD20 & {\bfseries 0.029} & 0.281 & 0.000 & {\bfseries 0.660} & 0.027 & 0.052 \\
CD3 & {\bfseries 0.210} & 0.297 & 0.004 & {\bfseries 0.417} & 0.094 & 0.142 \\
CD31 & {--} & {--} & -0.000 & {\bfseries 0.790} & {\bfseries 0.056} & 0.022 \\
CD34 & 0.051 & {\bfseries 0.315} & {--} & {--} & {\bfseries 0.080} & 0.080 \\
CD38 & {--} & {--} & {--} & {--} & {\bfseries 0.140} & {\bfseries 0.103} \\
CD4 & {\bfseries 0.182} & {\bfseries 0.352} & {--} & {--} & 0.044 & 0.089 \\
CD45 & {--} & {--} & {--} & {--} & {\bfseries 0.103} & {\bfseries 0.090} \\
CD45RO & {--} & {--} & {--} & {--} & {\bfseries 0.042} & {\bfseries 0.052} \\
CD68 & 0.072 & {\bfseries 0.340} & {--} & {--} & {\bfseries 0.082} & 0.111 \\
CD8 & {\bfseries 0.079} & 0.305 & 0.005 & {\bfseries 0.651} & 0.060 & 0.035 \\
COLLAGENIV & {--} & {--} & {--} & {--} & {\bfseries 0.111} & {\bfseries 0.072} \\
DAPI & 0.383 & {\bfseries 0.238} & {--} & {--} & {\bfseries 0.385} & 0.128 \\
FOXP3 & {--} & {--} & {--} & {--} & {\bfseries 0.079} & {\bfseries 0.058} \\
GATA3 & {--} & {--} & {--} & {--} & {\bfseries 0.097} & {\bfseries 0.075} \\
GRANZYMEB & {--} & {--} & {--} & {--} & {\bfseries -0.003} & {\bfseries 0.006} \\
HLADR & {--} & {--} & 0.010 & {\bfseries 0.595} & {\bfseries 0.059} & 0.076 \\
ICOS & {--} & {--} & {--} & {--} & {\bfseries 0.111} & {\bfseries 0.191} \\
IDO1 & {--} & {--} & {--} & {--} & {\bfseries 0.050} & {\bfseries 0.222} \\
KI67 & 0.097 & {\bfseries 0.240} & {--} & {--} & {\bfseries 0.106} & 0.151 \\
LAG3 & {--} & {--} & {--} & {--} & {\bfseries 0.141} & {\bfseries 0.122} \\
PANCK & 0.022 & {\bfseries 0.369} & {--} & {--} & {\bfseries 0.032} & 0.048 \\
PD1 & {\bfseries 0.199} & {\bfseries 0.243} & {--} & {--} & 0.137 & 0.114 \\
PDL1 & {\bfseries 0.121} & {\bfseries 0.287} & {--} & {--} & 0.054 & 0.180 \\
PODOPLANIN & {--} & {--} & {--} & {--} & {\bfseries 0.014} & {\bfseries 0.059} \\
TBET & {\bfseries 0.163} & {\bfseries 0.228} & {--} & {--} & {--} & {--} \\
VIMENTIN & {--} & {--} & {--} & {--} & {\bfseries 0.098} & {\bfseries 0.149} \\
VISTA & {--} & {--} & {--} & {--} & {\bfseries 0.094} & {\bfseries 0.168} \\
\bottomrule
\vspace{-0.4cm}
\end{tabular}
\end{table}

\begin{table}[ht]
\centering
\caption{\textbf{Per-marker virtual staining results on \citet{schurch2020coordinated}.} Pearson correlation and SSIM for every marker predicted by at least one model. Dashes mark markers absent from a model's output panel (GigaTIME: 14, HistoPlexer: 7, ROSIE: 33). Best performing model per marker and metric is shown in bold.}\label{app:tab:vs_per_marker_schurch2020coordinated}
\vspace{0.2cm}
\small
\begin{tabular}{lSSSSSS}
\toprule
 & \multicolumn{2}{c}{GigaTIME} & \multicolumn{2}{c}{HistoPlexer} & \multicolumn{2}{c}{ROSIE} \\
\cmidrule(lr){2-3} \cmidrule(lr){4-5} \cmidrule(lr){6-7}
Marker & {Pearson $\uparrow$} & {SSIM $\uparrow$} & {Pearson $\uparrow$} & {SSIM $\uparrow$} & {Pearson $\uparrow$} & {SSIM $\uparrow$} \\
\midrule
BCL2 & {--} & {--} & {--} & {--} & {\bfseries 0.037} & {\bfseries 0.115} \\
CD11C & {\bfseries 0.011} & {\bfseries 0.248} & {--} & {--} & -0.020 & 0.142 \\
CD138 & {\bfseries -0.026} & {\bfseries 0.155} & {--} & {--} & {--} & {--} \\
CD16 & {\bfseries 0.036} & 0.212 & 0.020 & {\bfseries 0.246} & {--} & {--} \\
CD163 & {--} & {--} & {--} & {--} & {\bfseries 0.024} & {\bfseries 0.035} \\
CD20 & {\bfseries 0.002} & 0.247 & -0.002 & {\bfseries 0.411} & -0.021 & 0.036 \\
CD21 & {--} & {--} & {--} & {--} & {\bfseries 0.013} & {\bfseries 0.083} \\
CD3 & {\bfseries 0.149} & 0.223 & 0.006 & {\bfseries 0.320} & 0.082 & 0.190 \\
CD31 & {--} & {--} & 0.000 & {\bfseries 0.822} & {\bfseries 0.027} & 0.022 \\
CD34 & {\bfseries 0.038} & {\bfseries 0.253} & {--} & {--} & 0.022 & 0.058 \\
CD38 & {--} & {--} & {--} & {--} & {\bfseries 0.107} & {\bfseries 0.125} \\
CD4 & {\bfseries 0.101} & {\bfseries 0.196} & {--} & {--} & 0.042 & 0.117 \\
CD44 & {--} & {--} & {--} & {--} & {\bfseries 0.038} & {\bfseries 0.124} \\
CD45 & {--} & {--} & {--} & {--} & {\bfseries 0.042} & {\bfseries 0.095} \\
CD45RO & {--} & {--} & {--} & {--} & {\bfseries 0.066} & {\bfseries 0.086} \\
CD68 & {\bfseries 0.067} & {\bfseries 0.285} & {--} & {--} & 0.062 & 0.109 \\
CD79A & {--} & {--} & {--} & {--} & {\bfseries 0.013} & {\bfseries 0.050} \\
CD8 & {\bfseries 0.063} & 0.266 & 0.008 & {\bfseries 0.493} & 0.054 & 0.068 \\
COLLAGENIV & {--} & {--} & {--} & {--} & {\bfseries -0.002} & {\bfseries 0.041} \\
EPCAM & {--} & {--} & {--} & {--} & {\bfseries 0.083} & {\bfseries 0.146} \\
FOXP3 & {--} & {--} & {--} & {--} & {\bfseries 0.075} & {\bfseries 0.114} \\
GATA3 & {--} & {--} & {--} & {--} & {\bfseries 0.059} & {\bfseries 0.078} \\
GRANZYMEB & {--} & {--} & {--} & {--} & {\bfseries -0.001} & {\bfseries 0.011} \\
HLADR & {--} & {--} & {\bfseries -0.008} & {\bfseries 0.560} & -0.029 & 0.057 \\
ICOS & {--} & {--} & {--} & {--} & {\bfseries 0.029} & {\bfseries 0.101} \\
IDO1 & {--} & {--} & {--} & {--} & {\bfseries 0.079} & {\bfseries 0.156} \\
KI67 & {\bfseries 0.060} & {\bfseries 0.228} & {--} & {--} & 0.059 & 0.154 \\
LAG3 & {--} & {--} & {--} & {--} & {\bfseries 0.093} & {\bfseries 0.140} \\
MELANA & {--} & {--} & {\bfseries 0.017} & {\bfseries 0.185} & {--} & {--} \\
PANCK & -0.002 & {\bfseries 0.272} & {--} & {--} & {\bfseries 0.007} & 0.042 \\
PD1 & {\bfseries 0.170} & 0.119 & {--} & {--} & 0.133 & {\bfseries 0.154} \\
PDL1 & {\bfseries 0.012} & {\bfseries 0.168} & {--} & {--} & 0.005 & 0.129 \\
PODOPLANIN & {--} & {--} & {--} & {--} & {\bfseries 0.021} & {\bfseries 0.051} \\
SMA & {--} & {--} & {--} & {--} & {\bfseries 0.049} & {\bfseries 0.068} \\
TBET & {\bfseries 0.103} & {\bfseries 0.225} & {--} & {--} & {--} & {--} \\
VIMENTIN & {--} & {--} & {--} & {--} & {\bfseries 0.066} & {\bfseries 0.132} \\
VISTA & {--} & {--} & {--} & {--} & {\bfseries 0.093} & {\bfseries 0.217} \\
\bottomrule
\vspace{-0.4cm}
\end{tabular}
\end{table}

\end{document}